\documentclass{article} %
\usepackage{iclr2027_conference,times}

\usepackage{amsmath,amsfonts,bm}

\def\eqref#1{equation~\ref{#1}}

\def\1{\bm{1}}

\DeclareMathAlphabet{\mathsfit}{\encodingdefault}{\sfdefault}{m}{sl}
\SetMathAlphabet{\mathsfit}{bold}{\encodingdefault}{\sfdefault}{bx}{n}

\usepackage{xcolor}
\usepackage{hyperref}
\usepackage{url}
\usepackage{xspace}
\usepackage{graphicx}
\usepackage[labelfont=bf,font=small,labelsep=period,skip=5pt]{caption}
\usepackage{subcaption}
\usepackage{pifont}
\usepackage{wasysym}

\newcommand{\yes}{\ensuremath{\CIRCLE}}
\newcommand{\half}{\ensuremath{\LEFTcircle}}
\newcommand{\no}{\ensuremath{\Circle}}
\usepackage{booktabs}
\usepackage{amsmath,amssymb}
\usepackage[capitalize,noabbrev]{cleveref}
\usepackage{float}
\usepackage{placeins}
\usepackage{microtype}
\usepackage{enumitem}
\setlist{leftmargin=*,itemsep=2pt,topsep=3pt,parsep=0pt}
\usepackage{sear_frontmatter}
\hypersetup{colorlinks=true,linkcolor=searlink,citecolor=searlink,urlcolor=searlink,
  pdftitle={SEAR: Self-Evolving Agentic Robotics},
  pdfauthor={Xinyi Wang, Bangzheng Li, Hanyu Wang, Lin Zhang, Andrea Yaoyun Cui, Zhenggang Tang, Shyamal Buch, Dejia Xu}}

\def\modelname{\texttt{\textbf{SEAR}}\xspace}
\def\modelfullname{\texttt{\textbf{S}elf-\textbf{E}volving \textbf{A}gentic \textbf{R}obotics}\xspace}

\newcommand{\searwebsiteurl}{https://sear.bot}
\newcommand{\seargithuburl}{https://github.com/Self-Evolving-Agentic-Robotics/SEAR}

\title{\modelname: Self-Evolving Agentic Robotics}  

\author{Xinyi Wang\textsuperscript{*,$\dagger$} \qquad Bangzheng Li\textsuperscript{*,$\dagger$} \qquad Hanyu Wang\textsuperscript{$\ddagger$} \qquad Lin Zhang\\[2pt]
Andrea Yaoyun Cui \qquad Zhenggang Tang \qquad Shyamal Buch\textsuperscript{\S} \qquad Dejia Xu\textsuperscript{$\dagger$,\S}\\[5pt]
\raisebox{-0.2\height}{\includegraphics[height=1.1em]{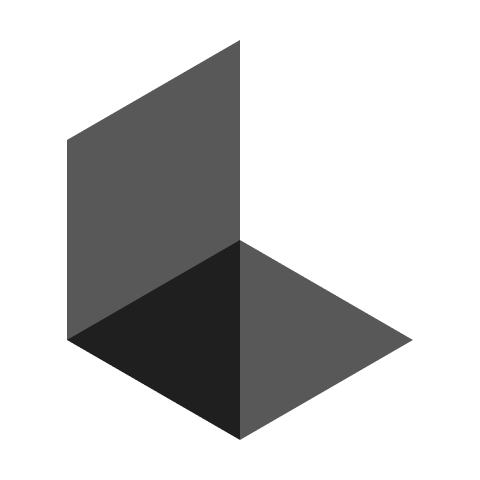}}\hspace{0.3em}Luma AI %
\\[3pt]
{\footnotesize\color{searmuted}\textsuperscript{*}Co-first authors \quad \textsuperscript{$\dagger$}Project leads \quad \textsuperscript{$\ddagger$}Real robot lead \quad \textsuperscript{\S}Co-last authors}\\[4pt]
\searbutton{\faGlobe}{Website}{\searwebsiteurl}\hspace{0.6em}\searbutton{\faGithub}{GitHub}{\seargithuburl}
}

\iclrfinalcopy %
\newcommand{\searheadlogo}{\raisebox{-3pt}{\includegraphics[height=22pt]{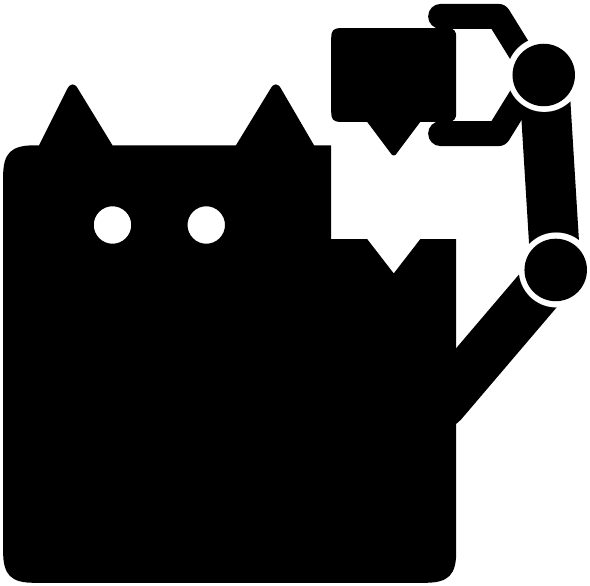}}}
\begin{document} \raggedbottom
\lhead{\searheadlogo}

\vspace*{-14pt}
\begin{searfrontmatter}
\begingroup
\centering
\setlength{\parskip}{0pt}%
\includegraphics[width=\linewidth]{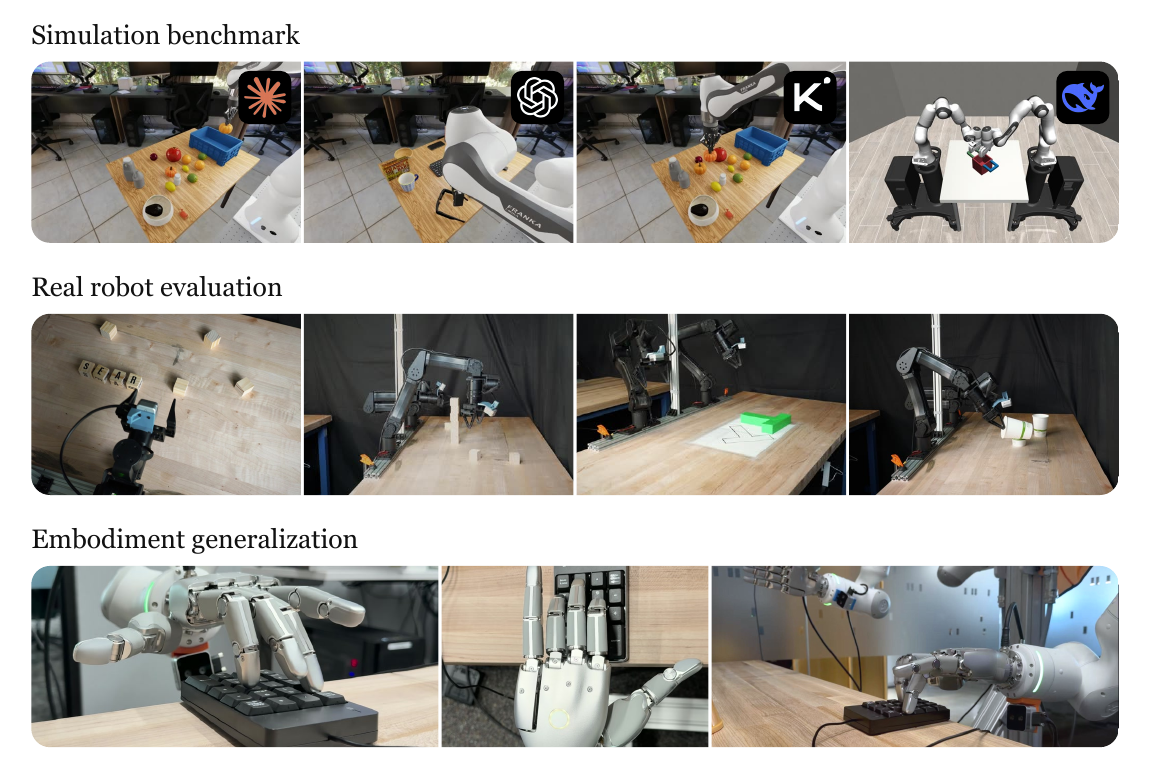}\par
\captionsetup{skip=3pt}%
\captionof{figure}{\modelname{} is a framework and benchmark for \textit{self-evolving agentic robotics}: frontier language-model agents that can improve robot performance from their own experience. We disentangle real progress across (top) 330 simulated tasks, two physics engines, and (middle) five physical tasks on a two-arm rig, tested on human-designed scenes. Bottom: the same harness on a dexterous hand.}
\vspace{2pt}
\label{fig:teaser}
\par\endgroup
\begin{abstract}
Frontier language-model agents can now improve robots on their own: they write control code, test it, and keep what works as skills and memories for the next task. We introduce \modelname{} (Self-Evolving Agentic Robotics), a framework and benchmark built around two questions. Is the agent really learning, or just trying more often? And does what it learns still help on new tasks, in a new simulator, and on a real robot? The framework runs an agent in three modes: it writes code that becomes the policy (code-as-policy), acts live in the control loop (agent-as-policy), or works as a researcher that fine-tunes a vision-language-action model (auto-research). The benchmark adds controls that separate retained experience from repeated attempts, familiar tasks, privileged simulator state, larger budgets, and luck. Across seven frontier language models, 330 simulated and five physical tasks, and over four years of agent runtime, we highlight three findings. First, the learning curve of a robotic agent depends both on the overall time budget and the task difficulty relative to model capability. Harder tasks make longer sessions worthwhile; on easier tasks, re-rolling short attempts does just as well. Second, control code and shared routines transfer across tasks and simulators, while task-specific solutions do not. On a real robot, we observe a performance gap between agent-designed practice and novel human-designed scenes. Third, different models evolve more effectively in different ways. Some do best acting live, others writing code first. And a text-only model keeps pace with a comparable multimodal one: perception helps with error assessment, but agentic reasoning and tool use can matter more. We release the benchmark and codebase to support future research.
\end{abstract}

\end{searfrontmatter}
\lhead{\searheadlogo}
\rhead{\modelname: Self-Evolving Agentic Robotics}

\section{Introduction}
\label{sec:intro}

Language-model (LM) agents can now improve robots on their own. Given a task and a way to check success, a coding agent writes a control program, runs it, reads what went wrong, and revises the program. It can save what worked as a skill for the next task, or collect its own rollouts and fine-tune a vision-language-action model (VLA) on them. Recent systems run this self-improvement loop in simulation and on real robots~\citep{capx, rats, aspire, enpire, waddle2026}. Robots need this ability because, once deployed, they meet objects, conditions, and tasks that no one prepared them for, and no engineer is there to rewrite their programs.

A rising success rate, however, is not proof that the agent learned. We ask two questions. \textbf{Is the agent really learning, or just trying more often?} A rising curve can come from resampling, since more independent attempts solve more problems~\citep{brown2024monkeys}, or from one lucky run, a larger budget, or simulator state a real robot never sees. \textbf{And does what it learns still help elsewhere?} A held-out initial state of the practiced task is the smallest possible step away from practice. Experience that is worth keeping should also help on new tasks, in a new simulator, and on a real robot. Prior systems each define their own protocol, so their results cannot answer either question or be compared with one another.

\begin{figure}[!tbp]
    \centering
    \includegraphics[width=\linewidth]{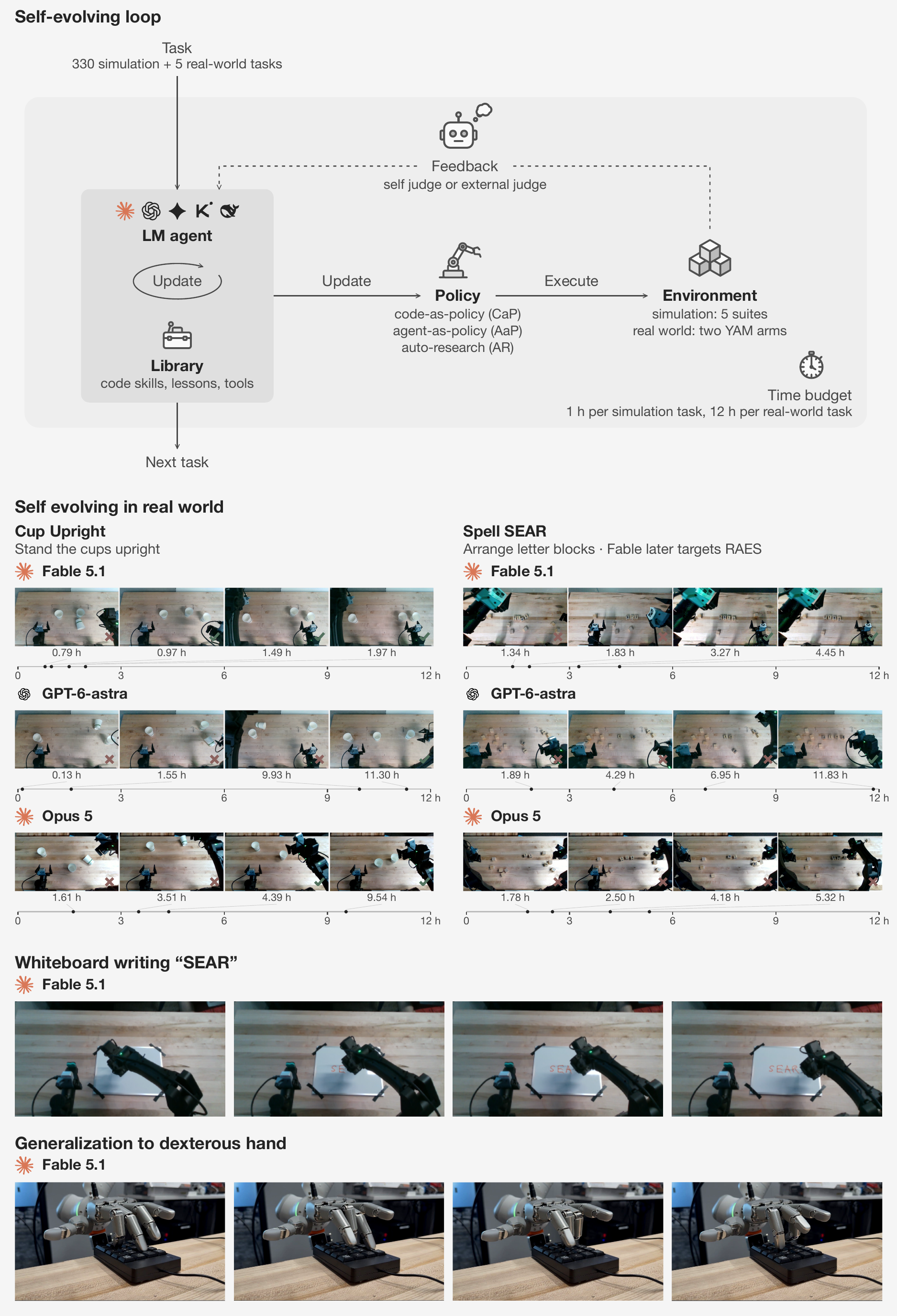}
    \vspace{-6pt}
    \caption{\textbf{The self-evolving loop, and what it looks like on a real robot.} Top: within a time budget, an LM agent with its library updates a policy, by code-as-policy (CaP), agent-as-policy (AaP), or auto-research (AR); the policy executes in simulation or on hardware, the agent's own checks or an external judge return feedback, and the library carries what worked into the next task (\cref{sec:sear}). Bottom: Cup Upright and Spell ``SEAR'' attempts by Fable~5.1, GPT-6-astra, and Opus~5, labeled with elapsed hours and outcomes (\cref{sec:sb2_hardware}), then a whiteboard-writing demo (\href{https://sear.bot}{website}) and a dexterous-hand task (\cref{sec:finding_dexterous}), both by Fable~5.1.}
    \label{fig:demos}
    \vspace{-8pt}
\end{figure}

We introduce \modelname{} (\modelfullname), a framework and benchmark for answering both (\cref{sec:sear}). The framework runs an LM agent through one harness in simulation and on hardware, in three modes: the agent writes a program that becomes the policy (code-as-policy), stays in the control loop and acts live (agent-as-policy), or fine-tunes a VLA on rollouts it collects (auto-research). In code-as-policy, the policy usually contains no learned action model: the LM writes ordinary Python that calls perception, inverse-kinematics, and servo tools. A pretrained VLA is also in the toolbox, but almost no final program uses it (\cref{sec:sb2_no_pixels}). Between tasks the agent keeps a \emph{library} of code skills and written memories. \Cref{fig:demos} shows the loop and a few physical tasks being learned this way. The benchmark pairs every comparison with a control. Three short restarts at the same total budget (best-of-3) and a run with no library separate learning from resampling. A sealed judge grades frozen policies on initial states the agent never sees. Repeated runs expose luck. Five generalization tiers measure how far a gain reaches.

We use \modelname{} to study seven frontier LMs on 330 simulated tasks in two physics engines and five physical tasks, with over four years of agent runtime in total. Three findings stand out:
\begin{itemize}
\item \textbf{The agents learn, but not always} (\cref{sec:sb2_learning}). The learning curve of a robotic agent depends both on the overall time budget and the task difficulty relative to model capability. Harder tasks make longer sessions worthwhile; on easier tasks, re-rolling short attempts does just as well. 
An inherited library helps most where a model succeeds sometimes but not reliably on its own. It adds little when the model almost always fails or almost always succeeds without it. On RoboLab, GPT-5.6-sol scores 46.9\% with no library, 42.4\% with its own library, and 60.6\% with another model's library, so experience saved as code transfers across models. A library can also be harmful when brittle skills or outright mistakes are carried forward.
\item \textbf{Code and shared routines transfer; task-specific solutions do not} (\cref{sec:sb2_transfer}). Grasp and control routines carry across tasks and simulators, and one program, written before the agent is told its tasks, succeeds on scenes it never explored when its routines work there. Code policies keep working under hardware faults that do not move the camera, and a trained VLA keeps working when the camera is moved. On a real robot, success in practice overstates success on new scenes.
\item \textbf{The LM matters, and so does how it likes to act} (\cref{sec:sb2_llm}). The strongest model leads every simulated family, and the gap between models widens as tasks get harder. A text-only model beats a multimodal model with the same general-capability score on the hardest family, RoboLab, with and without a library. On hardware, GPT-6-astra more than doubles its score when it acts live (from 41\% to 88\%), but acting live is slow: a live episode takes a median of 12.8 minutes of robot time, against 3.8 for a frozen code policy. Fable~5.1, placed in the live loop, writes and runs scripts for almost every command, and scores about the same either way.
\end{itemize}

Our contributions are:
\begin{itemize}
\item \textbf{An empirical study} of seven frontier LMs improving robots from their own experience, across three modes (code-as-policy, agent-as-policy, and auto-research), 330 simulated tasks in two physics engines, and five physical tasks.
\item \textbf{A benchmark} whose matched controls separate what an agent retains from what it gets by trying more often, seeing familiar tasks, reading privileged information, spending a larger budget, or being lucky, with a sealed judge that grades programs and trained checkpoints the same way. It extends to other MuJoCo- and Isaac-based tasks that expose a primitive API and a success predicate.
\item \textbf{An agentic framework and harness} that runs code-as-policy, agent-as-policy, and auto-research agents against the same environments, tools, and libraries in simulation and on physical robots (code: \url{https://github.com/Self-Evolving-Agentic-Robotics/SEAR}).
\end{itemize}

\section{\modelname: Framework and Benchmark}
\label{sec:sear}

\subsection{Three Ways to Run an Agent}
\label{sec:sb2_framework}

\modelname{} organizes LM-driven robot learning~\citep{capx, aspire, rats, enpire, gao2026a} by what the agent produces and what runs on the robot (\cref{fig:demos}):
\begin{itemize}
    \item \textbf{Code-as-policy (CaP).} The agent writes a program, and the program is the policy. No LM runs at execution time, so the policy is fast, and it can be replayed and re-graded without the agent.
    \item \textbf{Agent-as-policy (AaP).} The LM stays in the control loop and chooses each command from live observations. Every command waits on a model call, so episodes are slower and evaluation costs grow with each rollout.
    \item \textbf{Auto-research (AR).} The agent runs training experiments on another model~\citep{rank2026posttrainbench}: it collects rollouts from code policies it treats as teachers, curates them, fine-tunes the VLA $\pi_{0.5}$~\citep{black2025pi05}, evaluates it, and decides what to try next.
\end{itemize}
Every agent starts from the same toolbox: SAM3~\citep{carion2026sam} for segmentation, a pretrained $\pi_{0.5}$, and inverse-kinematics and servo interfaces. Each task starts in a fresh workspace. What persists from task to task is the \textbf{library}: versioned code skills and written memories harvested from the best policy of each finished task. AR checkpoints and data never carry over to the next task. Only skills and memories do. Simulation experiments use CaP and AR, because a judge can cheaply re-grade their frozen outputs. The hardware experiments also use AaP. \Cref{tab:sb2_landscape} places existing systems in this space.

\begin{table}[!h]
\centering
\caption{\textbf{Existing self-evolving robot agents as instances of \modelname{}.} \yes{} yes, \half{} partial, \no{} no, -- not reported, as stated in each system's paper or announcement (\cref{app:landscape}); \cref{tab:landscape_recent} adds seven systems released since mid-September 2026. \emph{Controls}: budget-matched restarts and a no-library run. \emph{Judge}: final policy graded on states and logic the agent cannot access.}
\label{tab:sb2_landscape}
\footnotesize
\setlength{\tabcolsep}{2.6pt}
\resizebox{\textwidth}{!}{%
\begin{tabular}{lcccccccl}
\toprule
& \multicolumn{3}{c}{Policy form} & & Library across & & & \\
\cmidrule(lr){2-4}
System & CaP & AaP & AR & Sim / real & tasks & Controls & Judge & Generalization measured \\
\midrule
CaP-X~\citep{capx}       & \yes & \half & \no & both & \half & \half & -- & perturbations, sim-to-real \\
ENPIRE~\citep{enpire}     & \half & \no & \yes & both & \half & \no & \half & qualitative \\
ASPIRE~\citep{aspire}     & \yes & \half & \no & both & \yes & \half & \half & perturbations, new tasks \\
RATs~\citep{rats}         & \yes & \half & \no & both & \yes & \half & \half & perturbations, new tasks, new sim. \\
Waddle~\citep{waddle2026} & \yes & \yes & \half & real & \yes & \no & \no & -- \\
\midrule
\modelname{} (ours)       & \yes & \yes & \yes & both & \yes & \yes & \yes & tiers T1--T5 (\cref{sec:sb2_protocol}) \\
\bottomrule
\end{tabular}%
}
\end{table}

\subsection{What Counts as Self-Evolving}
\label{sec:benchmark}

Evaluating a fixed policy is easy: roll it out on held-out states and count the successes. Evaluating a system that changes itself is harder, because its success rate can rise for five reasons that have nothing to do with retained experience. \modelname{} pairs each with one control (\cref{tab:confounds}).

\begin{table}[t]
\centering
\caption{\textbf{Five ways a self-evolving agent can look better than it is, and the control for each.}}
\label{tab:confounds}
\small
\setlength{\tabcolsep}{4pt}
\begin{tabular}{p{2.6cm}p{7.4cm}p{2.9cm}}
\toprule
Looks like learning & Control in \modelname{} & Results \\
\midrule
More attempts & One long session against the best of three restarts at a third of the budget each (\emph{best-of-3}); runs with and without a library (\emph{Lineage} vs.\ \emph{Vanilla}) & \cref{sec:sb2_sessions,sec:sb2_test_time} \\
Familiar tasks & Scores on TEST initial states never shown to the agent; five tiers of distance from practice & \cref{sec:sb2_transfer} \\
Privileged information & A sealed judge grades the frozen policy with state and logic the agent cannot reach & \cref{sec:seal,app:shortcuts} \\
Luck & Three independent repetitions of the whole run, library included & \cref{sec:sb2_mistakes} \\
A larger budget & Development time matched in every comparison; tokens, environment steps, and GPU time logged beside it & \cref{app:budgets} \\
\bottomrule
\end{tabular}
\end{table}

\subsection{Tasks}
\label{sec:simulation_tasks}

\begin{figure}[t]
    \centering
    \includegraphics[width=\linewidth]{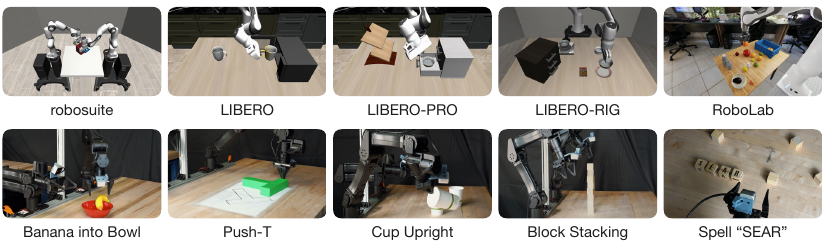}
    \vspace{-6pt}
    \caption{\textbf{The testbed.} Five simulated task families (top) and five physical tasks (bottom): Banana into Bowl, Push-T, Cup Upright, Block Stacking, and Spell ``SEAR''.}
    \label{fig:real_world_tasks_def}
    \vspace{-6pt}
\end{figure}

Five simulated families, from easiest to hardest (\cref{fig:real_world_tasks_def}):
\begin{itemize}
\item \textbf{robosuite}~\citep{zhu2020robosuite}, in MuJoCo~\citep{todorov2012mujoco}: 7 tasks, from lifting a cube to a two-arm handover.
\item \textbf{LIBERO}~\citep{liu2023libero}, in MuJoCo: 40 language-conditioned household tasks.
\item \textbf{LIBERO-PRO}~\citep{zhou2025liberopro}: 160 tasks made by changing the wording, objects, layout, or goal of the 40 LIBERO tasks. The changes also move tasks away from instances a model may have seen in pretraining.
\item \textbf{LIBERO-RIG}~\citep{liberorig}: 18 LIBERO tasks, each with one hidden hardware fault, such as a moved camera, a miscalibrated arm, an actuation error, or worn gripper pads. The 18 tasks cover ten kinds of fault (\cref{app:rig}).
\item \textbf{RoboLab}~\citep{yang2026robolab}, in Isaac Sim~\citep{NVIDIA_Isaac_Sim}: 105 cluttered, randomized tasks on a different physics engine and action API.
\end{itemize}
A run with a library walks the families in this order: robosuite, LIBERO, LIBERO-PRO, then either LIBERO-RIG or RoboLab. The five families hold 330 tasks in all. The physical platform has two YAM-Ultra arms with parallel-jaw grippers, a fixed overview camera, and two wrist cameras. Its five tasks test placement, pushing, reorientation, stacking, and language-conditioned arrangement (\cref{app:real_world_spec}).

\subsection{Protocol}
\label{sec:sb2_protocol}
\label{sec:controls}
\label{sec:tiers}

\textbf{Sealed judge.}\label{sec:seal} The agent works in its own container, where it can run local simulations, inspect failures, install packages, and train checkpoints. When the budget ends or the agent submits, its policy is frozen, and only the frozen program with its imported skills, or the AR checkpoint, is passed to a separate judge. The judge holds the held-out initial states and the grading logic, and ignores any success the policy claims. Each randomized task splits its initial states into DEV seeds the agent uses freely, VAL seeds the judge grades and reports back during development, and TEST seeds graded once at the end. Every reported simulation score is the TEST success rate over ten rollouts per task, and learning curves use VAL scores. Agents still found ways around the seal, such as reading object geometry from asset files; \cref{app:shortcuts} lists them.

\textbf{Conditions.}\label{sec:self_exploration}\label{sec:ar_evaluation} \emph{Vanilla} starts every task with an empty library. \emph{Lineage} is the same setup, but the agent inherits and extends its library task after task. \emph{Best-of-3} splits a Vanilla session's budget into three fresh attempts and keeps the one with the best VAL score. A code-as-policy session gets one hour per task (two on RoboLab). AR gets the same total development time for each suite, a group of tasks within a family. Each condition runs three times per model, except AR on LIBERO, LIBERO-PRO, and RoboLab (one run). \emph{Self-exploration} withholds the instruction and the judge: the agent practices on the scenes with its inherited library, proposes its own tasks, and only then sees the instruction and writes one final policy.

\textbf{Generalization tiers.} \textbf{T1} new initial states of the practiced task; \textbf{T2} the same tasks under changed conditions (LIBERO-PRO scenes, LIBERO-RIG hardware faults), run with frozen policies and no further practice; \textbf{T3} new tasks in the same simulator, after self-exploration; \textbf{T4} a new simulator and action API (MuJoCo to RoboLab's Isaac Sim), after self-exploration; \textbf{T5} the physical robot.

\textbf{Physical protocol.} On hardware nothing resets itself, so each task comes with a reset task the agent must also learn (take the banana out of the bowl, scramble the letters). An external evaluator judges before-and-after images, and practice stops after ten consecutive successes (Block Stacking has no early stop) or at a 12-hour budget per task. The frozen policy is then run once on each of ten scenes staged by an operator (five scenes for AaP, whose episodes are slower). The operator confirms the evaluator's verdict.

\textbf{Agents.} Simulation uses five LMs, each through a coding tool: Claude Opus~5 (Claude Code), GPT-5.6-sol (Codex), and Gemini-3.6-flash, Kimi-k3, and DeepSeek-v4-flash (all through OpenCode). A comparison of models is therefore a comparison of model-and-tool pairs. DeepSeek-v4-flash is text-only: it receives the same images and RGB-D tools as the other models but cannot view the images itself. Hardware uses Claude Fable~5.1, Opus~5, and GPT-6-astra, for seven LMs across simulation and hardware. %

\section{Is It Learning, or Just Trying More?}
\label{sec:sb2_learning}

\begin{finding}
Agents do learn from their own experience, but simpler explanations account for part of the rise in success. Splitting a session into three short attempts does as well as one long session when the model succeeds early. The long session wins only when a short attempt would end before the first success. An inherited library helps most when a model succeeds sometimes but not reliably on its own. It adds little when the model almost always fails or almost always succeeds. What it adds depends on what it contains: a useful skill raises success on later tasks, and a wrong belief saved in it is passed on to later tasks too.
\end{finding}

The results below depend on which language model drives the agent. The library helps most for models that do moderately well without it: a weak model writes little worth saving, and a strong model has little left to gain (\cref{sec:sb2_better_llm}). We first separate learning from trying more often, then ask what the library adds.

\begin{figure}[t]
    \centering
    \includegraphics[width=\linewidth]{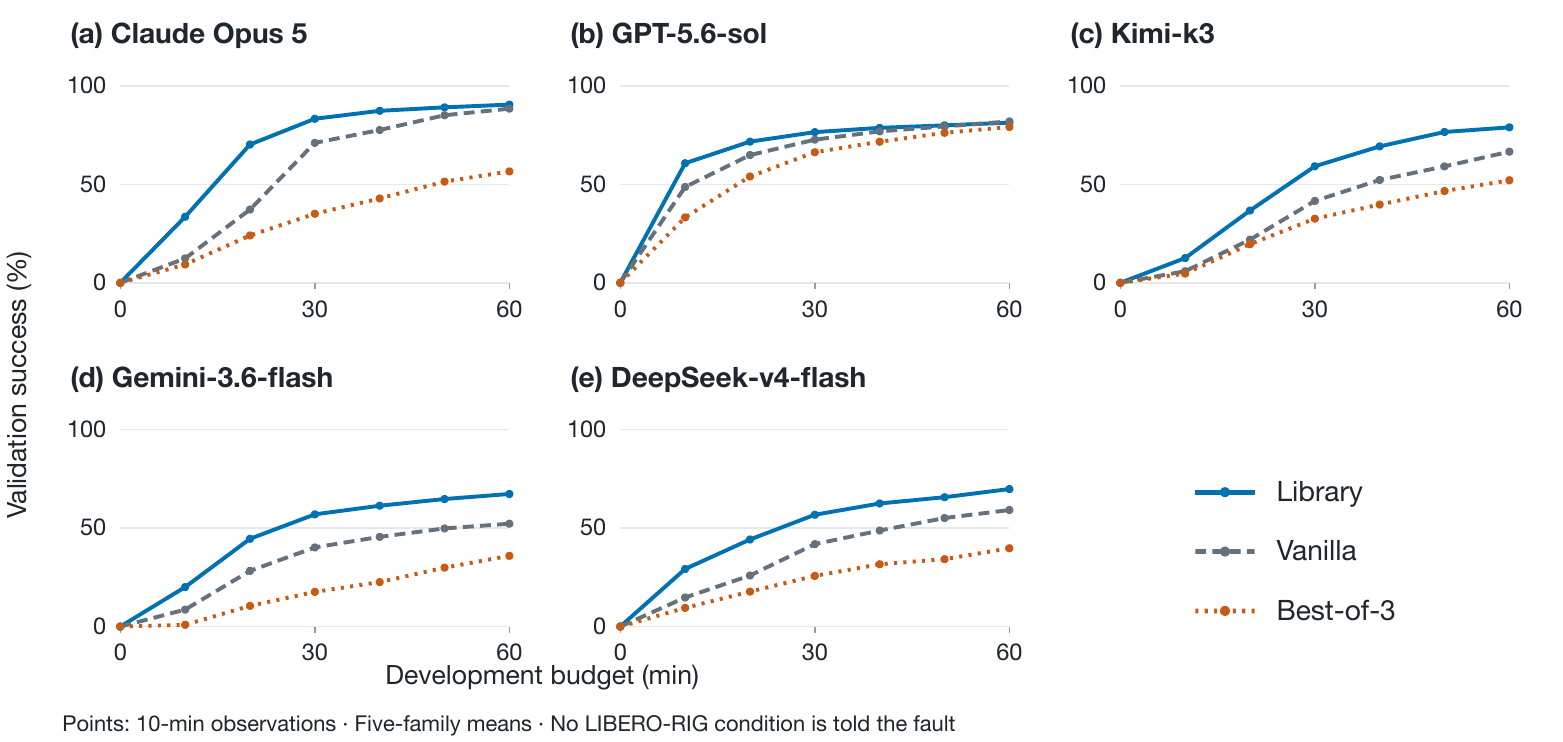}
    \vspace{-6pt}
    \caption{\textbf{Five models over development time.} Validation success, averaged over the five simulated families with equal weight (\cref{tab:fig4_by_family} gives each family), one panel per model: with an inherited library (Lineage, solid), without (Vanilla, dashed), and best-of-3, three attempts at a third of the budget each (dotted). RoboLab's 120-minute budget is rescaled onto the 60-minute axis.}
    \label{fig:avg_perf_vs_time}
    \vspace{-6pt}
\end{figure}

\subsection{One Long Session or Several Short Ones?}
\label{sec:sb2_sessions}
\label{sec:finding_resampling}
\label{sec:group_compute}

In code-as-policy, the LM's weights never change. The agent improves its program within a session and carries skills between tasks. At a fixed per-task budget, the Lineage margin over best-of-3 (BO3) therefore splits into two terms, one for sustained effort and one for inheritance:
\begin{align}
  \underbrace{\mathrm{Lineage} - \mathrm{BO3}}_{\text{margin}}
    = \underbrace{\bigl(\mathrm{Vanilla} - \mathrm{BO3}\bigr)}_{\text{one long session over three short}}
    + \underbrace{\bigl(\mathrm{Lineage} - \mathrm{Vanilla}\bigr)}_{\text{inherited library}} .
  \label{eq:decomposition}
\end{align}

\begin{figure}[t]
    \centering
    \includegraphics[width=\linewidth]{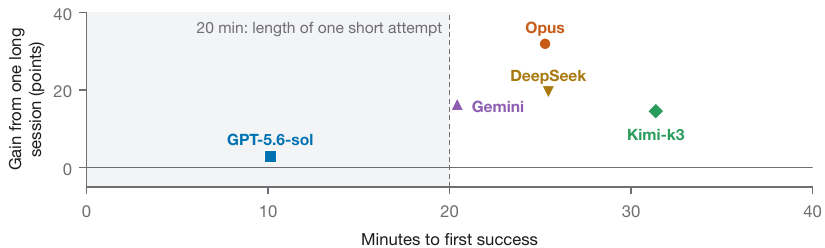}
    \vspace{-6pt}
    \caption{\textbf{One long session wins when a short attempt would end before the first success.} Each model's gain from one full-budget session (60 min) over the best of three short attempts (20 min) with the same total budget (Vanilla $-$ best-of-3, VAL success at the end of the budget), against its median time to first validation success without a library, both averaged over the five task families. Shading ends at 20 minutes, the length of one short attempt. RoboLab's budget is twice the others' (120 minutes), so its times are halved here; its actual times are twice those plotted. Per-family values are in \cref{tab:fig4_by_family}.}
    \label{fig:session_term}
    \vspace{-6pt}
\end{figure}

Best-of-3 is the fair version of trying more often: three fresh attempts at a third of the budget, keeping the one with the best validation score. On robosuite, the easiest family, one hour beats three 20-minute attempts for Opus (93.8 against 90.0\% TEST; \cref{tab:t1_main}) and, on validation, for every model but Gemini-3.6-flash, which is level (\cref{tab:fig4_by_family}). On RoboLab, the hardest, the sign of the session term depends on the model, and it follows one quantity (\cref{tab:decomposition}): when the model first succeeds. GPT-5.6-sol and Gemini-3.6-flash reach their first validation success at a median of 21 and 36 minutes, inside a 40-minute attempt, so each of the three attempts gets past the first success, and best-of-3 beats one two-hour session by 1.6 and 2.8 points. Opus and DeepSeek-v4-flash\footnote{DeepSeek-v4-flash's RoboLab runs with a library cover only the first 58 tasks: they stopped when the API provider retired the model. On LIBERO-RIG, DeepSeek is DeepSeek-v4.1-flash.} need about an hour, and Kimi-k3 nearly an hour and a half. A 40-minute attempt ends before their first success, and splitting costs them 5 to 24 points. Part of that cost is that short attempts often end without submitting a policy: Kimi-k3 submits a policy in 10\% of its 40-minute attempts against 68\% of its two-hour runs. Under LIBERO-RIG's hidden faults the session term is the largest in the study, 66 points for Opus, whose first success comes at a median of 28 minutes, after a 20-minute attempt has ended. Task difficulty alone therefore does not decide which setup is better. What decides it is the model's time to first success compared with the length of one short attempt. The same holds averaged over all five families (\cref{fig:session_term}): only GPT-5.6-sol reaches its first success inside a short attempt, and it gains 2.9 points from the long session, against 14.5 to 31.9 for the other four models.

\subsection{Experience Carries Forward, for Some Models}
\label{sec:sb2_test_time}
\label{sec:finding_inheritance}

\begin{figure}[t]
    \centering
    \includegraphics[width=\linewidth]{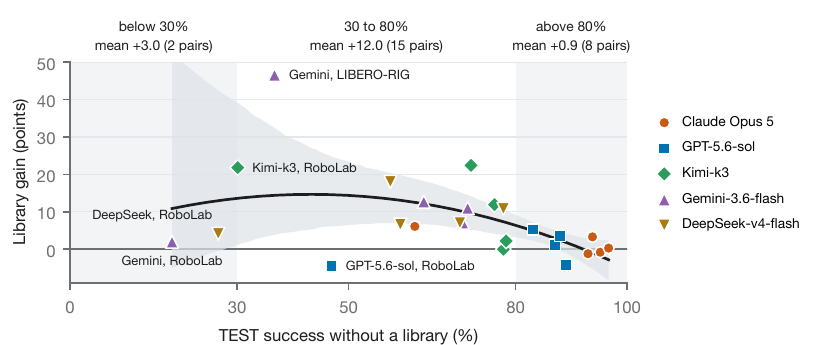}
    \vspace{-14pt}
    \caption{\textbf{The library helps most in the middle.} One point per model and family: TEST success without a library (Vanilla) against the library's gain (Lineage $-$ Vanilla), from \cref{tab:models}. Labels give the mean gain below 30\%, between 30 and 80\%, and above 80\%. Curve: least-squares quadratic fit over the 25 pairs, peaking at 43\% (gain $+14.7$); shading: 95\% bootstrap interval of the fit. }
    \label{fig:library_gain}
    \vspace{-6pt}
\end{figure}

\textbf{The library helps most in the middle.} How much the library adds depends less on which model inherits it than on how well that model does on a family without one (\cref{fig:library_gain}; \cref{fig:avg_perf_vs_time} gives each model's curves). Over the 25 model--family pairs of \cref{tab:models}, it adds 0.9 points on average where the model already scores above 80\% without it (Opus and GPT-5.6-sol on every family but RoboLab), 12.0 points between 30 and 80\%, and 3.0 below 30\% (Gemini-3.6-flash and DeepSeek-v4-flash on RoboLab). Above 80\%, there is little left to gain. Below 30\%, the model rarely reuses what it saved: DeepSeek-v4-flash imports library code in 20\% of its RoboLab policies, Gemini-3.6-flash calls most of its skills once, and Inkling-small, a sixth simulation model that we withdrew from the comparison, held 0 to 6 skills after 47 tasks (\cref{app:portraits}). Between 30 and 80\%, the gains are large: the library lifts Gemini-3.6-flash by 46.6 points under LIBERO-RIG's faults, and Kimi-k3 from 30.1 to 51.9\% on RoboLab, above GPT-5.6-sol with or without a library. Where the library does not change the final score, it still reaches that score sooner: on LIBERO-RIG, Opus and GPT-5.6-sol reach their final score by 30 minutes (96 and 89\% on validation), while their runs without a library are still at 75 and 81\%. These gains are larger than a rerun's noise: comparing each Lineage run with a held-out Vanilla repetition over 1,510 model--task pairs puts the library's help at 8 to 10 points, both on tasks the model never solved alone and on tasks it did (\cref{tab:library_null}). The one pair between 30 and 80\% where the library hurts, GPT-5.6-sol on RoboLab (46.9 to 42.4\%), shows that what the library holds matters as much as how well the model does without it.

\begin{figure}[t]
    \centering
    \includegraphics[width=\linewidth]{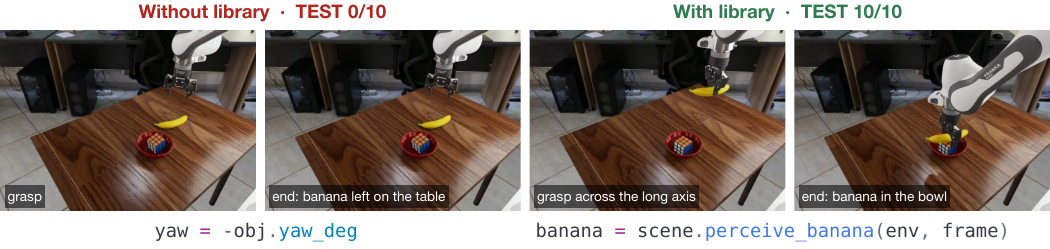}
    \vspace{-14pt}
    \caption{\textbf{What an inherited skill changes.} Kimi-k3 on the RoboLab task ``put the Rubik's cube and the banana in the bowl'', same scene. Without a library (left), the policy copies the gripper's yaw from the object pose, the fingers land on the banana, and all ten TEST states fail. With the library (right), it calls an inherited skill, written on an earlier banana task, that finds the banana's ridge and long axis and pinches across it; all ten TEST states succeed.}
    \label{fig:sb2_banana}
    \vspace{-6pt}
\end{figure}

\textbf{What the library holds matters.} \Cref{fig:sb2_banana} shows a useful inherited skill: one skill, written on an earlier task, turns a failed grasp into a reliable one. Kimi-k3's library fills with such routines (gripper control from measured opening, yaw alignment, pinch grasps), and its gain grows as they accumulate. GPT-5.6-sol's library holds mostly object-location recipes and few grasp checks. In inspected failures, it reuses thresholds and offsets from earlier tasks and keeps tuning them after they fail on a new scene. To separate the library's content from the model, we give GPT-5.6-sol libraries written by other models. On RoboLab, GPT-5.6-sol scores 46.9\% without a library and 42.4\% with its own, but 60.6\% with the library Opus wrote and 55.7\% with Kimi-k3's (three runs over all 105 tasks for each donor library). The model is the same, and only the inherited library differs. Experience saved as code and notes can transfer between models.

\subsection{What Carries Forward Includes Mistakes}
\label{sec:sb2_mistakes}
\label{sec:finding_variance}
\label{sec:group_limits}

Nothing in the loop requires a saved lesson to be checked again, so errors persist like skills do. In self-exploration, where the agent has no judge and must build its own success detector, one Opus run built a placement verifier in its first session, calibrated on sixteen episodes whose three failures happened not to move at all. The verifier declared success on 699 later rollouts and was wrong on 91\% of them. Because its verdict was also the stopping rule, 637 rollouts ended without another grasp attempt. The run saved the verdict in memory as a fact, and every later session inherited it. The run scored 4.6\% overall. Its two sibling runs, identical in setup but without that verifier, scored 15.5 and 23.6\%, three to five times higher. Runs with the same setup often end far apart: two DeepSeek-v4-flash Lineage runs differed by 20.9 points with two tasks flipping between 0 and 100, and on LIBERO the library moves Opus and GPT-5.6-sol by less than a run's standard deviation (94.4 against 95.2\% and 88.3 against 87.1\%). A single run cannot be told apart from luck, so \modelname{} reports the spread over three runs.

\section{Does What It Learns Transfer?}
\label{sec:sb2_transfer}
\label{sec:group_reach}

\begin{finding}
Success falls as the test moves further from what the agent practiced. The parts of the library that still help are the general ones: control code, grasp routines, and shared parsing, not solutions written for one task. Code policies cope with hardware faults they can observe, but fail when the camera moves. A VLA trained inside the loop loses some success when the camera moves instead of failing outright, but it matches code only on families where its code teacher produces many successful demonstrations to train on. On a real robot, success in practice overstates success in the test: during practice the agent resets the scene itself and rarely creates the hard starting states that an operator sets up for the test.
\end{finding}

\subsection{Down the Generalization Tiers}
\label{sec:sb2_tiers}
\label{sec:finding_perturbation}

\begin{table}[t]
\centering
\caption{\textbf{Success falls with distance from practice.} Mean TEST success (\%) of Claude Opus~5 policies per tier, over the practice--test family pairs where both the code policy and the VLA have results (\cref{tab:tier_cells} and \cref{fig:opus_tier_table} give every pair). T3 and T4 policies are written after self-exploration on the new tasks; T1 and T2 policies are frozen. T5 is not a transfer result like T1--T4: the agent starts from an empty library and practices each of the five hardware tasks on the physical robot, so the T5 entry is its score on new initial states of those same tasks (\cref{tab:model_scores}), with no VLA run on hardware.}
\label{tab:sb2_tier_means}
\small
\begin{tabular}{llcc}
\toprule
Tier & Graded on & Code-as-policy & VLA (AR) \\
\midrule
T1 & New initial states of the practiced tasks & 87.8 & 73.2 \\
T2 & The same tasks under changed scenes or hardware faults & 59.7 & 55.3 \\
T3 & New tasks in the same simulator & 33.3 & 43.1 \\
T4 & New tasks in a new simulator and action API & 28.7 & 10.2 \\
\midrule
T5 & New initial states of tasks practiced on the physical robot & 35.6 & --- \\
\bottomrule
\end{tabular}
\end{table}

\Cref{tab:sb2_tier_means} follows Opus's policies away from where they were practiced. Both policy forms lose ground at every simulation tier. The steps are uneven: the same code policy, practiced on LIBERO, scores 94.4\% on LIBERO's TEST states, 66.0\% under LIBERO-RIG's hardware faults, and 47.7\% under LIBERO-PRO's scene changes. A scene change costs more than a hardware fault, because a code policy re-detects the object before each move and so corrects a fault as it goes, as long as the camera stays where it was (\cref{sec:sb2_code_weights}).

\subsection{One Program, New Scenes}
\label{sec:sb2_new_scenes}
\label{sec:finding_transfer}
\label{sec:finding_unseen}

\textbf{The libraries are reused across tasks and simulators.} On the first half of RoboLab, Opus's three lineages imported 202 distinct skills 660 times, and skills written on the MuJoCo families were imported by about half of its RoboLab policies (two fifths for Kimi-k3, almost none for the other three models). The skills reused across simulators are control code, such as servo wrappers and grasp routines, not task solutions. One rim-pinch skill, written on LIBERO because a bowl was wider than the gripper, was imported by 66 later policies; RoboLab's four spatial-relation tasks were solved 10 of 10 on first contact, in all three lineages, by libraries that already held LIBERO's spatial skills and a memory that relation words name regions of the table (\cref{app:library_contents}).

\begin{figure}[t]
    \centering
    \includegraphics[width=\linewidth]{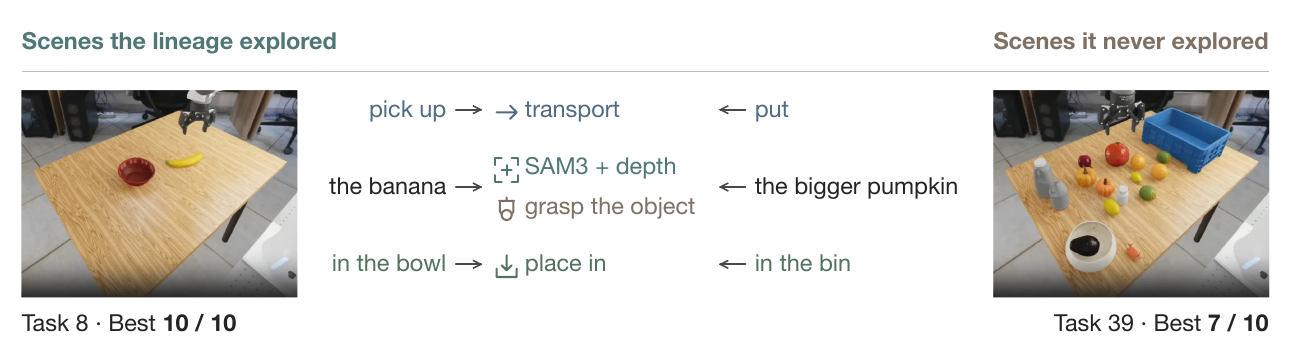}
    \vspace{-14pt}
    \caption{\textbf{One task-general program serves an explored and a never-explored scene.} A self-exploration lineage, never told its tasks, ends with one program that parses each instruction into a goal and calls shared routines. ``Put the bigger pumpkin in the bin'', in a scene the lineage never explored, calls the same grasp and placement as an explored banana-into-bowl task, and rises from 0 to 7 of 10 as those routines improve elsewhere.}
    \label{fig:unseen_parse}
    \vspace{-6pt}
\end{figure}

\textbf{A program written without knowing the task works on new scenes when the routines it needs already work.} In self-exploration, the agent gets neither task instructions nor a judge. It proposes its own practice and learns to map instructions onto common goals and reusable routines (\cref{fig:unseen_parse}). We froze one Opus lineage's task-general program at four points along its RoboLab sequence and graded each, unchanged, on one task per scene (\cref{fig:unseen_probe}). Success on the same 44 tasks rises from 15.5 to 31.6\%, and on twelve scenes the lineage never explored, from 5.8 to 22.5\%. The gain follows the routines: at the last checkpoint, never-explored tasks that need only routines that already work somewhere average 52\%, and those that need a routine never shown to work, such as removing an object from a container, average 1\%. Improvement concentrates in grasping: grasp failures fall from 58 to 29\% of graded episodes. When the instruction is finally disclosed, one rewrite adds a consistent margin over the frozen program: it is better in all 21 lineages of each of Opus, GPT-5.6-sol, and Kimi-k3 (median gains 7.8, 10.4, and 7.1 points; \cref{tab:selfexp_models}).

\subsection{Code and Weights Break Differently}
\label{sec:sb2_code_weights}
\label{sec:finding_code_vs_weights}

\begin{figure}[t]
    \centering
    \includegraphics[width=\linewidth]{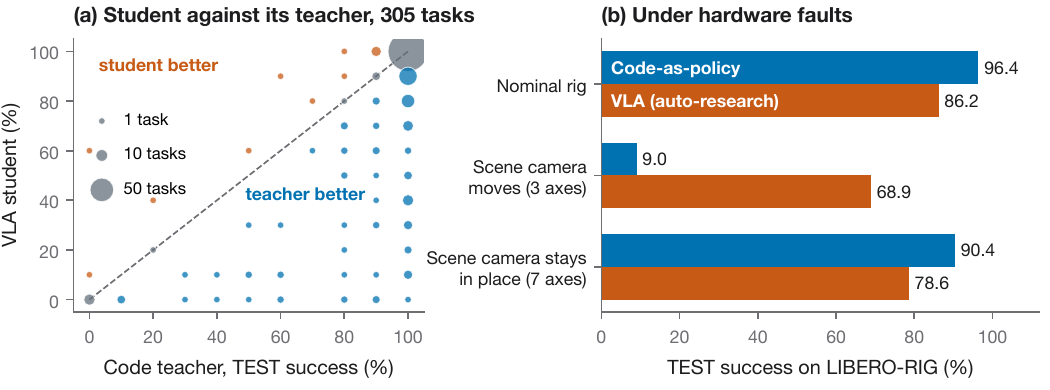}
    \vspace{-14pt}
    \caption{\textbf{Code and weights, side by side.} (a) The TEST success of the VLA (the student) against the code policy (its teacher), for the VLA Lineage sessions of an Opus auto-research agent, pooled over the 305 tasks of LIBERO, LIBERO-PRO, and RoboLab; marker area counts tasks. The student beats its teacher on 15 tasks (orange), trails it on 126 (blue), and ties on 164 (grey). (b) LIBERO-practiced Opus policies re-graded on LIBERO-RIG, averaged over the fault axes in each group (per-axis values in \cref{tab:vla_rig}). }
    \label{fig:sb2_teacher_student}
    \vspace{-6pt}
\end{figure}

In auto-research, an Opus agent trains one $\pi_{0.5}$ checkpoint per suite of tasks from the rollouts of per-task code teachers, within the same development time a code agent gets. The agent decides what data to collect, how to train, and whether to fix a teacher.

\begin{figure}[t]
    \centering
    \includegraphics[width=\linewidth]{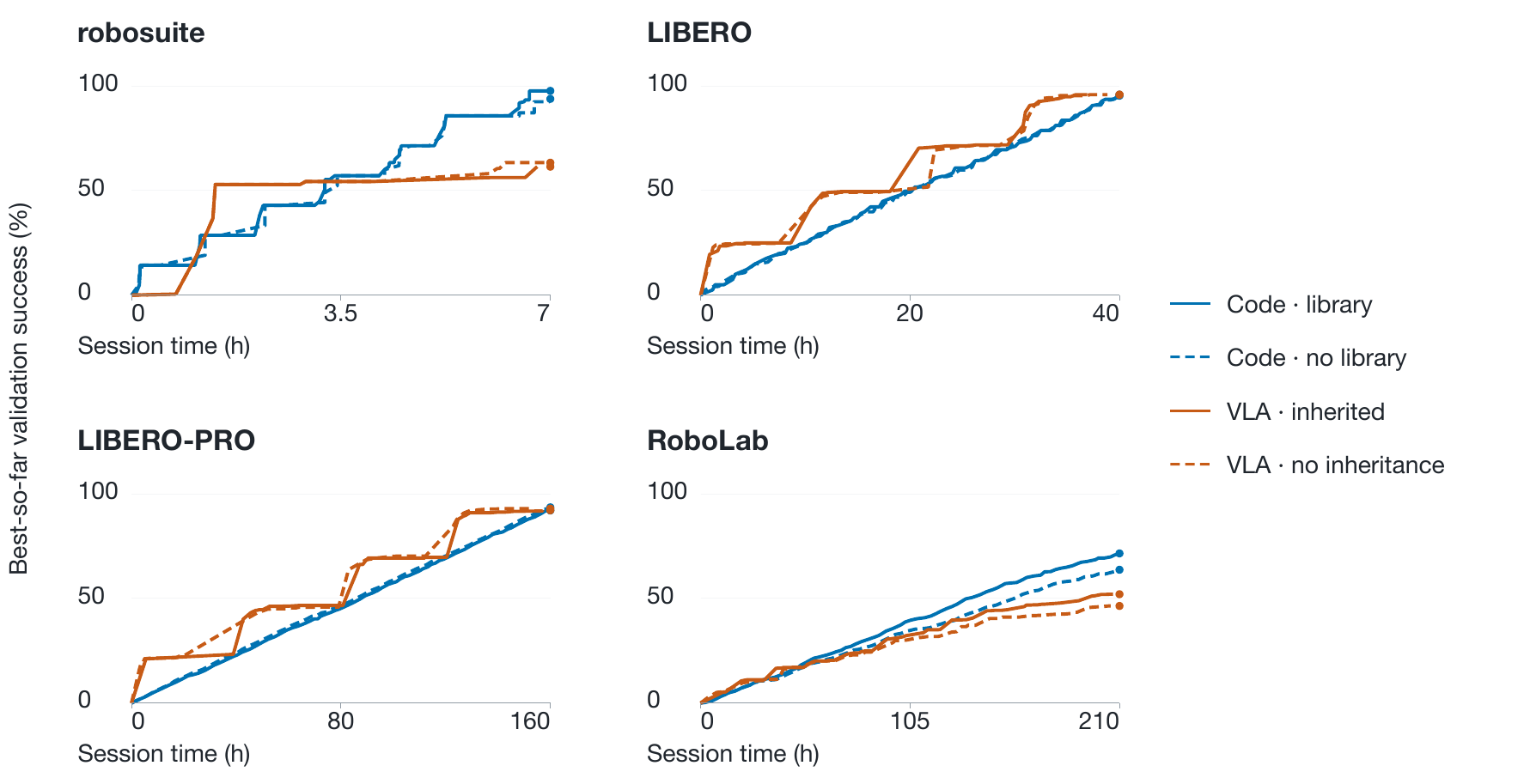}
    \vspace{-6pt}
    \caption{\textbf{Code against weights over accumulated session time.} Code (blue) against weights (orange) over accumulated session time, with (solid) and without (dashed) inherited state: best-so-far per task validation success averaged over a family's tasks, sessions laid end to end in curriculum order. 
    Session time includes GPU training for auto-research VLA.}
    \label{fig:cap_vs_vla}
    \vspace{-8pt}
\end{figure}

\textbf{Weights match code where demonstrations are plentiful.} On LIBERO and LIBERO-PRO the VLA scores within two points of code (94.0 vs.\ 94.4 and 90.6 vs.\ 91.8\%) and matches or beats it on about four tasks in five. It falls well short on robosuite (61.0 vs.\ 97.1) and RoboLab (47.2 vs.\ 68.0; \cref{tab:vla_vs_cap}). The limit is the number of demonstrations: RoboLab's slow simulator yields about one seventh as many teacher episodes per agent-hour as LIBERO. RoboLab tasks with fewer than 50 successful demonstrations score 20.8\% against 94.0\% with 300 or more (\cref{tab:vla_dose}). The learning curves show the same split over session time (\cref{fig:cap_vs_vla}). The student rarely exceeds its teacher (\cref{fig:sb2_teacher_student}a): of the 11 tasks whose teacher solves no TEST state, the VLA succeeds on at least one TEST state in 2 of them.

\textbf{Weights survive a moved camera; code does not.} A code policy turns pixels into world coordinates through an assumed camera pose. When the scene camera moves, nothing in the program checks that pose, and the LM that wrote the program is not running at test time to notice. Code falls to 9\%, while the VLA keeps 69\% (\cref{fig:sb2_teacher_student}b). Under the seven faults that leave the scene camera in place, the code policy re-detects the object before each move and holds 90\%. The VLA also handles new instructions better: trained on its own self-exploration data, where 95\% of episodes carry a distinct instruction, it reaches 42.6 and 43.6\% on held-out LIBERO and LIBERO-PRO tasks, level with or above the code policy written after the task instruction is disclosed (41.7 and 24.9\%).

\textbf{The two kinds of agent attend to different things.} Code agents open rendered images in 84 to 96\% of sessions; only 10 of 40 inspected VLA sessions ever look at a rollout of the policy they train. Two thirds of code agents' memories concern grasp and contact, while over a third of VLA agents' memories concern the training recipe (\cref{tab:memory_topics}). Code agents mostly debug how the robot interacts with objects. VLA agents mostly debug the training pipeline.

\subsection{On a Real Robot, Practice Is Not a Test}
\label{sec:sb2_hardware}
\label{sec:group_hardware}
\label{sec:finding_real_world}

\begin{table}[t]
\centering
\caption{\textbf{Held-out scores on the five physical tasks} (\%). Mean outcome (success 1, partial 0.5, failure 0) over ten operator-staged scenes per code-as-policy (CaP) task and five per agent-as-policy (AaP) task; Block Stacking reports the mean highest tower, capped at ten cubes, as a percentage. Each entry is one practice run per agent. Practice counts and costs are in \cref{tab:real_world,tab:real_world_aap}.}
\label{tab:model_scores}
\small
\begin{tabular}{l*{5}{c}}
\toprule
& \multicolumn{3}{c}{CaP: frozen code policy} & \multicolumn{2}{c}{AaP: agent acts live} \\
\cmidrule(lr){2-4}\cmidrule(lr){5-6}
Task & Fable 5.1 & GPT-6-astra & Opus 5 & Fable 5.1 & GPT-6-astra \\
\midrule
Banana into Bowl & 70.0  & 50.0 & 80.0 & 100.0 & 80.0  \\
Push-T           & 100.0 & 60.0 & 40.0 & 100.0 & 100.0 \\
Cup Upright      & 70.0  & 50.0 & 40.0 & 20.0  & 100.0 \\
Block Stacking   & 71.0  & 17.0 & 18.0 & 52.0  & 60.0  \\
Spell SEAR       & 40.0  & 30.0 & 0.0  & 60.0  & 100.0 \\
\midrule
Overall          & 70.2  & 41.4 & 35.6 & 66.4  & 88.0  \\
\bottomrule
\end{tabular}
\end{table}

On hardware the agent practices each task for up to 12 hours, starting from an empty library, and must also learn to reset the scene. \Cref{tab:model_scores} gives the frozen policies' scores on staged scenes: Fable~5.1 leads at 70.2\%, ahead of GPT-6-astra at 41.4 and Opus at 35.6.

\begin{figure}[t]
    \centering
    \includegraphics[width=\linewidth]{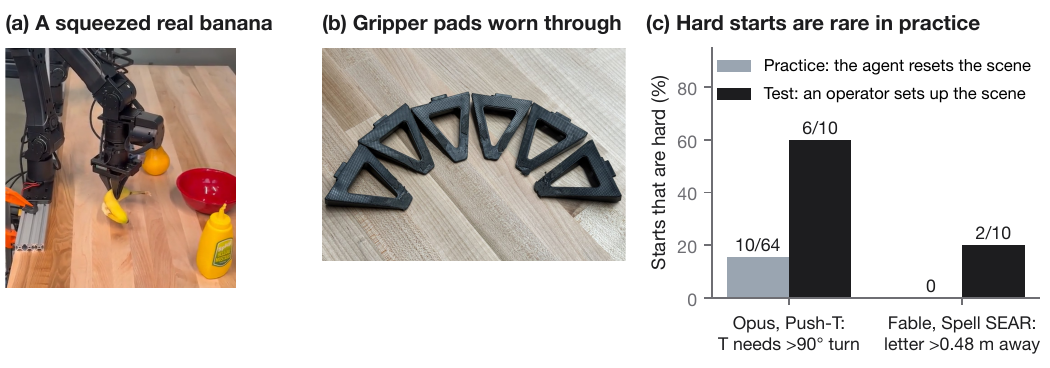}
    \vspace{-14pt}
    \caption{\textbf{What hardware exposes that simulation misses.} (a) A grasp that squeezes a real banana. (b) Gripper pads worn through during practice. (c) How often a task starts in a hard state: in practice, where the agent resets the scene itself, and in the test, where an operator sets it up (counts from the recorded start states).}
    \label{fig:sb2_hw_failures}
    \vspace{-6pt}
\end{figure}

\textbf{Ten practice successes in a row do not guarantee a high held-out score.} Six practice runs reached ten successes in a row, which ends practice. only one of them, Fable's Push-T, then succeeded on all ten staged scenes, and two (Fable's Spell SEAR, Opus's Cup Upright) succeeded on only four. The reason is that the agent's own resets choose the states it practices on (\cref{fig:sb2_hw_failures}c). Only 10 of Opus's 64 practice Push-T starts required turning the T by more than $90^\circ$, against 6 of the 10 staged scenes. Fable's reset, which scrambles the letters, never left a letter farther than 0.48\,m from an arm, but 2 staged Spell SEAR scenes did. A streak measured on states of the agent's own making can end practice before the hard states ever appear.

\textbf{Frozen policies fail at limits the agents set themselves.}\label{sec:finding_real_world_limits} We audited all 116 failed or partial evaluation outcomes; no scene was physically impossible, since each was solved by another agent, by the same agent in the other mode, or lay within measured reach. In 29 of 34 for Fable, 32 of 35 for Opus, and 26 of 47 for GPT-6-astra, the policy stopped at a limit, rule, or motion the agent had chosen itself, typically after a single incident in practice. Fable, for example, set a reach limit of 0.49\,m with a vertical gripper, while tilting the gripper by $15^\circ$ reaches 0.56\,m. The two staged scenes beyond its limit fail (\cref{app:real_world_failures}). A lesson learned once becomes a rule, and nothing in the loop tests it again.

\begin{figure}[t]
    \centering
    \includegraphics[width=\linewidth]{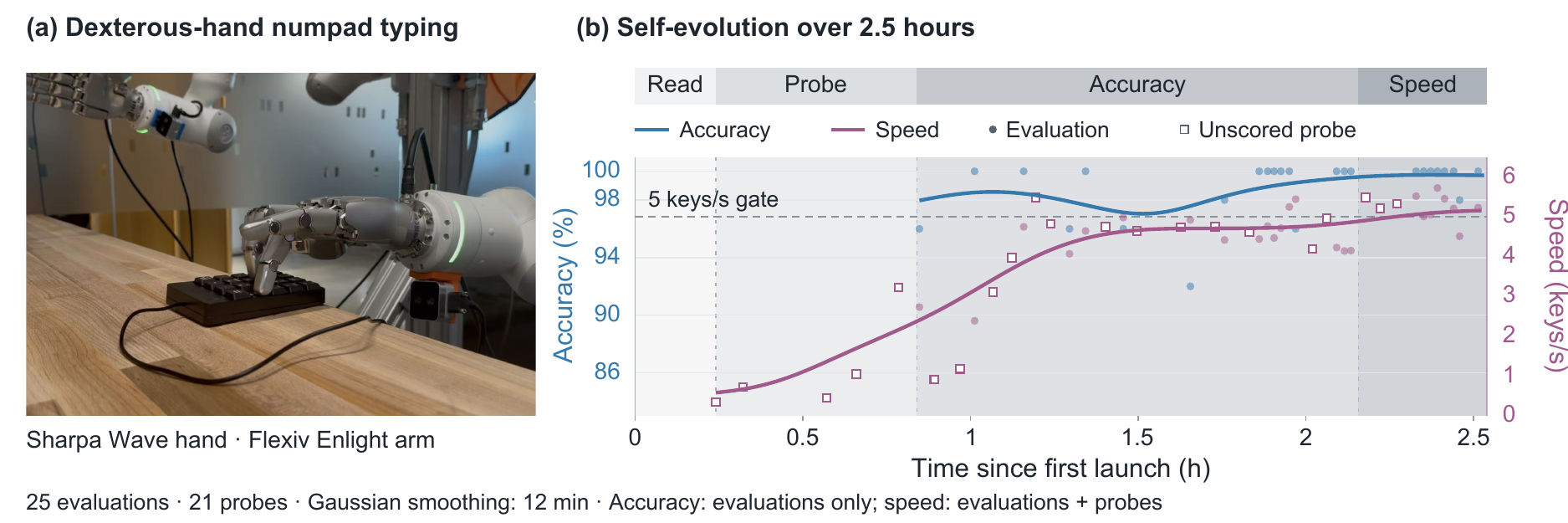}
    \caption{\textbf{Dexterous-hand numpad typing.} (a)~The Sharpa Wave hand on a Flexiv Enlight arm at a physical numpad. (b)~Accuracy (blue, left axis) and speed (purple, right axis) over 2.5~hours from first launch, including a restart. Dots are scored evaluations, hollow squares the speeds of unscored attempts; curves are Gaussian-smoothed with a 12-minute bandwidth. Shading marks phases annotated from the agent's journals; the dashed line is the 5~keys/s speed gate. Completing the task also requires at least 90\% accuracy over a sliding window of ten evaluations.}
    \label{fig:dex_numpad}
    \vspace{-8pt}
\end{figure}

\textbf{The loop runs on a second embodiment.}\label{sec:finding_dexterous} We ported the harness to a Sharpa Wave hand with 22 joints on a Flexiv arm, sharing no hardware with the two-arm rig. A Fable~5.1 agent learned to type random digits on a physical numpad at over five keys per second within 2.5 hours, and typed the first 100 digits of $\pi$ three times at 5.1 keys per second with one wrong digit in total across the three runs (\cref{fig:dex_numpad}); the learned typing policy also plays a short melody on a piano keyboard (\href{https://sear.bot}{project website}). The agent learned on this hand alone, on hardware only. An accurate simulated twin of the hand later proved hard to build. The demonstration therefore shows that the harness can be moved to a new robot, not that skills transfer between robots.

\section{What the Language Model Brings}
\label{sec:sb2_llm}

\begin{finding}
The language model sets how well the robot does: the strongest model leads on every family, and the gap between models widens as tasks get harder. What separates models is not whether they can see. A text-only model that reaches the scene through segmentation and depth tools can beat a multimodal model. Images help mainly with debugging. Models also have preferred ways to act, and the best mode for one model is not the best for another.
\end{finding}

\subsection{Better LM, Better Robot}
\label{sec:sb2_better_llm}
\label{sec:finding_models}
\label{sec:group_gate}

\begin{figure}[t]
    \centering
    \includegraphics[width=\linewidth]{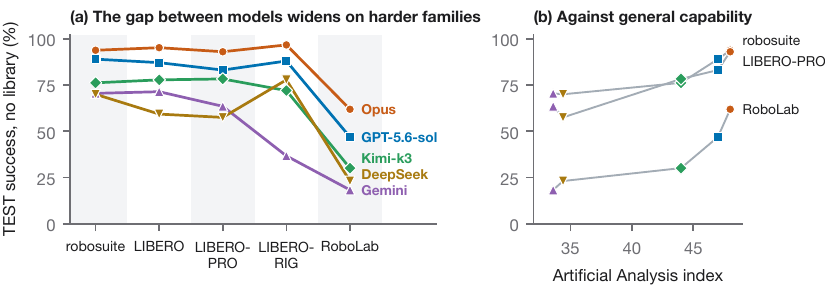}
    \vspace{-14pt}
    \caption{\textbf{Model scores across task families, without a library.} (a) TEST success, one line per model, families from easiest to hardest. (b) The same scores against the Artificial Analysis Intelligence Index~\citep{aaindex} for robosuite, LIBERO-PRO, and RoboLab; colors as in (a). Scores with a library are in \cref{tab:models}, and the library's effect is in \cref{fig:library_gain}. }
    \label{fig:model_scores}
    \vspace{-6pt}
\end{figure}

\Cref{fig:model_scores}a compares the five simulation models without a library. Opus has the highest score on every family, with or without a library (\cref{tab:models}). GPT-5.6-sol is second on every family without a library. With a library, Kimi-k3 overtakes it on robosuite, LIBERO-RIG, and RoboLab. The ranking follows general capability. On the Artificial Analysis Intelligence Index~\citep{aaindex} (high reasoning setting; \cref{fig:model_scores}b), Opus scores 48, GPT-5.6-sol 47, Kimi-k3 44, and Gemini-3.6-flash and DeepSeek-v4-flash 34 each. Without a library, the robot ranking follows this order on robosuite, LIBERO, LIBERO-PRO, and RoboLab. On LIBERO-RIG, where DeepSeek-v4.1-flash replaces DeepSeek-v4-flash, DeepSeek places third. Small differences on the index become large differences on the robot. Opus and GPT-5.6-sol are one index point apart, but 4.8 points of success apart on robosuite and 15.0 apart on RoboLab. The spread also grows with difficulty: without a library, the best and worst models are 23.8 points apart on robosuite and 43.6 on RoboLab. Under LIBERO-RIG's hidden faults, the gap is 60.0 points. Below some level of capability, the loop has nothing to work with. One Inkling-small run went 29 tasks in a row without a valid policy, so it had nothing to save in its library, and we withdrew the model (\cref{app:portraits}). The library pays off only above this capability floor (\cref{sec:sb2_test_time}). With three runs per model, the rankings are not firm. On the easier families, the spread between repeated runs often exceeds the gap between neighboring models (\cref{sec:sb2_mistakes}).

\subsection{Seeing Is Not the Bottleneck}
\label{sec:sb2_no_pixels}
\label{sec:finding_vision}
\label{sec:finding_failures}

A natural guess is that manipulation rewards the model that can see. Our results do not support this guess. DeepSeek-v4-flash is text-only, yet it combines the tools in its toolbox into working manipulation policies: nearly all of its RoboLab policies ask SAM3 for a mask of the named object and use the depth image to turn that mask into a 3-D position, and it produces a policy in 85\% of runs. On RoboLab, the hardest family, it beats the multimodal Gemini-3.6-flash, which has the same index score (34). It wins without a library (23.1 against 18.3\%) and with one (30.7 against 22.6\% on the 58 tasks both completed), though not in best-of-3. DeepSeek-v4-flash trails Gemini-3.6-flash on LIBERO and LIBERO-PRO (by 12.1 and 6.0 points without a library). When SAM3 misses an object, DeepSeek-v4-flash has no fallback. Among the multimodal models, GPT-5.6-sol never opens images directly and Gemini-3.6-flash opens one in 2\% of runs.

Images are used mainly for diagnosis. Opus views six to seven images per run. After viewing an image, it edits its code 38\% of the time, against a 24\% baseline, and Kimi-k3 edits after 68\% of its views. The failures these models diagnose are physical, not perceptual. Over 21,552 graded RoboLab episodes, the target's distance, rotation, and clutter barely predict success (correlation 0.04 in magnitude). What fails is the grasp. In episodes where the policy logged how wide the gripper was when it closed, the gripper had closed on air in 78\% of failed episodes and in 18\% of successful ones. Tasks that need 3-D relations, such as stacking, are among the easiest. Tasks that require grasping rims, handles, or thin shafts are the hardest. Few policies rely on a learned action model: only 2\% of no-library RoboLab policies call the pretrained $\pi_{0.5}$ in the toolbox. GPT-5.6-sol's policies that used it scored 16\%, against 54\% for those that did not.

\subsection{Models Have Preferences}
\label{sec:sb2_preferences}
\label{sec:finding_real_world_aap}

On hardware we also ran agent-as-policy: the LM stays in the loop at test time and re-observes the scene before every command (\cref{tab:model_scores}). We gave two models the same instructions, which ask for one command at a time but allow the model to write its own tools. The two models act in opposite ways. GPT-6-astra issues every command through a separate model call and requests about one image per arm motion. Acting live, it scores 88.0\%, against 41.4\% as a frozen code policy, and 95\% against 48\% on the four non-stacking tasks. It uses a fresh image to notice and correct a mistake in at least 9 of its 25 evaluation episodes. Fable~5.1 instead scripts the robot: its tool code grows from 202 to 3,576 lines over the five tasks, those scripts issue 99\% of its commands, and it looks at one image per seven motions. In effect, Fable~5.1 has turned live execution back into code-as-policy, and it scores about the same in both modes: 70\% on the four non-stacking tasks either way. Overall it scores 66.4\% live against 70.2\% frozen, a gap that comes from Block Stacking, whose live runs used a different stopping rule. Each model scores highest in the mode it prefers, though for Fable~5.1 the margin is small. Acting live is slower: a live episode takes a median of 12.8 minutes of robot time against 3.8 for a frozen code policy (\cref{tab:aap_latency}). These comparisons rest on one practice run per model and five staged scenes per task.

\section{Related Work}
\label{sec:related}

\textbf{Self-improving LM agents.} An LM agent can improve without weight updates by writing reflections on its failures into memory, as in Reflexion~\citep{shinn2023reflexion}, by growing a library of code skills, as Voyager does in Minecraft~\citep{wang2023voyager}, or by rewriting its own improvement procedure~\citep{zelikman2024stop} and harness~\citep{lin2026agentic, weng2026harness}. \modelname{}'s library combines the first two. \citet{lin2026disentangling} show that much apparent self-evolution in LM agents is lucky resampling. Our best-of-3 and no-library controls test for the same effect in robot learning, where feedback comes from physical execution. Benchmarks for research agents~\citep{chan2024mlebench, wijk2025rebench, starace2025paperbench, rank2026posttrainbench} share our emphasis on fixed budgets and isolated grading, and our sealed judge adapts EdgeBench's separate work and judge containers~\citep{edgebench}.

\textbf{LMs that write robot behavior.} Code as Policies~\citep{liang2023code} and ProgPrompt~\citep{singh2023progprompt} have an LM write executable robot programs from instructions; Eureka~\citep{ma2024eureka} and RoboGen~\citep{wang2024robogen} have it write the machinery that produces a policy, such as reward code or task scenes. Auto-research in \modelname{} belongs to the second line: the agent builds the data and training pipeline for a VLA~\citep{black2025pi05}.

\textbf{Self-evolving robot agents.} CaP-X~\citep{capx} benchmarks and improves coding agents for manipulation; ASPIRE~\citep{aspire} and RATs~\citep{rats} grow skill libraries through practice or play and reuse them on new tasks; ENPIRE~\citep{enpire} lets an agent improve its own robot policy on real hardware; Waddle~\citep{waddle2026} has agents control physical robots and share a skill library. A further wave appeared while this paper was being written. RPG~\citep{rpg2026}, RoboRSI~\citep{roborsi2026}, and Embodied Turing Machines~\citep{etm2026} evolve a shared skill or code library through practice and validate changes before keeping them; \citet{kuroki2026transfer} reuse an archive of agent-optimized programs on new tasks; \citet{wei2026harness} evolve the harness of an agent that acts in the loop and find that keeping only revisions that beat the current harness matters. Closer to our auto-research mode, SimEX~\citep{simex2026} has a coding agent build and correct a robot toolbox and its simulator, and Robo-COP~\citep{robocop2026} lets an orchestrator fine-tune its VLA on demonstrations it curates on RoboLab tasks. Each defines its own protocol (\cref{tab:sb2_landscape}), so their results cannot be compared, and none reports both a budget-matched restart control (our best-of-3) and a no-library control. \modelname{} provides one shared protocol that includes both controls, and runs code-as-policy, agent-as-policy, and auto-research under it. Three findings elsewhere echo ours. An agent that decides after each tool call succeeds more often than a program written in advance (56 against 24\%;~\citealp{urai2026}; see also \citealp{racap2026,pyrualean2026}). An LM orchestrator that plans, checks outcomes, and recovers around frozen VLAs and skills lifts LIBERO-PRO success from 12.8 to 53.3\% with no training, a gap its authors call the orchestration gap~\citep{pigey2026}. And 123 rounds of agentic skill discovery can fail to solve a single target task~\citep{wang2026whatstops}. Along the same lines, RoboHarness~\citep{roboharness2026} gives a coding agent tracked image keypoints and a few motion primitives, with no training, and reports beating VLA and world-action models on BEHAVIOR household tasks (at twice the challenge's step budget) and lifting between 68 and 86 of 100 unseen objects, depending on the LM. LIBERO~\citep{liu2023libero} was itself built for lifelong learning. \modelname{} asks what a library is worth once the agent, rather than a training loop, decides what to keep. Interface standards such as XPolicyLab~\citep{xpolicylab2026} are complementary.

\textbf{Benchmarks of LM agents that use robots.} RobotWorld~\citep{robotworld2026} evaluates frontier agents that act through robot tool calls on 84 simulated tasks across manipulation, locomotion, driving, and aerial control, one episode per task, and finds that models differ by the kind of goal they solve, as our models differ by how they prefer to act (\cref{sec:sb2_preferences}). RLE-Bench~\citep{rlebench2026} tests coding agents as robot learning engineers across control, policy learning, perception, and mechanical design. EmbodiedSWE~\citep{embodiedswe2026} benchmarks coding agents on long-horizon dexterous tasks and turns their verified solutions into VLA training data, noting that the solutions tend to be specialized to single task instances. These benchmarks measure what an agent can do in one attempt. \modelname{} measures what the agent keeps from practice and how far from practice that knowledge still works.

\section{Discussion}
\label{sec:discussion}

\begin{figure}[!htb]
    \centering
    \includegraphics[width=0.8\linewidth]{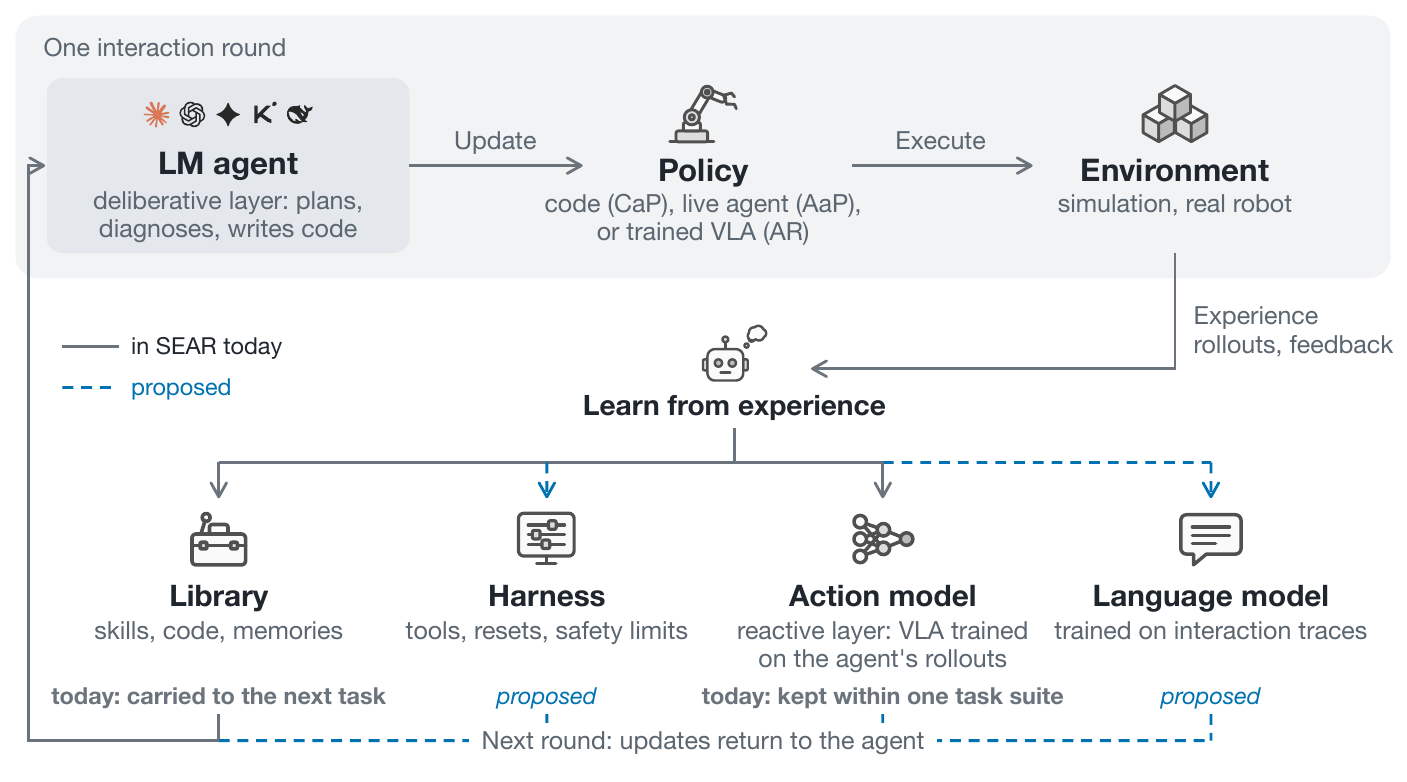}
    \vspace{-6pt}
    \caption{\textbf{The proposed co-evolving loop.} Experience from each interaction round can update four components that return to the agent. Solid grey: what \modelname{} does today (the library carries over between tasks; a VLA is trained but stays with its task suite). Dashed blue: proposed (harness revision and training the LM on its own traces).}
    \label{fig:sear_loop}
    \vspace{-8pt}
\end{figure}

\textbf{A continual learning loop.}\label{sec:sb2_loop}\label{sec:discussion_coevolution} Today a \modelname{} agent carries one thing from task to task: its library of code skills, tools, and written memories. A trained action model stays with the suite that produced it. The natural next step is a loop in which everything the agent touches improves together. Every round of practice produces rollouts and feedback, and that experience can flow into four places: the library, as reusable code and memories; the harness, meaning the sensors, tools, reset procedures, and safety limits the agent works through; the action model, fine-tuned on the agent's own rollouts with failures filtered out or turned into recovery data; and eventually the LM itself, trained on its own interaction traces. Whatever improves is handed back to the agent for the next round. Seen this way, the persistent state of the whole system plays the role that weights play elsewhere, except that much of it is readable, editable knowledge. Our results point at one architecture for that loop: the LM as a deliberative layer that plans, diagnoses, and writes code, and a VLA as a reactive layer for continuous control, trained on what the LM delegates to it. Code policies break when the camera moves, but the VLA does not. The VLA, in turn, is limited by the demonstrations that code supplies (\cref{sec:sb2_code_weights}). Each layer therefore covers the other's failure. \Cref{fig:sear_loop} sketches the loop, and \cref{app:sb2_coevo} describes what evaluating it would require. \modelname{} is built to test whether such a loop changes the shape of the improvement curve, not only its height.

\textbf{What the controls bought.}\label{sec:sb2_controls_bought} Each finding depends on a control a single learning curve would not have. Without best-of-3, we could not tell when one long session beats three restarts. Without the no-library run, the library would seem to help every model. Without repeated runs, the run that collapsed after saving a false verdict (\cref{sec:sb2_mistakes}) would read as a model difference. Without the sealed judge and its audit, the shortcuts of \cref{app:shortcuts} would count as skill. Without the tiers, a gain on held-out states would pass for transfer. \Cref{app:design_rules} turns these findings into design rules for anyone building such a system, and \cref{app:limitations} lists the limitations of this study.

\section{Conclusion}
\label{sec:conclusion}

When an agent improves a robot from its own experience, a rising success rate is not enough evidence that it is learning. \modelname{} asks two questions: is the agent learning or just trying more often, and does what it learns help on tasks other than the one it practiced? It answers both with matched controls, a sealed judge, and five tiers of distance from practice. Across seven frontier LMs, the answers are mixed in ways a single learning curve would hide. Long sessions pay off only for models that are slow to reach a first success. Libraries help most for models that succeed sometimes but not reliably, and they can also hurt. Control code and shared routines transfer, while task-specific solutions do not, and a success streak in practice does not guarantee success on scenes someone else sets up. Finally, the choice of LM, and how that LM prefers to act, decide much of what the robot can do.

\newpage
\appendix
\section{Results in Detail}
\label{app:details}

This appendix holds the evidence behind the main text that did not fit there, in the order of \cref{sec:sb2_learning,sec:sb2_transfer,sec:sb2_llm}.

\subsection{Session, Library, and Model Scores as Numbers}
\label{app:main_tables}

\begin{table}[t]
\centering
\caption{\textbf{Session and library terms} of \cref{eq:decomposition}, the values behind \cref{fig:library_gain}; TEST success (\%); BO3 is best-of-3. RoboLab: one 120-minute session or three 40-minute attempts; LIBERO-RIG: 60 minutes or three of 20. First success: median time of the first validation success over the Vanilla runs that ever succeed. LIBERO-RIG best-of-3 is read over tasks with a graded winner (all 18; 9 for Gemini-3.6-flash). $^{\dagger}$Over the first 58 RoboLab tasks, where DeepSeek-v4-flash's runs stopped when the model was retired (Lineage 30.7, Vanilla 26.6 on those tasks). $^{\S}$Its successor, DeepSeek-v4.1-flash.}
\label{tab:decomposition}
\footnotesize
\setlength{\tabcolsep}{3pt}
\begin{tabular}{llccccc}
\toprule
& & First success & & & Session term & Library term \\
Family & Model & (median min) & Best-of-3 & Vanilla & Vanilla $-$ BO3 & Lineage $-$ Vanilla \\
\midrule
RoboLab & Claude Opus 5     &  58 & 57.1 & 61.9 & $+4.8$  & $+6.1$ \\
        & GPT-5.6-sol       &  21 & 48.5 & 46.9 & $-1.6$  & $-4.5$ \\
        & Kimi-k3           &  88 &  6.4 & 30.1 & $+23.7$ & $+21.8$ \\
        & Gemini-3.6-flash  &  36 & 21.1 & 18.3 & $-2.8$  & $+2.0$ \\
        & DeepSeek-v4-flash &  59 & 14.8 & 23.1 & $+8.3$  & $+4.1^{\dagger}$ \\
\midrule
LIBERO-RIG & Claude Opus 5     &  28 & 30.6 & 96.7 & $+66.1$ & $+0.3$ \\
           & GPT-5.6-sol       &  11 & 82.8 & 88.0 & $+5.2$  & $+3.5$ \\
           & Kimi-k3           &  43 & 57.8 & 72.0 & $+14.2$ & $+22.4$ \\
           & Gemini-3.6-flash  &  22 & 35.6 & 36.7 & $+1.1$  & $+46.6$ \\
           & DeepSeek-v4-flash$^{\S}$ &  28 & 25.6 & 77.8 & $+52.2$ & $+10.8$ \\
\bottomrule
\end{tabular}
\end{table}

\begin{table}[t]
\centering
\caption{\textbf{Model scores across task families, with and without a library}, the values behind \cref{fig:model_scores,fig:library_gain}. TEST success (\%), mean $\pm$ SD over three runs (two where marked $^{\P}$). \textbf{Bold}: highest in each column. $^{\dagger}$First 58 RoboLab tasks only, where DeepSeek-v4-flash's runs stopped when the model was retired; its Vanilla score on the same tasks is $26.6 \pm 2.2$. $^{\S}$DeepSeek-v4.1-flash, its successor.}
\label{tab:models}
\footnotesize
\setlength{\tabcolsep}{3pt}
\renewcommand{\arraystretch}{1.15}
\resizebox{\textwidth}{!}{%
\begin{tabular}{l*{10}{c}}
\toprule
& \multicolumn{2}{c}{robosuite}
& \multicolumn{2}{c}{LIBERO}
& \multicolumn{2}{c}{LIBERO-PRO}
& \multicolumn{2}{c}{LIBERO-RIG}
& \multicolumn{2}{c}{RoboLab} \\
\cmidrule(lr){2-3}\cmidrule(lr){4-5}\cmidrule(lr){6-7}\cmidrule(lr){8-9}\cmidrule(lr){10-11}
Model & Lineage & Vanilla & Lineage & Vanilla & Lineage & Vanilla & Lineage & Vanilla & Lineage & Vanilla \\
\midrule
Claude Opus 5
& \textbf{\boldmath$97.1 \pm 1.4$} & \textbf{\boldmath$93.8 \pm 7.0$}
& \textbf{\boldmath$94.4 \pm 2.5$} & \textbf{\boldmath$95.2 \pm 1.1$}
& \textbf{\boldmath$91.8 \pm 0.5$} & \textbf{\boldmath$93.0 \pm 1.8$}
& \textbf{\boldmath$97.0 \pm 2.7$} & \textbf{\boldmath$96.7 \pm 2.2$}
& \textbf{\boldmath$68.0 \pm 3.5$} & \textbf{\boldmath$61.9 \pm 1.0$} \\
GPT-5.6-sol
& $84.8 \pm 6.4$ & $89.0 \pm 2.2$
& $88.3 \pm 3.9$ & $87.1 \pm 2.3$
& $88.4 \pm 2.8$ & $83.1 \pm 3.3$
& $91.5 \pm 5.7$ & $88.0 \pm 7.1$
& $42.4 \pm 0.6$ & $46.9 \pm 1.8$ \\
Gemini-3.6-flash
& $77.6 \pm 6.6$ & $70.5 \pm 0.8$
& $82.4 \pm 3.9$ & $71.4 \pm 4.1$
& $76.2 \pm 1.2^{\P}$ & $63.5 \pm 3.8$
& $83.3 \pm 18.9^{\P}$ & $36.7 \pm 4.0$
& $20.3 \pm 2.4$ & $18.3 \pm 3.6$ \\
Kimi-k3
& $88.1 \pm 2.2$ & $76.2 \pm 5.9$
& $77.8 \pm 1.3$ & $77.8 \pm 1.8$
& $80.5 \pm 3.5$ & $78.3 \pm 3.5$
& $94.4 \pm 5.3$ & $72.0 \pm 5.8$
& $51.9 \pm 4.9$ & $30.1 \pm 2.9$ \\
DeepSeek-v4-flash
& $77.0 \pm 13.5$ & $70.0 \pm 1.4$
& $65.9 \pm 9.1$ & $59.3 \pm 3.6$
& $75.5 \pm 4.5$ & $57.5 \pm 1.2$
& $88.6 \pm 0.4^{\S}$ & $77.8 \pm 1.5^{\S}$
& $30.7 \pm 5.1^{\dagger}$ & $23.1 \pm 2.1$ \\
\bottomrule
\end{tabular}%
}
\end{table}

\subsection{The Library Against a Rerun's Noise}
\label{app:library_null}

We pool 1,510 matched model--task pairs across robosuite, LIBERO, LIBERO-PRO, and RoboLab, each with three Vanilla repetitions and three Lineage repetitions (two on 171 pairs). Comparing means, Lineage scores higher than Vanilla on 96 of the 214 pairs where every Vanilla repetition scores zero. On the other 1,296 pairs it scores higher on 511 and lower on 354. Because a pair's Vanilla score is itself noisy, some of these changes would appear in any rerun. We therefore compare each Lineage repetition with a third Vanilla repetition, both read against the mean of the first two Vanilla repetitions. On the 276 pairs where the first two Vanilla repetitions score zero, a Lineage repetition scores above zero in 32.5\% of cases and at least 50\% in 23.1\%; the third Vanilla repetition does so in 22.5 and 13.0\%. On the 1,234 pairs with some Vanilla success, a Lineage repetition scores above the Vanilla mean in 33.7\% of cases and below it in 23.1\%; the third Vanilla repetition, in 26.1 and 31.9\%. The library therefore helps by 8 to 10 points on both kinds of pair. On unsolved pairs the margin comes from LIBERO-PRO and RoboLab; on LIBERO a third Vanilla repetition does as well. Bucketing pairs by the full Vanilla mean would be biased by regression to the mean, which is why the split-repetition comparison is used (\cref{tab:library_null}).

\begin{table}[H]
\centering
\caption{\textbf{The library against a rerun's noise.} Share of comparisons (\%) in which one repetition scores above (or below) the reference, the mean of the first two Vanilla repetitions of the same model--task pair; 1,510 pairs over robosuite, LIBERO, LIBERO-PRO, and RoboLab. A Lineage repetition is compared with what a third Vanilla repetition does against the same reference.}
\label{tab:library_null}
\small
\setlength{\tabcolsep}{4pt}
\begin{tabular}{p{3.3cm}lccc}
\toprule
Pairs & Outcome & Lineage rep. & Third Vanilla rep. & Margin \\
\midrule
276 unsolved (first two Vanilla at 0) & above 0 & 32.5 & 22.5 & $+10.0$ \\
 & at least 50\% & 23.1 & 13.0 & $+10.1$ \\
1,234 with some Vanilla success & above the reference & 33.7 & 26.1 & $+7.6$ \\
 & below the reference & 23.1 & 31.9 & $-8.8$ \\
\bottomrule
\end{tabular}
\end{table}

\subsection{Library Crossover}
\label{sec:finding_crossover}

We run GPT-5.6-sol on RoboLab with libraries written by Opus or Kimi-k3. For each task, the recipient gets the library version the donor held before attempting that task, with the same filters as in Lineage; the task, recipient model, and two-hour budget stay the same, and the crossover runs do not add their own experience to the donor library. One selected donor lineage per model, three runs of GPT-5.6-sol over all 105 tasks per donor, 630 runs in all; each task's score averages its graded runs, leaving out seven judge crashes on policies that scored zero on validation. A useful library also reaches a higher score sooner. On the runs matched to the crossover (whose finals differ slightly from \cref{tab:fig4_by_family}), GPT-5.6-sol with Opus's library reaches 55.2\% best-so-far validation success at half budget, above its no-library final 51.4\%; For comparison, Kimi-k3 with its own library reaches 32.5\% at half budget, above its Vanilla final 28.8\%. In the crossover, the median time to first success among successful runs changes little; the library helps more runs succeed or reach higher scores within the budget.

\subsection{Replicate Spread and a Rig Defect}

Not every collapse is variance. (A self-exploration lineage starts from a \emph{donor} library evolved on an earlier family; its \emph{frozen} policy is the task-general program left at the end of exploration, and its \emph{authored} policy the one rewritten once the instruction is disclosed.) The three Opus self-exploration replicates seeded from the LIBERO-RIG donor first read 26.7, 34.1, and 0.5\% on RoboLab with the frozen policy, but the collapsed replicate was one of 23 lineages across four models that never discovered a wrong-axis gripper offset in the tool frame, a rig defect. Rerun from the start under the corrected frame, it finished at 25.0\% frozen and 31.8\% authored, close to but just below its two siblings, which had found the offset on their own. Its Gemini-3.6-flash counterparts did not recover: twelve rerun Gemini-3.6-flash lineages solved 11 of 1,429 graded tasks with the frame fixed. Only two self-exploration separations between models survive the replicate spread: Gemini-3.6-flash at or near zero on every RoboLab donor against an Opus mean of 24 to 37\%, and Kimi-k3 ahead of Opus on LIBERO, 38.0 against 26.9\% for the frozen policy.

\subsection{Generalization Tiers, Cell by Cell}

\begin{figure}[H]
    \centering
    \includegraphics[width=\linewidth]{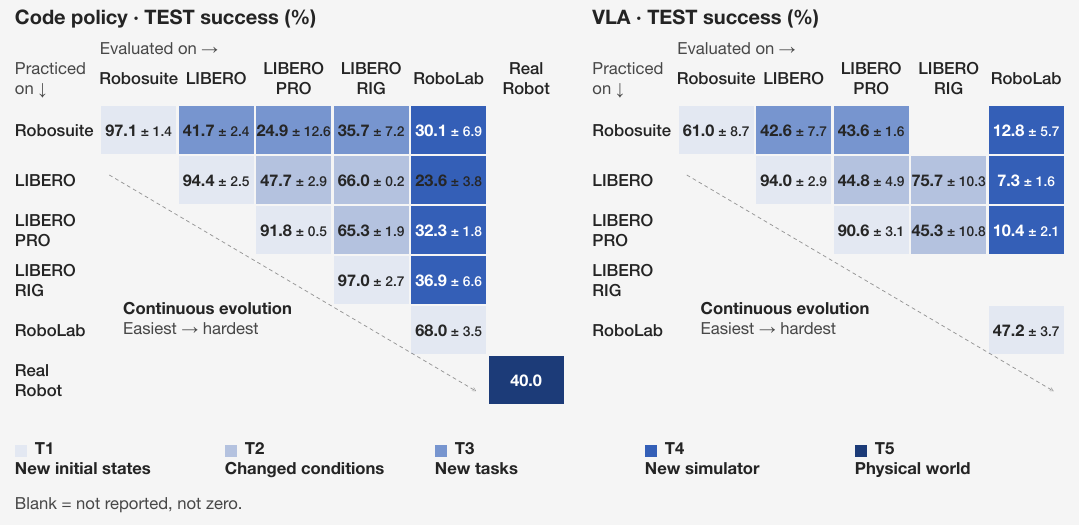}
    \vspace{-6pt}
    \caption{\textbf{Where a gain comes from and how far it reaches, for Opus.} Tiered TEST success (\%) of code-as-policy (CaP) and auto-research (AR) agents driven by Claude Opus~5. Rows are the family an agent practiced in, columns the family its policy was graded on, both ordered easiest to hardest. The diagonal is the practiced family (T1); cells to its right are graded at increasing distance from practice (T2--T4). \Cref{tab:tier_cells} gives the numbers. The two VLA LIBERO-RIG cells average each checkpoint's own-suite cells over the ten fault axes. The real-robot cell (T5, 40.0) is Opus's frozen physical code policy on the four non-stacking tasks; that run starts from an empty library (\cref{sec:sb2_hardware}), so it is not a transfer from simulation.}
    \label{fig:opus_tier_table}
\end{figure}

\textbf{A fix for one hardware fault does not carry to another.} We take the policy each lineage wrote for a LIBERO-RIG course task and run it, frozen, on the same task under the other faults. It scores within $-10$ to $+10$ points of the policy the same model had written for that task on LIBERO-PRO, before it met any fault ($+2$ for Opus, $-4$ for GPT-5.6-sol, $-10$ for Kimi-k3, $+10$ for DeepSeek-v4-flash, and $+3$ for Gemini-3.6-flash on validation at the moderate level; \cref{app:rig_level2}). Success also falls with severity (67, 60, and 52\% on validation from the moderate to the most severe level), so a severe fault approaches the cost of a scene change; proprioception noise is the axis where code's correction runs out (73, 37, and 4\%; \cref{tab:rig_by_axis}). The no-library control shows the same profile, so this pattern comes from code-as-policy with SAM3 perception, not from the learning loop. End-to-end VLA policies degrade instead of breaking: at the same moderate level, pooled over eight models, their success falls from 93 to 45\% when the scene camera is moved and to 75\% under a camera-to-arm calibration error~\citep{liberorig}.

\textbf{Without the instruction, reach depends on the donor library and on the model.} In self-exploration each lineage starts from a library evolved on an earlier family (its donor). The robosuite, LIBERO, and LIBERO-PRO donors are successive snapshots of one library, and the LIBERO-RIG donor branches from it. On RoboLab, a simulator none of them saw, Opus lineages seeded from the four donors reach 30.1, 23.6, 32.3, and 36.9\% once the instruction is disclosed; apart from the robosuite snapshot, more experience in the donor gives more reach, a tendency rather than a rule at three replicates. For Kimi-k3 and GPT-5.6-sol the LIBERO-PRO donor is the best of the four and the LIBERO-RIG donor among the worst (\cref{tab:selfexp_models}), so what a course adds to a library depends on the model that inherits it.

\begin{figure}[H]
    \centering
    \includegraphics[width=\linewidth]{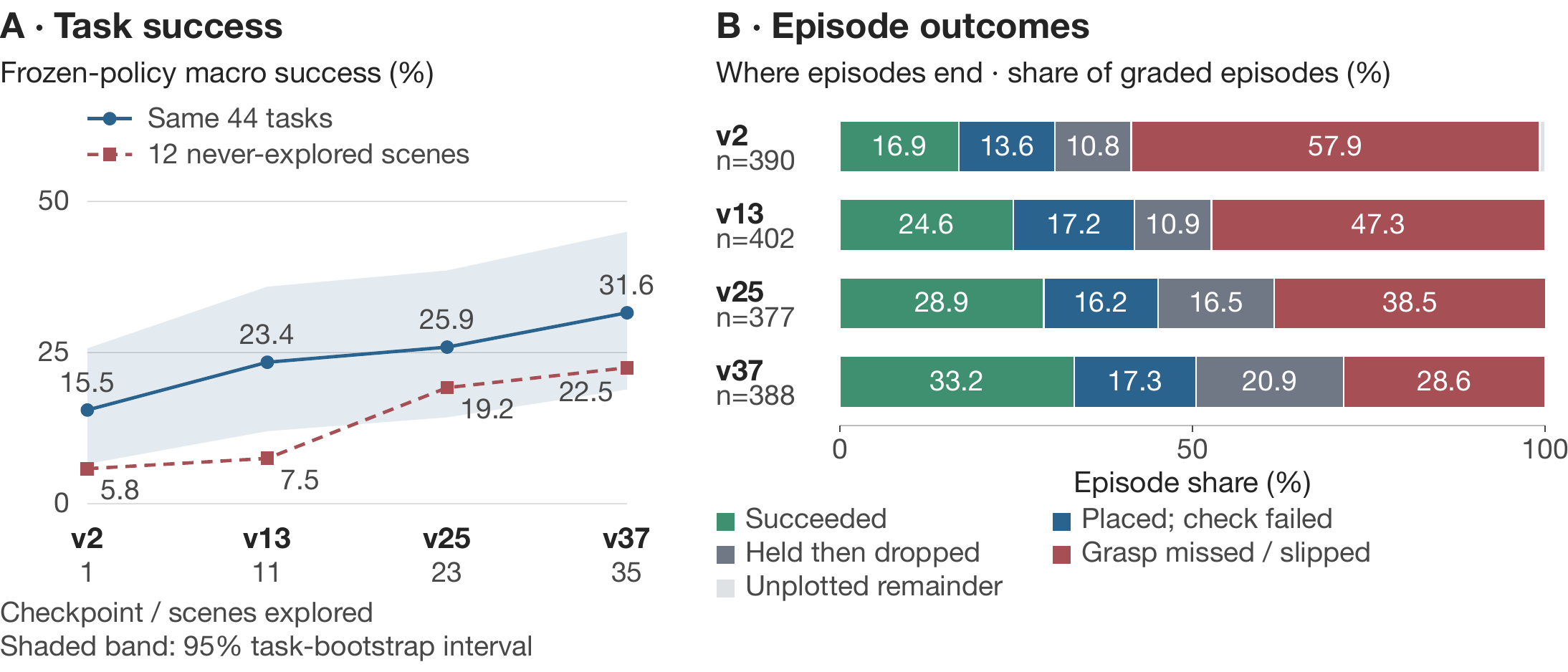}
    \vspace{-6pt}
    \caption{\textbf{A task-blind lineage graded on scenes it has not yet reached.} One Opus self-exploration lineage on RoboLab, its task-general program frozen at library versions v2, v13, v25, and v37 (after exploring 1, 11, 23, and 35 of its 47 scenes) and graded ten times on the same 44 probe tasks, one per scene, including scenes it had not yet reached. Nothing is rewritten per task. (A) Success on the same 44 tasks and on the 12 scenes the lineage never explored; grey is the 95\% bootstrap interval over tasks. (B) Where graded episodes end. This is one lineage graded on one fixed layout per task, so the intervals cover tasks, not lineages. Of 33 tasks graded before and after their scene was explored, 26 did not change.}
    \label{fig:unseen_probe}
\end{figure}

\subsection{Code and Weights}

\begin{table}[H]
\centering
\caption{\textbf{Code against weights, family by family.} Opus code (CaP) and weights (VLA) lineages. \emph{VLA $\geq$ CaP} compares the first trial of each paradigm task by task. \emph{All ten TEST states solved} is the share of graded (trial, task) cells with perfect TEST success: three CaP trials per family, three VLA trials on robosuite, one elsewhere.}
\label{tab:vla_vs_cap}
\small
\begin{tabular}{lcccccc}
\toprule
& & \multicolumn{2}{c}{Mean TEST success} & & \multicolumn{2}{c}{All ten TEST states solved} \\
\cmidrule(lr){3-4} \cmidrule(lr){6-7}
Family & Tasks & CaP & VLA & VLA $\geq$ CaP & CaP & VLA \\
\midrule
robosuite  &   7 & 97.1 & 61.0 & 14\% & 86\% & 29\% \\
LIBERO     &  40 & 94.4 & 94.0 & 80\% & 85\% & 75\% \\
LIBERO-PRO & 160 & 91.8 & 90.6 & 79\% & 81\% & 70\% \\
RoboLab    & 105 & 68.0 & 47.2 & 48\% & 46\% & 14\% \\
\bottomrule
\end{tabular}
\end{table}

\begin{table}[H]
\centering
\caption{\textbf{Code and weights under each LIBERO-RIG fault axis.} TEST success (\%) of LIBERO-practiced Opus policies on LIBERO-RIG at the moderate level. The control-frequency axis is excluded from every LIBERO-RIG result: the code-as-policy harness runs the rig at its nominal rate and cannot apply it. The VLA row is the LIBERO lineage checkpoints graded on the cells of their own suite, three to six checkpoint-cells per axis.}
\label{tab:vla_rig}
\small
\def\rigrot#1{\rotatebox{55}{\makebox[0pt][l]{#1}}}
\begin{tabular}{l*{11}{c}}
\toprule
\rule{0pt}{1.85cm}Fault & \rigrot{nominal} & \rigrot{cam.\ orientation} & \rigrot{cam.\ position} & \rigrot{cam.\ field of view} & \rigrot{wrist-cam mount} & \rigrot{gripper pads} & \rigrot{frame rotation} & \rigrot{actuation gain} & \rigrot{base pose} & \rigrot{table height} & \rigrot{proprio.\ noise} \\
\midrule
CaP & 96.4 & 7.0 & 8.7 & 11.3 & 95.3 & 94.2 & 95.6 & 93.4 & 92.1 & 88.2 & 73.7 \\
VLA & 86.2 & 56.7 & 60.0 & 90.0 & 55.0 & 66.7 & 83.3 & 81.7 & 81.7 & 83.3 & 98.3 \\
\bottomrule
\end{tabular}
\end{table}

\textbf{Teacher and student.} \Cref{fig:sb2_teacher_student}a pairs each task's VLA Lineage TEST score with the TEST score of the per-task code teacher whose rollouts it trained on (robosuite is omitted because its teachers have validation scores only). The student scores above its teacher on 15 of 305 tasks (2 LIBERO, 9 LIBERO-PRO, 4 RoboLab). Of the 11 tasks whose teacher solves no TEST state (3 LIBERO-PRO, 8 RoboLab), the student succeeds on some states of 2, both LIBERO-PRO. The agent can modify a teacher to collect the data it needs, but in these runs that rarely lifts a task the teacher cannot do at all.

\textbf{What the two libraries keep.} A code library holds executable control skills, 306 to 322 per RoboLab lineage, of which a final program imports a median of five. A VLA library holds the research process: collection and training scripts and a memory of what went wrong, with no checkpoints or data. That is enough to spare a lineage from rediscovering the label format: nine of twelve Vanilla sessions checked it by open-loop replay, against two of twelve Lineage sessions. Inheriting the research process pays only where the task is hard: a VLA lineage ties the no-library VLA on LIBERO (94.0 each), trails it on LIBERO-PRO (90.6 against 92.4), and leads it on RoboLab (47.2 against 43.1; \cref{tab:t1_main}), each from one run. Extra development time pays through data: on RoboLab, sessions that spent their second half collecting more data gained about four times as much as sessions that retrained on the same data (\cref{tab:vla_extra_time}); tasks that require two or more placements score about half as well as one-placement tasks, while instructions that require reasoning about attributes or counts cost nothing detectable (\cref{tab:vla_placements}). One session found that the harness's velocity labels, differenced from teacher commands, replayed to 4.37\,rad of arm travel where the teacher had produced 14.80, an error the training loss did not show.

\begin{table}[H]
\centering
\caption{\textbf{Code agents write about the world; VLA agents write about the pipeline.} Share of memory entries touching each topic. CaP: the 2{,}368 memory entries of the three corrected-frame RoboLab lineages' latest libraries and the 80 of the robosuite-and-LIBERO library. VLA: the 3{,}924 headings and bullets in the memory files of the 20 VLA lineages. An entry can touch several topics; topics are keyword classes.}
\label{tab:memory_topics}
\small
\begin{tabular}{lcccc}
\toprule
& \multicolumn{2}{c}{CaP} & \multicolumn{2}{c}{VLA} \\
\cmidrule(lr){2-3} \cmidrule(lr){4-5}
Share of memory entries touching\ldots & RoboLab & LIBERO & RoboLab & LIBERO/PRO \\
\midrule
grasp / contact geometry        & 69\% & 64\% & 11\% & 12\% \\
perception / grounding          & 58\% & 42\% &  8\% & 10\% \\
frame / calibration / kinematics & 46\% & 38\% &  7\% & 15\% \\
motion / controller             & 40\% & 39\% &  8\% &  8\% \\
training recipe / checkpoint    &  6\% &  6\% & 36\% & 39\% \\
training data / labels          &  2\% &  8\% & 31\% & 21\% \\
grading noise                   &  1\% &  0\% & 20\% & 17\% \\
inference settings              &  1\% &  0\% & 17\% & 15\% \\
\bottomrule
\end{tabular}
\end{table}

\subsection{Design Rules}
\label{app:design_rules}
\label{sec:design_rules}

Eight rules for anyone building a self-evolving robot agent, each from a finding of this study:
\begin{enumerate}
\item Qualify the base model on the hardest family first; a model below a minimum level of capability writes almost nothing worth saving, so its own library cannot help it (\cref{sec:sb2_better_llm}).
\item Spend on grounded perception tools rather than on a multimodal model, and give image access for diagnosis (\cref{sec:sb2_no_pixels}).
\item Measure a model's time to first success on a pilot and make sessions at least that long; beyond that, scale with parallel long runs. Report TEST, not VAL, scores (\cref{sec:sb2_sessions}).
\item Seed hard deployments with a library built around control and grasp routines, and test a library from another model if a model's own does not help (\cref{sec:sb2_test_time,sec:sb2_new_scenes}).
\item Choose between code and weights by the kind of change you expect: code for faults the program can observe, and weights for camera motion and new instruction wording. When training weights, spend the budget on demonstrations first (\cref{sec:sb2_code_weights}).
\item Never let a checker the agent wrote itself decide when practice stops unless an independent check confirms it, and record where each memory came from so it can be re-validated (\cref{sec:sb2_mistakes}).
\item Set up the start state of each evaluation independently of the agent's own resets, and do not end practice because of a success streak on scenes the agent reset itself (\cref{sec:sb2_hardware}).
\item Keep re-observation in the loop where things go wrong, and schedule re-tests of rules written after a single incident (\cref{sec:sb2_preferences,sec:finding_real_world_limits}).
\end{enumerate}

\subsection{Limitations}
\label{app:limitations}
\label{sec:limitations}

Three runs per condition bound what we can distinguish, and replicate spread often exceeds the gap between neighboring models. The budget-matched VLA runs on LIBERO, LIBERO-PRO, and RoboLab are one run each, and every physical result is one practice run per agent. Agent-as-policy uses five staged scenes per task, has no Opus run, and uses a different stopping rule for Block Stacking. The physical runs start from an empty library, so we measure learning on hardware, not transfer from simulation to hardware. Each model runs through its own coding tool, so a model comparison is a comparison of model-and-tool pairs. Budgets are matched in wall-clock time, not compute. The sealed judge protects runtime state, but it does not hide everything an agent can read (\cref{app:shortcuts}). A zero score in three runs does not show that a model can never solve a task, only that it did not within the budget. Finally, model rankings change when the agent is not given the task instruction: in self-exploration Kimi-k3 matches or beats Opus on five of seven settings (\cref{tab:selfexp_models}).

\subsection{Shortcuts the Agents Found}
\label{app:shortcuts}
\label{sec:finding_shortcuts}

The seal is not absolute (\cref{app:sb2_protocol}), and agents found and used its openings without being asked, as reasoning models do when an evaluation leaves one~\citep{bondarenko2025specification}. The most common was reading static asset files: Opus parsed the object XML in 115 of 179 LIBERO runs and 552 of 1,015 LIBERO-PRO runs, and the scene USD in 1,533 of 1,743 RoboLab runs; Kimi-k3 read object XML in 284 of 1,260 non-RoboLab runs. This was allowed, and it helped, but a physical robot cannot read an object's geometry before acting. Policies can also call the environment's success check, \texttt{env.check\_success()}, which we leave callable as a stop signal; it returns only whether the task is done, and the judge grades with its own copy. How often a policy stops or branches on it depends on the model: 47\% of GPT-5.6-sol's graded RoboLab policies against 5\% of Opus's. 

Task text also leaks. LIBERO encodes each instruction in the task file's name, its language block, and its goal predicate, so a sealed condition must strip all three. When we audited the libraries meant to seed our task-blind lineages, the post-LIBERO Opus library already quoted a hidden instruction word for word, in a memory recording ``Turn on the stove'', and the post-LIBERO-PRO library matched 10 of 40 LIBERO and 27 of 160 LIBERO-PRO instructions, so we could not use it. The harness now refuses exploration sessions that touch held-out task identifiers and rewrites prompts that contain hidden material. In one pilot we dropped, we also tried to hide the success check: the patch covered only our wrapper, and Opus and GPT-5.6-sol found \texttt{env.env.check\_success()} within a single session. A seal has to be tested by trying to break it, not by checking that the patch applied.

\subsection{Sources for the Landscape Table}
\label{app:landscape}

\begin{table}[t]
\centering
\caption{\textbf{Self-evolving robot agents released since mid-September 2026}, in the columns of \cref{tab:sb2_landscape}. \yes{} yes, \half{} partial, \no{} no, -- not reported, from each paper's full text. \emph{Controls}: budget-matched restarts and a no-library run. \emph{Judge}: final policy graded on states and logic the agent cannot access.}
\label{tab:landscape_recent}
\footnotesize
\setlength{\tabcolsep}{2.6pt}
\resizebox{\textwidth}{!}{%
\begin{tabular}{lcccccccl}
\toprule
& \multicolumn{3}{c}{Policy form} & & Library across & & & \\
\cmidrule(lr){2-4}
System & CaP & AaP & AR & Sim / real & tasks & Controls & Judge & Generalization measured \\
\midrule
\citet{kuroki2026transfer} & \yes & \no & \no & sim & \yes & \half & \half & new tasks \\
Harness evolution~\citep{wei2026harness} & \no & \yes & \no & sim & \half & \no & \half & new tasks, new benchmark \\
SimEX~\citep{simex2026} & \yes & \no & \no & both & \half & \no & \half & new simulator, sim-to-real \\
RPG~\citep{rpg2026} & \half & \yes & \no & both & \yes & \no & \half & perturbations, sim-to-real \\
Robo-COP~\citep{robocop2026} & \half & \yes & \yes & both & \no & \no & \half & new initial states \\
Embodied TM~\citep{etm2026} & \yes & \no & \no & sim & \yes & \no & \half & new layouts, new tasks \\
RoboRSI~\citep{roborsi2026} & \half & \yes & \half & both & \yes & \no & \half & perturbations \\
\bottomrule
\end{tabular}%
}
\end{table}

\Cref{tab:sb2_landscape} records what each system's own paper or announcement reports; ``partial'' marks a feature that appears in a limited form. \textbf{CaP-X}~\citep{capx}: multi-turn program generation with execution feedback and an RL-trained coder (CaP-RL); simulation (robosuite, LIBERO-PRO, BEHAVIOR) and real robots; a task-agnostic skill library built from pooled rollouts and evaluated on the same tasks; a skill-library ablation without matched restarts; transfer to perturbations and from simulation to a real arm. \textbf{ENPIRE}~\citep{enpire}: the agent edits training code for learned policies (and can synthesize code policies) on a fleet of real bimanual stations, with simulation in Gym-PushT and RoboCasa; cross-task transfer appears through a written summary in one case; fixed test seeds and protected metrics in some settings, but agent-built verifiers. \textbf{ASPIRE}~\citep{aspire}: code-as-policy with a growing skill library inherited by later tasks; an empty-library ablation but no budget-matched restarts; split debug, validation, and evaluation seeds with simulator ground-truth APIs forbidden; perturbations and unseen long-horizon tasks, with a limited real-robot study. \textbf{RATs}~\citep{rats}: code-as-policy agents build a library through play and reuse it on downstream tasks; a no-play ablation and a compute-matched baseline but no independent restarts; held-out tasks, perturbations, a held-out simulator, and a preliminary real study. \textbf{Waddle}~\citep{waddle2026}: agents control real arms by viewing cameras, writing code, and re-planning live, share a skill library, and in one case train an ACT policy; the announcement reports no controls or held-out evaluation. The rows of \cref{tab:landscape_recent} come from each paper's full text. \textbf{Kuroki \& Tang}~\citep{kuroki2026transfer}: a coding agent optimizes frozen programs for RoboCasa tasks and reuses an archive of four of them on nine new tasks; a no-archive comparison runs under shared session limits but unequal total compute; held-out test states, though the policies read ground-truth simulator segmentation. \textbf{Harness evolution}~\citep{wei2026harness}: an optimizer agent evolves the harness of a task agent that acts through a 3-D interface; selection-rule ablations but no restart or no-library control; held-out LIBERO tasks, while the RoboCasa campaign has no held-out set. \textbf{SimEX}~\citep{simex2026}: per-task programs over a persistent toolbox built within one task family; evaluation in a different simulator and on a real two-arm robot, after five scored adaptation rollouts in the evaluation environment. \textbf{RPG}~\citep{rpg2026}: an offline coding model revises a shared skill library and system prompt that a runtime agent uses; held-out seeds, but improvement uses the same success check and privileged simulator state; its rounds are not budget-matched. \textbf{Robo-COP}~\citep{robocop2026}: an orchestrator writes per-episode programs that call a VLA and fine-tunes that VLA, per task; held-out initial states of the same tasks. \textbf{Embodied Turing Machines}~\citep{etm2026}: coding agents grow one shared library and frozen per-task programs on RoboDojo; held-out test layouts, with simulator ground truth used during development. \textbf{RoboRSI}~\citep{roborsi2026}: agents repair a shared skill hierarchy and compose it at run time; most results improve on the evaluation tasks themselves, and only LIBERO-Plus uses a frozen held-out evaluation.

\section{Additional Tables}
\label{sec:appendix}

\begin{table}[H]
\centering
\caption{\Cref{fig:rig_perturbation} as numbers: the LIBERO-practiced Opus policies re-graded on LIBERO-RIG, VAL success (\%) averaged over tasks per perturbation axis and severity, the lineage as the mean of three runs over all 40 LIBERO tasks and the best-of-3 winners over the same 40 tasks. Columns s3 to s5 are severity levels 3 (moderate) to 5 (most severe). The lineage's s3 TEST column gives the values behind \cref{fig:opus_tier_table}; its mean over the ten axes, 65.9 pooled and $66.0 \pm 0.2$ over the three runs, is that figure's LIBERO$\to$LIBERO-RIG code-as-policy cell (T2). LIBERO-RIG's control-frequency axis is omitted: the code-as-policy harness builds the rig at the nominal 20 Hz and cannot apply it, so it is excluded from every LIBERO-RIG result in this paper. Complete grades only.}
\label{tab:rig_by_axis}
\small
\begin{tabular}{lccccccc}
\toprule
& \multicolumn{4}{c}{Library lineage} & \multicolumn{3}{c}{No-library best-of-3} \\
\cmidrule(lr){2-5} \cmidrule(lr){6-8}
Axis & s3 & s4 & s5 & s3 TEST & s3 & s4 & s5 \\
\midrule
frame rotation & 95.9 & 94.9 & 89.7 & 95.6 & 73.8 & 71.2 & 68.0 \\
actuation gain & 95.3 & 85.7 & 67.4 & 93.4 & 70.2 & 60.8 & 51.5 \\
proprioception noise & 73.4 & 37.1 & 4.0 & 73.7 & 62.0 & 40.8 & 6.2 \\
camera orientation & 6.8 & 7.1 & 7.5 & 7.0 & 2.8 & 2.5 & 3.0 \\
camera position & 8.7 & 11.4 & 10.2 & 8.7 & 2.8 & 4.8 & 6.2 \\
camera field of view & 12.6 & 8.8 & 6.8 & 11.3 & 17.2 & 5.4 & 3.2 \\
wrist-camera mount & 96.7 & 96.8 & 96.8 & 95.3 & 76.2 & 74.8 & 75.8 \\
base pose & 93.3 & 78.8 & 60.9 & 92.1 & 73.2 & 61.5 & 47.2 \\
table height & 87.0 & 80.2 & 76.8 & 88.2 & 56.0 & 51.0 & 48.5 \\
gripper pads & 95.6 & 94.8 & 94.7 & 94.2 & 76.0 & 75.8 & 75.2 \\
\midrule
Macro over the ten axes & 66.5 & 59.5 & 51.5 & 65.9 & 51.0 & 44.8 & 38.5 \\
Nominal rig & \multicolumn{3}{c}{97.8} & 96.4 & \multicolumn{3}{c}{80.3} \\
\bottomrule
\end{tabular}
\end{table}

\begin{table}[H]
\centering
\caption{The cells of \cref{fig:opus_tier_table} as numbers: TEST success (\%) of the Claude Opus 5 code-as-policy (CaP) and auto-research (AR) policies, rows the family practiced in and columns the family graded on, mean $\pm$ sample standard deviation over three replicates where replicated; the AR diagonal cells of LIBERO, LIBERO-PRO, and RoboLab are one trial each, with $\pm$ over their four or twelve suites. The diagonal is T1, the off-diagonal cells T2 to T4 as \cref{fig:opus_tier_table} marks them. The T5 cell, 40.0, is the physical lineage of \cref{tab:real_world} and is not repeated here. For the T3/T4 CaP cells the frozen exploration policy, graded with the instruction only at run time, scores (robosuite$\to$LIBERO) $26.9 \pm 7.6$ (robosuite$\to$LIBERO-PRO) $14.6 \pm 9.5$ (robosuite$\to$LIBERO-RIG) $24.3 \pm 11.4$ (robosuite$\to$RoboLab) $25.8 \pm 3.7$ (LIBERO$\to$RoboLab) $19.1 \pm 5.8$ (LIBERO-PRO$\to$RoboLab) $26.7 \pm 2.5$ (LIBERO-RIG$\to$RoboLab) $28.6 \pm 4.8$. $^{\S}$Two of the three replicates found a wrong-axis gripper offset in the simulated tool frame (a rig defect) on their own; the third was rerun from the start with the offset corrected (authored 34.5, 44.3, and 31.8; the last over the 115 of its 120 exploration tasks that were graded).}
\label{tab:tier_cells}
\small
\begin{tabular}{llccccc}
\toprule
Practiced on & Policy & robosuite & LIBERO & LIBERO-PRO & LIBERO-RIG & RoboLab \\
\midrule
robosuite & CaP & $97.1 \pm 1.4$ & $41.7 \pm 2.4$ & $24.9 \pm 12.6$ & $35.7 \pm 7.2$ & $30.1 \pm 6.9$ \\
 & AR & $61.0 \pm 8.7$ & $42.6 \pm 7.7$ & $43.6 \pm 1.6$ & --- & $12.8 \pm 5.7$ \\
\midrule
LIBERO & CaP &  & $94.4 \pm 2.5$ & $47.7 \pm 2.9$ & $66.0 \pm 0.2$ & $23.6 \pm 3.8$ \\
 & AR &  & $94.0 \pm 2.9$ & $44.8 \pm 4.9$ & $75.7 \pm 10.3$ & $7.3 \pm 1.6$ \\
\midrule
LIBERO-PRO & CaP &  &  & $91.8 \pm 0.5$ & $65.3 \pm 1.9$ & $32.3 \pm 1.8$ \\
 & AR &  &  & $90.6 \pm 3.1$ & $45.3 \pm 10.8$ & $10.4 \pm 2.1$ \\
\midrule
LIBERO-RIG & CaP &  &  &  & $97.0 \pm 2.7$ & $36.9 \pm 6.6^{\S}$ \\
 & AR &  &  &  & --- & --- \\
\midrule
RoboLab & CaP &  &  &  &  & $68.0 \pm 3.5$ \\
 & AR &  &  &  &  & $47.2 \pm 3.7$ \\
\bottomrule
\end{tabular}
\end{table}

\begin{table}[t]
\centering
\caption{Self-exploration for every model that ran it: TEST success (\%) of the policy authored once the instruction is disclosed (authored) and of the task-general exploration policy graded with the instruction only as a run-time input (frozen), for each donor library and target family of \cref{fig:opus_tier_table}, mean $\pm$ sample standard deviation over three replicate lineages. The Opus column is that figure's T3/T4 code-as-policy cells and their frozen-policy ablation; DeepSeek-v4-flash did not run this campaign. The RoboLab rows are under the corrected tool frame: lineages that never discovered the wrong-axis gripper offset were rerun from session zero (one Opus, three GPT-5.6-sol, seven Kimi-k3, and all twelve Gemini-3.6-flash lineages, each through its 48 sessions except Gemini's LIBERO-RIG-donor third replicate at 45), and the lineages that had found the offset themselves keep their full-course reading. A fully errored cell scores zero.}
\label{tab:selfexp_models}
\small
\setlength{\tabcolsep}{3.5pt}
\begin{tabular}{llcccc}
\toprule
Donor $\to$ target & Policy & Claude Opus 5 & GPT-5.6-sol & Kimi-k3 & Gemini-3.6-flash \\
\midrule
robosuite$\to$LIBERO & authored & $41.7 \pm 2.4$ & $34.7 \pm 13.6$ & $43.7 \pm 4.4$ & $10.2 \pm 2.9$ \\
 & frozen & $26.9 \pm 7.6$ & $20.9 \pm 6.1$ & $38.0 \pm 4.5$ & $8.3 \pm 7.2$ \\
\midrule
robosuite$\to$LIBERO-PRO & authored & $24.9 \pm 12.6$ & $21.5 \pm 12.1$ & $26.7 \pm 7.2$ & $6.6 \pm 6.2$ \\
 & frozen & $14.6 \pm 9.5$ & $11.3 \pm 9.0$ & $20.8 \pm 8.9$ & $4.2 \pm 4.4$ \\
\midrule
robosuite$\to$LIBERO-RIG & authored & $35.7 \pm 7.2$ & $23.3 \pm 20.5$ & $27.7 \pm 5.2$ & $1.2 \pm 2.0$ \\
 & frozen & $24.3 \pm 11.4$ & $8.2 \pm 7.1$ & $8.9 \pm 8.2$ & $2.0 \pm 2.5$ \\
\midrule
robosuite$\to$RoboLab & authored & $30.1 \pm 6.9$ & $16.6 \pm 10.6$ & $30.9 \pm 3.1$ & $0.0 \pm 0.0$ \\
 & frozen & $25.8 \pm 3.7$ & $9.8 \pm 6.3$ & $23.0 \pm 3.1$ & $0.0 \pm 0.0$ \\
\midrule
LIBERO$\to$RoboLab & authored & $23.6 \pm 3.8$ & $24.8 \pm 9.4$ & $30.0 \pm 8.5$ & $3.9 \pm 3.9$ \\
 & frozen & $19.1 \pm 5.8$ & $15.2 \pm 6.2$ & $22.8 \pm 5.1$ & $2.1 \pm 2.3$ \\
\midrule
LIBERO-PRO$\to$RoboLab & authored & $32.3 \pm 1.8$ & $28.3 \pm 2.9$ & $36.5 \pm 6.4$ & $0.0 \pm 0.0$ \\
 & frozen & $26.7 \pm 2.5$ & $17.0 \pm 1.7$ & $24.1 \pm 7.8$ & $0.0 \pm 0.0$ \\
\midrule
LIBERO-RIG$\to$RoboLab & authored & $36.9 \pm 6.6$ & $17.2 \pm 7.8$ & $12.9 \pm 15.8$ & $1.9 \pm 1.7$ \\
 & frozen & $28.6 \pm 4.8$ & $9.5 \pm 4.8$ & $8.4 \pm 11.2$ & $0.0 \pm 0.0$ \\
\bottomrule
\end{tabular}
\end{table}

\begin{table}[t]
\centering
\caption{The conditions of \cref{fig:avg_perf_vs_time} at the end of the budget, per family: best-so-far VAL success (\%) averaged over each family's cells at 60 minutes of self-evolve time (120 on RoboLab), and the equal-weight mean that the figure plots. First success: median minutes to the first validation success over the no-library runs that ever succeed; its Mean halves RoboLab's time to put it on the 60-minute scale of the other families. \Cref{fig:session_term} plots, per model, the Mean of no library minus best-of-3 against the Mean first success. The LIBERO-RIG with-library value is the mean of the seed-matched runs and the no-library value the mean of three 60-minute Vanilla repetitions, neither told the fault, on the 18-cell course; the best-of-3 control is read over the cells with a graded winner. RoboLab is the corrected-frame campaign over all 105 cells of its curriculum, all three conditions on the same cells; $^{\dagger}$DeepSeek-v4-flash's three RoboLab conditions are read on the first 58 cells, where its runs stopped when the model was retired.}
\label{tab:fig4_by_family}
\small
\setlength{\tabcolsep}{2.4pt}
\begin{tabular}{llcccccc}
\toprule
Model & Condition & robosuite & LIBERO & LIBERO-PRO & LIBERO-RIG & RoboLab & Mean \\
\midrule
Claude Opus 5 & with library & 97.6 & 95.2 & 92.5 & 96.1 & 71.1 & 90.5 \\
 & no library & 93.8 & 95.7 & 93.2 & 96.8 & 62.8 & 88.5 \\
 & best-of-3 & 87.1 & 62.5 & 49.3 & 31.1 & 52.8 & 56.6 \\
 & first success, min & 13.7 & 27.9 & 27.7 & 28 & 58 & 25.3 \\
\midrule
GPT-5.6-sol & with library & 85.7 & 90.5 & 90.5 & 92.4 & 47.8 & 81.4 \\
 & no library & 91.9 & 90.5 & 86.1 & 90.2 & 51.3 & 82.0 \\
 & best-of-3 & 90.0 & 94.5 & 73.7 & 82.5 & 55.0 & 79.2 \\
 & first success, min & 7.5 & 10.0 & 11.6 & 11 & 21 & 10.1 \\
\midrule
Kimi-k3 & with library & 89.5 & 77.8 & 80.7 & 93.5 & 53.3 & 79.0 \\
 & no library & 76.7 & 78.2 & 79.0 & 70.7 & 28.4 & 66.6 \\
 & best-of-3 & 57.1 & 67.8 & 71.4 & 58.3 & 5.8 & 52.1 \\
 & first success, min & 13.7 & 28.1 & 28.0 & 43 & 88 & 31.4 \\
\midrule
Gemini-3.6-flash & with library & 80.0 & 82.0 & 76.8 & 73.9 & 24.1 & 67.3 \\
 & no library & 70.5 & 70.6 & 63.4 & 37.4 & 19.4 & 52.3 \\
 & best-of-3 & 71.4 & 37.5 & 30.5 & 18.3 & 21.9 & 35.9 \\
 & first success, min & 11.2 & 25.9 & 25.0 & 22 & 36 & 20.4 \\
\midrule
DeepSeek-v4-flash & with library & 80.5 & 67.2 & 77.5 & 90.8 & 33.2$^{\dagger}$ & 69.9 \\
 & no library & 74.3 & 59.2 & 58.3 & 75.6 & 28.6$^{\dagger}$ & 59.2 \\
 & best-of-3 & 64.3 & 46.0 & 47.4 & 23.9 & 17.2$^{\dagger}$ & 39.8 \\
 & first success, min & 13.7 & 28.0 & 28.0 & 28 & 59 & 25.4 \\
\bottomrule
\end{tabular}
\end{table}

\begin{table}[t]
\centering
\caption{\textbf{Opus on the practiced task under all three conditions.} Claude Opus~5 on held-out seeds of the practiced task (T1), TEST success (\%), mean $\pm$ SD over three repeated runs where repeated; the auto-research cells on LIBERO, LIBERO-PRO, and RoboLab are one trial each, with $\pm$ over their four or twelve suites. \emph{Lineage} carries the library forward; \emph{Vanilla} is the matched no-library session; \emph{Best-of-3} spends the same budget as three independent attempts and keeps the best by validation score. $^{*}$End-of-run TEST of the validation-selected winners on the 47 robosuite and LIBERO tasks. LIBERO-RIG: the 18 cells of the course at the moderate level, the lineage over three seed-matched runs, Vanilla over three 60-minute runs per cell that are not told the fault. %
$^{\P}$Mean over the four non-stacking physical tasks, ten operator-staged scenes each (\cref{tab:model_scores}).}
\label{tab:t1_main}
\small
\begin{tabular}{lccccc}
\toprule
& \multicolumn{3}{c}{Code-as-policy} & \multicolumn{2}{c}{Auto-research VLA} \\
\cmidrule(lr){2-4} \cmidrule(lr){5-6}
Family & Lineage & Vanilla & Best-of-3 & Vanilla & Lineage \\
\midrule
robosuite   & $97.1 \pm 1.4$ & $93.8 \pm 7.0$ & 90.0 & $61.4 \pm 10.0$ & $61.0 \pm 8.7$ \\
LIBERO      & $94.4 \pm 2.5$ & $95.2 \pm 1.1$ & 73.8$^{*}$ & $94.0 \pm 3.6$ & $94.0 \pm 2.9$ \\
LIBERO-PRO  & $91.8 \pm 0.5$ & $93.0 \pm 1.8$ & 67.2 & $92.4 \pm 4.9$ & $90.6 \pm 3.1$ \\
LIBERO-RIG  & $97.0 \pm 2.7$ & $96.7 \pm 2.2$ & 30.6 & --- & --- \\
RoboLab     & $68.0 \pm 3.5$%
& $61.9 \pm 1.0$%
& 57.1%
& $43.1 \pm 9.3$ & $47.2 \pm 3.7$ \\
real robot  & 40.0$^{\P}$ & --- & --- & --- & --- \\
\bottomrule
\end{tabular}
\end{table}

\begin{figure}[t]
    \centering
    \includegraphics[width=\linewidth]{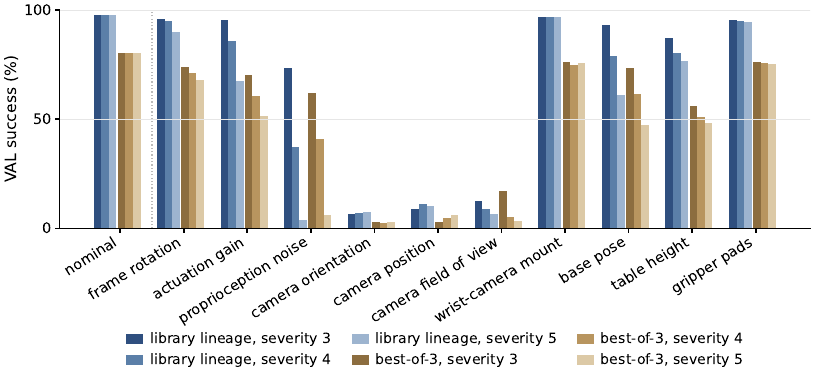}
    \vspace{-6pt}
    \caption{\textbf{Opus policies under LIBERO-RIG faults, with and without a library.} LIBERO-practiced Opus policies re-graded on LIBERO-RIG by fault axis and severity, validation success averaged over tasks: the library lineage (three lineages, all 40 tasks) against the no-library best-of-3 winners on the same tasks. \Cref{tab:rig_by_axis} lists the values.}
    \label{fig:rig_perturbation}
    \vspace{-8pt}
\end{figure}

\section{Benchmark Protocol Details}
\label{app:sb2_protocol}
\paragraph{Limits of the seal.}
The seal is not absolute, and we report its limits beside each result. Work environments have internet access, so public LIBERO initial states remain obtainable even though we remove them from the work image. RoboLab's validation feedback describes the VAL layouts, so an agent could use it to reconstruct them. Static task assets and task files are readable, and the success predicate is callable as a stop signal; \cref{sec:finding_shortcuts} reports what agents did with each. Audits found no reconstruction of test layouts in the inspected logs, but cannot establish that it was impossible.

\textbf{Budgets.} A CaP session lasts one hour per robosuite, LIBERO, LIBERO-PRO, or LIBERO-RIG task and two hours per RoboLab task, with up to eight agent-initiated submissions plus periodic host grades. AR receives the same total development time as CaP on LIBERO, LIBERO-PRO, and RoboLab, spent as one session per suite; robosuite's AR budget differs and is given in \cref{app:budgets} with the submission and early-stopping rules. Matching wall clock does not match compute, so we record environment steps, rollouts, submissions, API and token spend, and GPU time separately.

\textbf{Repetitions.} Every repeated campaign reports its count. Three repetitions of a Lineage develop three independent libraries; three repetitions of Vanilla develop three independent policies. Means and sample standard deviations are over repetitions, never over evaluation episodes. Where a campaign has one trial (the budget-matched VLA analyses on LIBERO, LIBERO-PRO, and RoboLab; every physical lineage), we say so beside the result.

\subsection{Terms}
\label{sec:terms}

\Cref{tab:terms} is a short reference for terms used in a specific sense; the text gives the authoritative definitions.
\begin{table}[t]
\centering
\caption{\textbf{Glossary.} Terms used in a specific sense in this paper, and where each is introduced.}
\label{tab:terms}
\footnotesize
\begin{tabular}{@{}l p{8.9cm} l@{}}
\toprule
Term & Meaning & Where \\
\midrule
\multicolumn{3}{@{}l}{\textbf{\emph{Policy forms}}} \\
\addlinespace
Code-as-policy (CaP) & The agent delivers a program; no language model runs at execution time. & \cref{sec:sear} \\
Auto-research (AR) & The agent runs training experiments that fine-tune a VLA from rollouts of its own code policies. & \cref{sec:sear} \\
Agent-as-policy (AaP) & The language model stays in the control loop and chooses each action from live observations. & \cref{sec:sear} \\
\addlinespace
\multicolumn{3}{@{}l}{\textbf{\emph{What persists}}} \\
\addlinespace
Library & The code skills and textual memories that persist between tasks; each task inherits and extends it. & \cref{sec:sear} \\
\addlinespace
\multicolumn{3}{@{}l}{\textbf{\emph{Settings and conditions}}} \\
\addlinespace
Vanilla & The no-library control: instruction and judge are given, and every task (in AR, every suite) starts from an empty library. & \cref{sec:sear} \\
Lineage & Vanilla with the library inherited from all earlier tasks; also one such run, with its own evolving library. & \cref{sec:sear} \\
Self-exploration & Neither instruction nor judge is given; the agent practices on its own and learns the task only afterwards. & \cref{sec:self_exploration} \\
Best-of-3 & Three independent no-library attempts at a third of the budget each; the best by validation score is kept. & \cref{sec:controls} \\
\addlinespace
\multicolumn{3}{@{}l}{\textbf{\emph{Evaluation}}} \\
\addlinespace
Judge & The sealed grader that rolls a frozen policy out on held-out states and scores it with the environment's own predicate. & \cref{sec:seal} \\
DEV / VAL / TEST & Initial-state bands: free for development; graded during development with feedback; graded once at the end and never shown. & \cref{sec:controls} \\
Cell & One task under one condition; the unit every score averages over. & \cref{app:rig} \\
Suite & The group of tasks one auto-research checkpoint is trained on: a LIBERO task suite, a LIBERO-PRO perturbation axis, or a block of RoboLab tasks. & \cref{sec:sb2_code_weights} \\
Tiers T1--T5 & How far the graded conditions are from practice: new initial states, perturbed conditions, unseen tasks, a new simulator, hardware. & \cref{sec:tiers} \\
\bottomrule
\end{tabular}
\end{table}

\section{Seed Band Allocations}
\label{app:seed_details}

For LIBERO and LIBERO-PRO, DEV uses initial states 0--29, VAL uses states 30--39, and TEST uses states 40--49. Work images remove the VAL and TEST initial states. RoboLab uses separate cached files containing 30 DEV, 10 VAL, and 10 TEST layouts; only the DEV layouts are supplied to the work environment. For robosuite, development permits agent-chosen seeds, VAL uses documented public seeds, and TEST seeds are derived from a judge-side private salt. For RoboLab, the salt permutes the cached TEST layouts rather than generating their contents.

Submission feedback includes aggregate success, per-seed outcomes, and judge metrics. On RoboLab, these metrics also expose VAL layout descriptions, allowing agents to replay those layouts locally.

\section{Interaction Budgets}
\label{app:budgets}

\textbf{Per-task budgets.} Robosuite, LIBERO, LIBERO-PRO, and LIBERO-RIG tasks receive a one-hour development budget, while RoboLab tasks receive two hours to accommodate longer simulation times. Each run permits up to eight agent-initiated submissions. Additional host-side evaluations run periodically, provide feedback, and are eligible for policy selection without consuming the manual submission allowance. Runs may terminate early upon perfect VAL success or exhaustion of the submission allowance.

\textbf{Budgets in other settings.} Self-exploration sessions receive two hours per scene, followed by a single authored rewrite once the instruction is disclosed. AR trains one checkpoint per suite, with ten hours per LIBERO suite, forty per LIBERO-PRO axis, and 17.5 per RoboLab suite, matching CaP's total budget for each of these families. Robosuite's two AR sessions receive five hours each; its curves use the first seven hours to match CaP, while TEST summaries use final checkpoints. In self-exploration AR receives six hours per three-scene group, or ten hours per RoboLab scene; the T3/T4 AR cells of \cref{fig:opus_tier_table} train on a curriculum of twelve scenes on LIBERO and LIBERO-PRO and 47 on RoboLab.

\section{Resampling Decomposition}
\label{app:resampling}

This appendix specifies the selection rule and evaluation readings behind \cref{sec:finding_resampling}. The main comparison uses Lineage, the full-budget no-library Vanilla run, and best-of-3 short no-library attempts.

\paragraph{How a run is scored.} Each run is graded through automatic snapshots every five minutes, agent-initiated submissions, and a final grade of the policy on disk when the budget expires. Each grade evaluates ten VAL initial states, and selection retains the policy with the highest valid VAL score. The same within-run rule applies to all conditions. TEST is a disjoint band used only after selection and never guides policy development or library harvesting.

\paragraph{Two readings of the same runs.} The learning curves in \cref{fig:avg_perf_vs_time} show best-so-far VAL performance at a development-time cutoff. Each best-of-3 attempt receives one third of the displayed total time. \Cref{tab:t1_main} and \cref{sec:finding_resampling} instead report TEST scores of VAL-selected policies after the runs end, including their final grades. These readings differ in both evaluation band and which submissions are eligible. They should not be substituted for one another or interpreted as equal compute: independent restarts have separate submission allowances and can consume different numbers of model tokens and environment steps.

\paragraph{Library and session terms.} At a fixed current-task development budget, the margin over best-of-3 decomposes as
\begin{align}
\mathrm{Lineage}-\mathrm{BO3}
&=(\mathrm{Lineage}-\mathrm{Vanilla})
 +(\mathrm{Vanilla}-\mathrm{BO3}).
\end{align}
The first term estimates the benefit of inherited state under matched current-task resources; the experience used to construct the library remains part of the lineage's history. The second compares sustained iteration with independent shorter attempts. Both kinds of run can revise policies within a session, so this is a comparison of budget allocation rather than a control with all learning disabled. The LIBERO, LIBERO-PRO, and LIBERO-RIG best-of-3 attempts are told their 20-minute limit, which makes them submit earlier and more often than an untold attempt, while the robosuite and RoboLab attempts and the Lineage and Vanilla sessions are not told their budget. Caveats for RoboLab (a partly completed course and a corrected tool frame) are stated next to the results.

\paragraph{Selection and aggregation.} For a task $t$, let $v_i(t)$ be the best VAL score of short attempt $i$, and $q_i(t)$ its selected policy's TEST score. The reported best-of-3 TEST score is
\begin{align}
 i^*(t)&=\operatorname*{arg\,max}_{i\in\{1,2,3\}}v_i(t), &
 \mathrm{BO3}_{\mathrm{TEST}}(t)&=q_{i^*(t)}(t).
\end{align}
It is not the maximum of the three TEST scores. Family scores average over the specified tasks; repeated Lineage and Vanilla runs are summarized across independently developed policies or libraries.

\textbf{Selecting by validation overstates the winner for a model that submits often.} Best-of-3 keeps the attempt with the highest validation score, and that attempt is often the one that overfit the validation states. GPT-5.6-sol submits about twice as often as any other model, and its RoboLab winners lose 6.6 points from validation to test; Opus's lose none, and the other three lose 2 points or less. Even after this loss, GPT-5.6-sol still scores higher with best-of-3 than with one session.

\textbf{With three times the budget, resampling full-length runs pays for every model.} Full-length runs vary a lot. The best of three full-budget Vanilla repetitions, which costs three times the budget, scores 17 points above a single repetition on RoboLab (15 when chosen by validation score) and 5 on robosuite. So when compute allows, run several full-length sessions in parallel rather than splitting one budget into short ones. At a fixed budget, whether to split depends on the model's time to first success.

\section{Real-World Harness Details}
\label{app:real_world_harness}

\paragraph{Perception and control tools.}
The toolbox combines pretrained perception models with deterministic geometry and control. SAM3~\citep{carion2026sam} provides text-prompted instance segmentation, returning masks, bounding boxes, and confidence scores. A vision-language model tool answers agent-defined questions about one or more images, optionally in a specified JSON format. It is instructed to describe visible properties (object identity, text, orientation, spatial relations) and to express uncertainty when observations are ambiguous. Its role is restricted to perception; task planning and action selection remain with the agent or generated policy. These perceptual estimates are separate from the external evaluator's judgments.

Geometry tools map image observations into robot coordinates using calibrated projection, ray--plane intersection, and valid depth measurements. They support plane fitting, object localization, footprint and orientation estimation, and rectified top-down views. Per-camera metadata identifies whether depth is available and metrically valid. Geometry tools remain available throughout a run, while learned model tools are enabled according to the run configuration.

Calibrated inverse-kinematics and servo interfaces support Cartesian motion, gripper control, and coordinated arm movement subject to workspace, joint-rate, and inter-arm separation constraints. Agents may query the motion planner without executing a command and replay recorded images to develop perception routines between physical attempts.

\section{Physical Campaign Protocol}
\label{app:real_world_protocol}

The physical campaign comprises five tasks and evaluates three coding agents, Claude Fable~5.1 and Claude Opus~5 through Claude Code and GPT-6-astra through Codex, all at maximum reasoning effort, that revise executable policy bundles while keeping model weights fixed. Each lineage begins with an empty library, without a seed policy or simulation reference solution. Subsequent tasks inherit the preceding policy bundle and library. The library records reusable code, procedures, and textual memories together with their provenance. Code packages are harvested from successful executions supported by valid evidence, while procedures and memories may additionally retain lessons from failed attempts.

Each task receives a 12-hour wall-clock budget covering policy revision and physical interaction. Compared agents share the task order (Banana into Bowl, Push-T, Cup Upright, Block Stacking, Spell SEAR), hardware interfaces, available tools, and evaluator configuration, and run on a frozen harness package whose file set is checked by checksum on every rig; a lineage may move between rigs from one task to the next. A task may terminate early after ten consecutive scored forward-task successes. Early termination additionally requires successful evidence from both arms, and reset attempts, diagnostic probes, and outcomes marked unknown or invalid are excluded from the early-stopping window. Block Stacking has no early stop: the tower persists between attempts, an attempt succeeds when the free-standing column ends taller than it started, and toppling the tower is a separate reset task.

\paragraph{Offline evaluation.} After the CaP campaign, each task's policy bundle is frozen and evaluated on operator-staged scenes. The frozen evaluator scores each episode and the operator confirms or amends the verdict. \Cref{app:real_world_results} specifies episode counts, scoring rules, and which practice segments enter the totals. Every offline episode we checked ran on the rig on which its task was practiced.

\section{Physical Campaign: Detailed Practice and Evaluation}
\label{app:real_world_results}

\paragraph{Reading the detailed tables.} \Cref{tab:real_world,tab:real_world_aap} separate practice from evaluation. CaP evaluates frozen policy bundles; AaP keeps the LLM in the control loop. Each task inherits the preceding workspace and has a 12-hour practice budget. Bold times mark the streak: ten consecutive scored forward successes with both arms represented. Other entries show the budget, followed by forward successes/attempts. CaP counts scored forward attempts; AaP counts all recorded forward evaluations, including unresolved verdicts and requests closed at the budget limit. For Block Stacking, a practice attempt is one block placement, successful when the column ends taller than it started; there is no early stop.

Evaluation scores in these tables are out of ten. Non-stacking scores are ten times the mean outcome (success: one; partial: one half; failure: zero). Block Stacking scores average the maximum cube count attained in each episode, capped at ten. The final row averages all five tasks before rounding. CaP uses ten scenes per non-stacking task and ten tower episodes per model. When a scene is restaged, only its final evaluation is counted. AaP uses five scenes or tower episodes per task.

\paragraph{Resource accounting.} Total attempts include forward attempts, resets, and probes. Unscored probes exclude probes submitted for a judge verdict, although those remain in total attempts. All totals use the reported lineage segments: when a task has multiple launches, the selected segment is the one inherited by the next task, or the latest launch for the final task. CaP hardware time sums those segments; AaP hardware time sums their launch-to-end durations, rounded to the nearest minute.

\begin{table}[t]
\centering
\caption{CaP practice and evaluation details. Scores are out of ten. Practice entries give time to the ten-success streak (bold) or the 12-hour budget, then scored forward successes/attempts. Bold mean score marks the best model. Counting conventions are defined in the accompanying text.}
\label{tab:real_world}
\small
\begin{tabular}{lcccccc}
\toprule
& \multicolumn{2}{c}{Fable~5.1} & \multicolumn{2}{c}{Opus~5} & \multicolumn{2}{c}{GPT-6-astra} \\
\cmidrule(lr){2-3} \cmidrule(lr){4-5} \cmidrule(lr){6-7}
Task, in lineage order & practice & score & practice & score & practice & score \\
\midrule
Banana into Bowl & \textbf{1h46}\;\;10/10 & 7.00 & \textbf{2h50}\;\;13/14 & 8.00 & 12h\;\;27/45 & 5.00 \\
Push-T           & \textbf{6h34}\;\;26/29 & 10.00 & 12h\;\;8/58 & 4.00 & 12h\;\;20/23 & 6.00 \\
Cup Upright      & \textbf{4h18}\;\;14/19 & 7.00 & \textbf{9h33}\;\;68/105 & 4.00 & 12h\;\;27/46 & 5.00 \\
Block Stacking      & 12h\;\;88/106 & 7.10 & 12h\;\;15/210 & 1.80 & 12h\;\;3/19 & 1.70 \\
Spell SEAR       & \textbf{4h37}\;\;11/18 & 4.00 & 12h\;\;1/45 & 0.00 & 12h\;\;9/15 & 3.00 \\
\midrule
Hardware time & 29h27 & & 48h33 & & 60h13 & \\
Total attempts & 355 & & 672 & & 597 & \\
Unscored probes & 8 & & 72 & & 358 & \\
\midrule
Mean score (out of 10) & & \textbf{7.02} & & 3.56 & & 4.14 \\
\bottomrule
\end{tabular}
\end{table}

\begin{table}[t]
\centering
\caption{AaP practice and evaluation details. Scores are out of ten. Practice entries give time to the ten-success streak (bold) or the 12-hour budget, then successful/recorded forward evaluations. Bold mean score marks the best model. No Opus~5 AaP evaluation is available.}
\label{tab:real_world_aap}
\small
\begin{tabular}{lcccc}
\toprule
& \multicolumn{2}{c}{Fable~5.1} & \multicolumn{2}{c}{GPT-6-astra} \\
\cmidrule(lr){2-3} \cmidrule(lr){4-5}
Task, in lineage order & practice & score & practice & score \\
\midrule
Banana into Bowl & \textbf{2h15}\;\;10/10 & 10.00 & \textbf{4h23}\;\;10/10 & 8.00 \\
Push-T & \textbf{7h03}\;\;17/19 & 10.00 & \textbf{9h56}\;\;10/10 & 10.00 \\
Cup Upright & \textbf{2h43}\;\;11/13 & 2.00 & \textbf{8h40}\;\;10/10 & 10.00 \\
Block Stacking & 12h\;\;33/62 & 5.20 & 12h\;\;14/23 & 6.00 \\
Spell SEAR & \textbf{3h36}\;\;12/15 & 6.00 & 12h\;\;11/13 & 10.00 \\
\midrule
Hardware time & 27h41 & & 46h58 & \\
Total attempts & 224 & & 133 & \\
Unscored probes & 16 & & 50 & \\
\midrule
Mean score (out of 10) & & 6.64 & & \textbf{8.80} \\
\bottomrule
\end{tabular}
\end{table}

\begin{table}[t]
\centering
\caption{\textbf{A model in the loop makes every episode several times longer.} Rig time per TEST evaluation episode under the two execution modes: median minutes from the evaluation request to the verdict (number of episodes in parentheses). CaP is the frozen policy bundle of the offline evaluation; AaP keeps the model in the loop during evaluation. The last column is the ratio of the two medians over all episodes of the task.}
\label{tab:aap_latency}
\small
\setlength{\tabcolsep}{4pt}
\begin{tabular}{lccccc c}
\toprule
& \multicolumn{3}{c}{CaP, frozen policy} & \multicolumn{2}{c}{AaP, model in the loop} & \\
\cmidrule(lr){2-4} \cmidrule(lr){5-6}
Task & Fable~5.1 & Opus~5 & GPT-6-astra & Fable~5.1 & GPT-6-astra & AaP / CaP \\
\midrule
Banana into Bowl & 1.3 (12) & 1.4 (10) & 1.3 (10) & 7.8 (5) & 12.8 (5) & 6.7 \\
Push-T           & 5.0 (12) & 4.1 (10) & 8.3 (10) & 12.8 (5) & 38.1 (6) & 3.7 \\
Cup Upright      & 1.6 (10) & 1.8 (10) & 1.6 (10) & 10.5 (5) & 8.9 (5) & 5.6 \\
Block Stacking   & 9.7 (11) & 6.9 (10) & 5.5 (10) & 15.2 (5) & 28.7 (5) & 3.4 \\
Spell SEAR       & 5.2 (10) & 3.8 (10) & 8.1 (10) & 7.4 (6) & 23.5 (5) & 2.7 \\
\midrule
All episodes, median & \multicolumn{3}{c}{3.8} & \multicolumn{2}{c}{12.8} & 3.4 \\
\bottomrule
\end{tabular}
\end{table}

\section{Real-World Task Specifications}
\label{app:real_world_spec}
\paragraph{Hardware setup.}
The platform comprises two i2rt YAM Ultra manipulators, each with six revolute joints and a LINEAR\_4310 parallel-jaw gripper. The arms are fixed side by side along one edge of the table, with approximately 0.51~m between their bases, and face a shared workspace.

Three RealSense RGB-D cameras provide complementary views. A fixed D435 overview camera is mounted approximately midway between the arm bases, 0.86~m above the tabletop, and faces into the workspace. Its optical axis points approximately $20^\circ$ away from vertically downward, equivalent to $70^\circ$ below horizontal. A D405 is rigidly mounted on each wrist. In the local flange frame, each wrist camera is offset approximately 70~mm laterally from the flange axis and 70--76~mm along it, with its optical axis tilted inward by approximately $25^\circ$ relative to the flange's forward ($+z$) axis. Its orientation relative to the table therefore changes with the arm pose.

All three cameras operate at 30~frames/s. The D435 records color at $1920\times1080$ and depth at $848\times480$; each D405 records both color and depth at $848\times480$. Depth is aligned to the corresponding color view. The overview-camera pose, wrist camera-to-flange transforms, table plane, and relative arm-base transform define the mapping from image observations to the shared workspace; wrist-camera poses are updated using forward kinematics.

\paragraph{Objects and task criteria.}
Initial object positions and orientations vary within the workspace. The five tasks use the following objects and completion criteria.
\begin{itemize}
    \item \textbf{Banana into Bowl.} Yellow foam bananas are approximately 220~mm long. The red bowl is approximately 55~mm high, with a rim diameter of approximately 164~mm and a base diameter of 93~mm. Success requires placing and releasing one banana inside the initially empty bowl. The banana may protrude beyond the rim, but must rest inside the bowl rather than on its rim or against its exterior. Other objects must remain undisturbed; incidental bowl displacement of up to approximately one bowl radius is permitted.
    \item \textbf{Push-T.} The green 3D-printed block has a $200\times200$~mm footprint and a height of 40~mm. Its crossbar is $200\times50$~mm, with a centered $150\times50$~mm stem extending from it.\footnote{CAD model for the T-shaped block: \url{https://cad.onshape.com/documents/f1140134e38f6ed6902648d5/w/a78cf81827600e4ff4058d03/e/f35f57fb7589f72e05c76caf}.} A matching T-shaped outline is drawn on white paper. Success requires non-prehensile pushing that aligns both the position and orientation of the block with the outline; partial overlap or visible rotational misalignment is insufficient. Grasping is permitted only to recover a block that has moved completely off the paper, after which pushing must resume.
    \item \textbf{Cup Upright.} White paper cups have a truncated-cone shape, with a height of 106~mm, a mouth diameter of 89~mm, and a base diameter of 62~mm. Cup count and initial orientations vary. Success requires turning one lying or inverted cup upright, opening upward, and releasing it without disturbing cups that were already upright. All cups must remain on the table.
    \item \textbf{Block Stacking.} Ten identical wooden cubes, each 38.3~mm on a side, are initially scattered on the table. The objective is to build a single tower as high as possible. The tower must stand without support from either arm, a gripper, or another object. The score is the maximum standing tower height reached during the episode, reported as a layer count alongside its height above the table; a subsequent collapse does not erase the attained height.
    \item \textbf{Spell SEAR.} Approximately twelve wooden cubes, each about 40~mm on a side, carry the same letter on all six faces and include duplicate and distractor letters. Success requires exactly one cube per requested letter in a single unstacked row within a central $450\times220$~mm region. Letters must be visible, in the requested order, and upright within approximately $30^\circ$ of the reading orientation; unused cubes must lie outside this region. During practice, the requested left-to-right target may be SEAR or RAES, corresponding to reading \modelname{} from left to right or right to left in the displayed overview; \cref{fig:real_world_evolution_c} labels the direction of each example. Final evaluation uses \modelname{} read from left to right by a viewer facing the table from the arm bases.
\end{itemize}

\paragraph{Human scoring.}
Human evaluators review the recorded outcomes against these criteria. For the four non-stacking tasks, full completion receives 1, partial completion receives 0.5, and failure receives 0. A banana resting on the bowl rim is an example of partial completion. Invalid or incorrectly staged trials are repeated. For Block Stacking, evaluators verify the maximum standing layer count from the recordings, with depth measurements providing the corresponding height.

\section{Physical Campaign: Where Frozen Policies Stop}
\label{app:real_world_failures}

This appendix details the audit summarized in \cref{sec:finding_real_world_limits}. For every failed or partial evaluation unit we located, in the evaluation record and the frozen policy, the check that ended progress, and classified it by what stopped the robot: a physical impossibility; a refusal by the harness, namely its safety guard, motion planner, joint limits, or step budget; or a limit the agent wrote itself, such as a reach cap, tolerance, clearance margin, grasp restriction, disabled step, or stopping rule. A scene counts as physically possible when another agent, the same agent in the other mode, or the platform's measured grasp reach shows that the motion can be done. The harness enforces joint limits, a workspace box per arm, a 0.25~m keep-out region around the other arm's base, and a speed limit near the table. It does not model objects on the table, and its planner checks straight-line paths locally, so a refusal does not establish that a target is unreachable. \Cref{tab:real_world_failures} gives the counts. The units comprise Fable's and Opus's frozen-policy episodes, block-placement calls from GPT-6-astra's frozen Block Stacking policy and its other frozen-policy episodes, and Fable's live AaP episodes. Because block-placement calls and whole-task episodes are different units, the counts describe the audited failures rather than a common episode-level failure rate across models. Camera images were not part of this audit; statements about contact between objects rest on logged motion and stopping heights.

\textbf{Further audited cases (\cref{sec:finding_real_world_limits}).} In Fable's Spell SEAR run, a letter-straightening step turned a letter the wrong way on its only practice use; the agent disabled the step after that single attempt and left it disabled, and the frozen policy leaves a crooked letter in two evaluation scenes. Its code for clearing the middle of the table runs in 6 of 10 evaluation scenes but never ran during practice. Opus's Block Stacking policy measured cubes at 37--39\,mm but in its ninth hour fixed the size at 28--33\,mm to match the judge's nominal value, and it interprets normal arm tracking error as contact, raising release heights by 8--19\,mm; only 1 of 11 cubes placed on a column remains there in the last six evaluation episodes, against 7 of 15 in the first four. GPT-6-astra's Cup Upright policy lays the cup back down and stops after a refused turn with about two thirds of its steps unused, and Opus's Push-T policy stops 80 to $170^\circ$ from the goal with 8,000 to 11,000 of its 15,000 steps left. Most stops not attributable to the policy's own rules were harness refusals of poses those policies selected.

\begin{table}[h]
\centering
\caption{What stopped each failed or partial evaluation unit. A unit is an episode, except for GPT-6-astra's Block Stacking, where each block-placement call counts. \emph{Harness refusal}: all three of Fable's arise from poses its own code selected when a working alternative existed; Opus's are one refusal it steered into and two grasps that missed on the left arm's provisional calibration; most of GPT-6-astra's refuse paths its own rules chose, and 42 of its 47 units trace to its own choices.}
\label{tab:real_world_failures}
\small
\begin{tabular}{llccccc}
\toprule
Agent & Mode & Units & Impossible & Harness refusal & Own limit & Other \\
\midrule
Fable~5.1 & CaP and AaP & 34 & 0 & 3 & 29 & 2 \\
Opus~5 & CaP & 35 & 0 & 3 & 32 & 0 \\
GPT-6-astra & CaP & 47 & 0 & 20 & 26 & 1 \\
\bottomrule
\end{tabular}
\end{table}

\textbf{Banana into Bowl.} No frozen failure dropped a banana, set it beside the bowl, or knocked the bowl; every full failure is a move refused or abandoned, and each policy tries its first plan once. Fable grasps with the gripper tilted by at most $15^\circ$, so a banana 0.60~m out was refused although tilted grasps reach it. GPT-6-astra quits when both of its two camera poses are unavailable, having removed an earlier fallback within minutes of its first launch without any incident, and lifts or carries with a single wrist angle. In one scene the banana and bowl are 0.76~m apart and each arm reaches only one of them. No frozen policy passes the banana between arms; both live agents did, though with more motion steps than the frozen budget allows. Opus releases over the bowl center, 30~mm above the rim, wherever it holds the banana. It wrote this rule before its first robot attempt, and two bananas came to rest across the rim.

\textbf{Push-T.} The staged scenes often require a large turn. Fable pushes up to 10.5~cm, turning the block 13 to $37^\circ$ per push, and re-measures after every push. Opus shortened its pushes to 35~mm in its fourth hour, after pushes appeared to spin the block by 65 to $171^\circ$ at a time when its model of the tool--block contact was wrong and some headings were misread. It corrected the model at hour 7.5 but never revisited the push length. Its left-arm pushes delivered a median 8\% of the predicted turn against 64\% for the right arm, and its two stopping rules, written for the final centimetres, fired 80 to $170^\circ$ from the goal. The $16^\circ$ cap on predicted turn per push in its code was never reached, because a 35~mm push cannot be predicted to turn the block by more than about $14^\circ$. GPT-6-astra rejected correct outline fits scoring 7.8 and 8.6~mm against a fixed 7~mm limit, unchanged since its first hour, and ended both attempts without touching the block. Elsewhere its finishing tolerance, tighter than the judge usually applies, sent it into extra pushes that worsened the pose, and each of its pushes costs about 835 motion steps because it lifts to a camera pose after every push.

\textbf{Cup Upright.} Fable's only method pinches the rim through the cup's opening, which works only when the opening faces an arm. It chose the method in its first 20 minutes, following a suggestion in the task card, and never tried the body grasp that both other agents used on the same scenes. All three of its frozen failures are refusals with no motion, and one was caused by a 0.20~m clearance from the parked arm, inherited from Push-T and never tested, where the harness requires 0.12~m. Opus's near-horizontal body grip left the cup upright in 2 of 9 stand-ups, with no check that the grip holds. GPT-6-astra grips the body and turns the cup in the air. With the cup held 0.40--0.435~m from the arm base, 17 of 18 practice turns succeeded; at 0.454~m or more, all 5 were refused at the wrist. Its rule for choosing where to turn was revised twice, each time to fix the latest incident, and in one evaluation scene it picked the farther spot. After any refused turn it lays the cup down and stops. Its live agent instead turned the cup flat on the table and re-gripped it.

\textbf{Block Stacking.} Fable places cubes within a median 2.9~mm, and the error does not grow with height. Its stall at about nine cubes looks like a wrist limit, but follows from its own choice to square each cube with the one below: from layer seven it turned the gripper 28 to $41^\circ$ off the line from the arm base, and at that height the planner accepted 17 of 17 moves with the gripper within $10^\circ$ of the line and 4 of 34 when it was 30 to $43^\circ$ off. In practice it rarely built above six cubes after its record in hour three. Opus's placement code carried a height correction between calls that counted the arm's normal 3~mm tracking error as contact. The correction grew to 17--20~mm; cubes that stayed were released a median 0.2~mm below the true top and cubes that fell a median 6.5~mm above it, and the cube size it had fixed at 28--33~mm plausibly added to the error. When GPT-6-astra released a cube, it almost always stayed, 6 of 7 times, but most calls stopped before releasing. A low carrying path that keeps 170~mm from the tower and about 92~mm from every other cube, whatever that cube's height, blocked 8 calls, and replaying its planner with a height rule it wrote later finds a path in 7 of them. A $6.1^\circ$ heading tolerance, 200-step descent allowances, and a camera-coverage threshold ended 8 more, and harness reach refusals on approaches it had chosen ended 3. It spent 5.7 of its 12 hours on a narrow-grasp method for crowded cubes, which produced one practice success, and never considered moving cubes apart.

\textbf{Spell SEAR.} No frozen policy put a wrong letter in the row. Staged scenes 6 to 10 add a crowded middle, letters lying across the target row, and required letters 0.53--0.58~m from the nearer arm base, and no frozen policy solved scenes 6 to 9. Besides the crooked and distant letters described in \cref{sec:finding_real_world_limits}, Fable twice lowered a letter that, judging from its logged stopping height, came to rest on a required letter still standing near its slot, because its clearing code removes only unneeded blocks. GPT-6-astra's frozen policy refuses any placement whose fully open hand, modeled as an $86\times144$~mm rectangle, would touch another block. The rectangle only grew during practice, its notes explicitly refuse to shrink it, and it caused 13 refusals, while its step budget ran out in 5 of 10 scenes. Its live agent reordered letters, cleared distractors, and tilted the tool for distant letters, and succeeded on every staged scene it was given. Every evaluation scene requested SEAR, so reading direction was not tested in evaluation.

\textbf{Earlier work is not carried forward.} Fable's counted Spell SEAR segment starts from its Block Stacking workspace. An earlier launch of the same task, not the one counted, had already picked distant letters with a tilted gripper and re-checked each letter's angle after placing it, but inheritance follows the task order, so the counted segment never saw that work. More generally, only 1 to 25\% of a bundle's substantive code lines recur in the same agent's bundle for the previous task. The exception is Block Stacking to Spell SEAR, where Fable keeps about 40\% and GPT-6-astra about 47\% by reusing their cube-handling code for letter blocks, while Opus keeps 3 to 6\%. No frozen policy reads the persistent library.

\textbf{How the sessions are spent.} Across the five tasks, GPT-6-astra made 8,023 tool calls, about 133 per hour, against 2,660 for Fable and 3,528 for Opus, and it read harness documentation and source 911 times, against 185 and 163. After changing code, Fable ran an offline check before the next hardware attempt in 62\% of cases, GPT-6-astra in 47\%, and Opus in 30\%; after a failure, Fable opened an additional camera image before retrying in 44\% of cases, against 7\% for Opus. Opus never ran the Block Stacking reset task and cleared the spelled word once in 12 hours, whereas Fable ran a scored reverse attempt after almost every forward attempt. For every agent the robot was executing an attempt for only 34--42\% of session time.

\section{Physical Campaign: Practice Behavior in Detail}
\label{app:real_world_practice}
This section expands \cref{sec:sb2_hardware}.

\begin{figure}[tbp]
    \centering
    \includegraphics[width=0.92\linewidth]{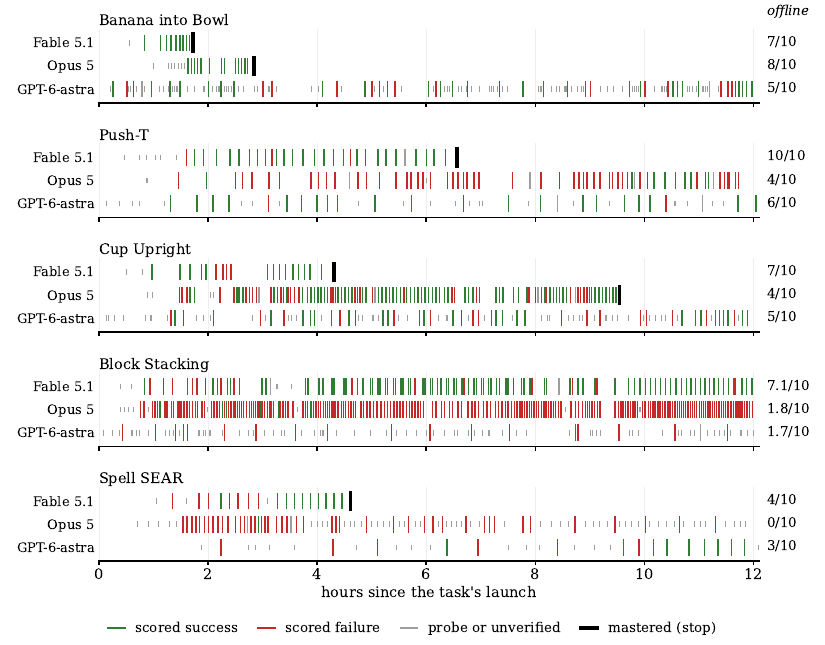}
    \vspace{-10pt}
    \caption{\textbf{Every physical practice attempt against time.} Every forward attempt of the physical CaP campaign against hours since the task's launch: tall ticks are scored attempts (green success, red failure), short grey ticks unscored attempts, and the black bar the tenth consecutive scored success, where the run stops; Block Stacking has no early stop. The right margin gives each task's held-out score out of ten.}
    \label{fig:real_world_timeline}
\end{figure}

\paragraph{Streaks.}\label{sec:finding_real_world_practice} Fable reaches the ten-success streak on all four non-stacking tasks, using 29.5 hours of hardware time across the five tasks; Opus reaches it on two, and GPT-6-astra on none in 60 hours (\cref{tab:real_world}). segments that reached the streak average 67\% on staged scenes, against 38\% for the six non-stacking segments that ran out of budget. \Cref{fig:real_world_timeline} shows every practice attempt.

\textbf{Probes and scene variation.} GPT-6-astra spends 358 of its 597 attempts, 60\%, on unscored probes, compared with 8 of 355 for Fable. Opus uses 72 probes in total, 51 of them on Spell SEAR. On that task, GPT-6-astra spends five hours practicing pick and place through the clearing task before attempting the Spell SEAR task; its last seven scored spellings then succeed, three short of the streak. Many of GPT-6-astra's probes deliberately change the next starting state: with its jaws closed it pushes the empty bowl 20--40\,mm to a new position; it sets up a T rotated by $180^\circ$; and it performs 48 of its 52 cup resets as probes to place cups farther out. These actions follow an instruction to succeed across layouts. At least three of its success streaks break on the scored attempt following such a probe.

\textbf{Reset states versus staged states.} In every one of Fable's practice spellings, its scrambling reset leaves at most one block in the middle of the table and no required letter farther than 0.48\,m from an arm base; in evaluation, 8 of 10 scenes have blocks in the middle and 2 have a required letter 0.53--0.58\,m away. Only 10 of Opus's 64 practice Push-T starts require a turn of more than $90^\circ$, compared with 6 of the 10 staged scenes. GPT-6-astra's reset always leaves the T within 175\,mm of its outline, whereas the staged starts are 158--269\,mm away. Fable's cup reset leaves every cup with its opening toward an arm. In Fable's Cup Upright run a dropped cup came to rest facing away from both arms; the agent continued without resolving that configuration, 14 refused resets did not count toward the streak, and it reached the streak with 7.7 of its 12 hours unused.

\textbf{Live execution.} Under agent-as-policy, Fable's code grows from 202 lines after Banana into Bowl to 3,576 lines in 30 files after Spell SEAR, with at least 15 files issuing robot commands; these scripts account for 99\% of its 33,532 practice commands and 98\% of its 5,916 evaluation commands, and on Block Stacking and Spell SEAR background loops open and close most practice attempts without a separate model decision. GPT-6-astra issues each of its 7,537 commands through a separate model call, requests about one image per arm motion against Fable's 0.14, and writes ten small helpers, none of which moves the robot. Fable's scripted Cup Upright routine closes on the cup's paper wall in all 18 visible practice grasps and in none of 23 evaluation grasps; its single evaluation success comes after switching to a body-squeeze routine written during practice.

\textbf{Cost.} GPT-6-astra issues its first robot attempt 3 to 8 minutes after each launch, while Fable and Opus first spend 23 to 53 minutes reading and writing code. Per scored forward success, practice costs about \$3.7 in API spend for Fable, \$11 for Opus, and \$18 for GPT-6-astra. For the Cup Upright segment the recorded API costs are about \$96, \$254, and \$368, respectively.

\textbf{Losing objects.} Opus's Spell SEAR policy knocks the S off the table in the fourth hour, the E in the sixth, and the R in the ninth. From hour 4.4 its own scene description reports that the word cannot be completed because the S is missing. Nevertheless it spends its last 7.5 hours, and 62 of its 101 attempts, continuing the task; its memory records the loss of the S and E, but further code revision cannot replace the missing letters. Practice ends with 1 success in 45 scored spelling attempts, and the frozen policy scores 0 of 10 on the staged scenes. In the third hour of Opus's Cup Upright run an attempt knocks a lying cup off the table; the next 169 attempts, including the ten-success streak, operate on the remaining two-cup scene.

\textbf{Block Stacking.} The task asks for the tallest tower, and the run's score is the maximum height reached, but the feedback for each placement is a binary success whenever the column ends taller than it started. Fable reaches 310\,mm, about eight cubes, in its third hour, then cycles through building towers of five or six cubes, toppling them, and rebuilding (\cref{fig:real_world_evolution_b}). Placements that increase the height of these shorter towers continue to earn successful verdicts even though the record does not improve. Placement success and progress toward a higher tower must therefore be reported separately; the pattern does not by itself establish that the feedback caused the rebuilding.

\section{Physical Campaign: Attempt Galleries}
\label{app:real_world_gallery}

\begin{figure}[p]
    \centering
    \includegraphics[width=\linewidth]{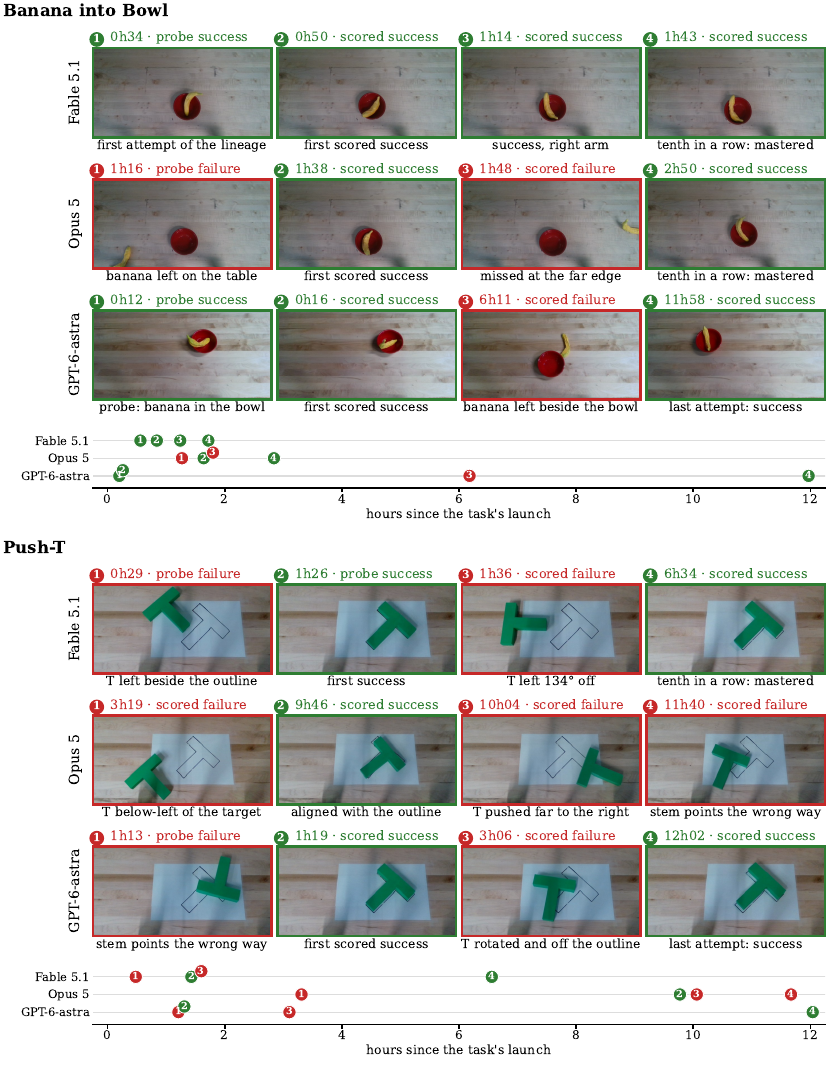}
    \vspace{-14pt}
    \caption{Self-evolution on hardware, every agent on every task (continued in \cref{fig:real_world_evolution_b,fig:real_world_evolution_c}). Each block is one task, with one row per agent and four attempts per row, numbered in session order: the overview camera at the end of the attempt, with the elapsed session time, the attempt kind, and the judge's verdict (green success, red failure). The axis under each block places the four numbered attempts of each agent at their session times, so the rows read as time lapses of 12-hour sessions. For Opus~5 on Banana into Bowl and Fable~5.1 on the other tasks the attempts are chosen to show how the session developed; the other rows generally show the first recorded attempt, the first scored success, a later scored failure, and the last recorded scored attempt. The displayed Opus and GPT-6-astra Push-T failures, GPT-6-astra Spell SEAR failure, and Fable Cup Upright examples are selected for visually unambiguous outcomes rather than proximity to a fixed session time. Each lineage starts from an empty library on Banana into Bowl and inherits its own earlier tasks' workspace thereafter.}
    \label{fig:real_world_evolution}
\end{figure}

\begin{figure}[p]
    \centering
    \includegraphics[width=\linewidth]{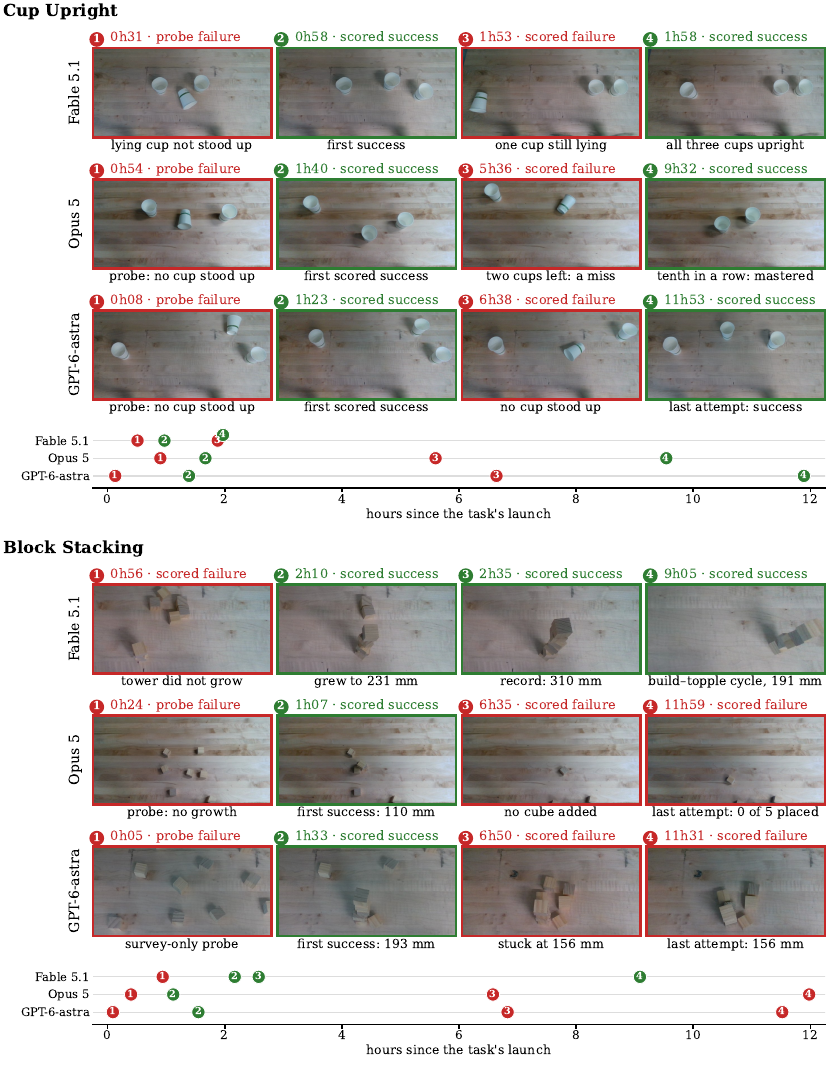}
    \vspace{-14pt}
    \caption{Self-evolution on hardware, continued from \cref{fig:real_world_evolution}: Cup Upright and Block Stacking. Opus's Cup Upright attempts after hour 2.9 practice a two-cup scene, because one cup left the table. On Block Stacking, Fable's last tile is the build-and-topple cycle it settled into after its 310~mm record; Opus's last stacking success comes at hour 5.3 and it has only single cubes after hour 6.3, and GPT-6-astra's tower sits at 156~mm.}
    \label{fig:real_world_evolution_b}
\end{figure}

\begin{figure}[p]
    \centering
    \includegraphics[width=\linewidth]{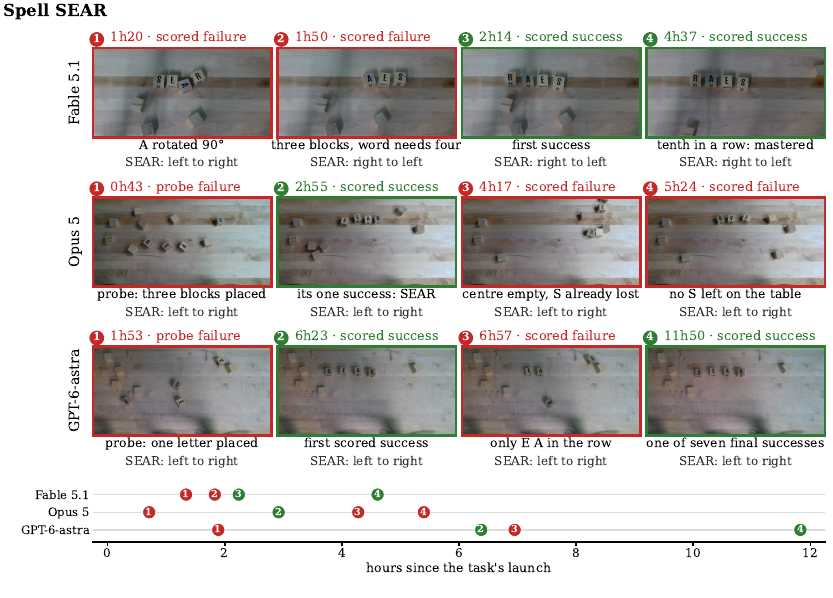}
    \vspace{-14pt}
    \caption{Self-evolution on hardware, continued from \cref{fig:real_world_evolution_b}: Spell SEAR. Each tile states the instructed reading direction for \modelname{} in the displayed camera view. Left-to-right spells SEAR; right-to-left appears as RAES from camera left to right. Fable tiles 2--4 request RAES, while its first tile and all displayed Opus and GPT-6-astra tiles request SEAR; RAES is a valid forward spelling target, not a failed SEAR arrangement. Opus's one success comes at hour 2.9; by hour 4.3 its S has left the table, and its last recorded scored attempt, at hour 5.4, reads a table with no S. GPT-6-astra's first success comes at hour 6.4, and its last seven scored spellings all succeed.}
    \label{fig:real_world_evolution_c}
\end{figure}
\FloatBarrier

\section{LIBERO-RIG: Design and Protocol}
\label{app:rig}

This section summarizes the design of LIBERO-RIG~\citep{liberorig}, the benchmark that supplies the hardware-fault condition of tier T2 (\cref{sec:tiers}) and one of the five simulated families. The companion paper also evaluates nine end-to-end VLA models under the same protocol, which we do not reproduce here.

\paragraph{What LIBERO-RIG is.} LIBERO-RIG measures how robust a policy is to perturbations of the physical parameters of the robot it runs on, with the task held fixed. It keeps the instruction, the object identities, the success predicate, and the nominal initial-state identifiers of every LIBERO~\citep{liu2023libero} task, and changes only the rig, the robot hardware and the way it is set up, one axis at a time, by a prescribed amount. The rig is parameterized by the extrinsics and intrinsics of the scene and wrist cameras, the calibration between camera and arm, the gain and rate of the controller, the noise in the arm's reported state, the base pose of the arm, the table height, and the geometry of the end-effector pads. Formally, the rig configuration enters the episode in three places: the sensing map, through which the policy observes the world (camera mounting and optics, proprioception noise); the actuation map, through which its commands reach the world (camera-to-arm calibration, gain); and the physical arrangement of robot, table, and end effector (base pose, table height, pad geometry). A shared initial-state identifier therefore fixes the task's start, not an identical geometry or an identical first observation. The base benchmark is LIBERO's four ten-task suites, LIBERO-Spatial, LIBERO-Object, LIBERO-Goal, and LIBERO-10, simulated in MuJoCo~\citep{todorov2012mujoco} through robosuite~\citep{zhu2020robosuite}, each task with 50 fixed initial states, a scene camera, a wrist camera, proprioception, and a nominal control rate of 20 Hz.

\paragraph{Axes and fault types.} \Cref{tab:rig_axes} lists the ten axes used in this paper. The organizing idea is the type column, which classifies each fault by its direct effect on the camera images at a matched arm and object configuration. An observable fault changes at least one camera image of that configuration, as with a moved camera or a shifted robot base. A latent fault changes no camera image; it shows only in a non-visual input or in how the arm responds to a command, as with a rotated calibration or noise on the reported arm pose. Gripper-pad geometry sits between the two, since the pads occupy a few pixels of the wrist view. The split matters for mitigation: image augmentation targets what a policy sees, so by itself it does not address a fault that leaves the image of a given state unchanged. Each axis is a small, exactly specified transform of the simulator model or of the command stream. Action-frame rotation, for example, rotates the commanded translation and rotation about the world vertical before the controller's scaling and clipping, a controlled proxy for one rotational component of a calibration error.

\begin{table}[h]
\centering
\caption{The ten LIBERO-RIG perturbation axes used in this paper. LIBERO-RIG also defines a control-frequency axis, the control loop running at a rate other than the trained 20\,Hz; the code-as-policy harness runs the rig at the nominal rate and cannot apply that fault, so the axis is excluded from every result here. Type: O observable (at least one camera image changes at a matched configuration), L latent (no camera image changes), W weakly observable (visible only in the wrist view at close range). The last column gives the range from the mildest to the severest level.}
\label{tab:rig_axes}
\footnotesize
\begin{tabular}{@{}l c p{5.6cm} p{3.1cm}@{}}
\toprule
Axis & Type & What physically changes & Levels 1 to 5 \\
\midrule
camera orientation & O & the scene camera's mount is twisted; its position does not move & 2 to 25 degrees \\
camera position & O & the scene camera is remounted elsewhere around the workspace, still aimed at it & 5 to 50 degrees of arc \\
camera field of view & O & a different lens: the vertical field of view changes from the nominal 45 degrees & $\pm$2 to $\pm$18 degrees \\
wrist-camera mount & O & the wrist camera's bracket is twisted and slid along its mounting arm & 1 to 15 degrees, 2 to 30 mm \\
action-frame rotation & L & camera-to-arm miscalibration: every command executes in a frame rotated about vertical & 2 to 25 degrees \\
actuation gain error & L & commands execute too strong or too weak by a scale factor & $\pm$5 to $\pm$50 percent \\
proprioception noise & L & Gaussian noise on the sensed hand pose reported to the policy; the controller reads the true state & 2 to 50 mm, 0.5 to 8 degrees \\
robot-base pose shift & O & the floor mount is moved and rotated; the objects stay where they are & 1 cm, 2 degrees to 15 cm, 25 degrees \\
table-height change & O & the work surface is raised or lowered; objects move with it (two suites) & $\pm$1 to $\pm$10 cm \\
gripper-pad geometry & W & the end-effector pads are worn down or oversized; fingers and joints are untouched & $\pm$5 to $\pm$25 percent \\
\bottomrule
\end{tabular}
\end{table}

\paragraph{Severity levels and conditions.} Every axis has five settings, s1 to s5. For every axis these form a ladder from the smaller tested magnitudes to a rebuilt rig, with the first two levels at 2 and 5 degrees of camera twist and 1 and 2 cm of base displacement; the magnitudes are not claimed to represent measured real-world drift, and a given level does not carry the same physical severity across axes. This paper uses level 3, which we call moderate, throughout: the 18-task course, the T2 re-grades of practiced policies, and every LIBERO-RIG number in the main text are at the moderate level. The one exception is a severity sweep that also grades policies at levels 4 and 5, which we call severe (\cref{tab:rig_by_axis}, \cref{fig:rig_perturbation}, and \cref{tab:rig_level2}); each of these labels its level. A condition is one combination of suite, axis, and level, and spans all ten tasks of the suite. Table height applies only to the two suites whose arenas are height-adjustable, so the ten axes give 190 perturbed conditions plus four nominal ones. Every episode's parameters are pinned in checksummed manifests rather than sampled at run time, and every condition starts its tasks from the same fixed initial states as the unperturbed baseline, so every policy meets the same perturbation on the same start.

\paragraph{How a policy is measured.} The score is the raw success rate under each condition, read against the same policy's unperturbed rate on the same suite. Raw rates answer the deployment question directly, how often the policy works on this rig, and the gap to the nominal rate is the perturbation cost; a retention companion, the ratio of the two, allows a capability-adjusted reading. Each policy receives a per-axis profile rather than a single number, which tells a practitioner which rig changes it tolerates and which it does not. An episode starts from its fixed initial state, settles for ten control steps under a no-op command, and hands control to the policy; it ends as a success the moment LIBERO's predicate holds and as a failure when the budget runs out, with the step budget scaled to the control rate so that the simulated-time budget stays constant. The policy receives no fault identity, severity, rate, or calibration metadata, and any hidden state is reset between episodes. Two things must hold for a perturbed condition to measure the policy rather than the benchmark: the task must remain solvable, and its goal must remain visible. LIBERO-RIG tests both with fairness checks, replaying LIBERO's recorded demonstrations under the exact inverse of the perturbation where an inverse exists, following the demonstration's path with a closed-loop tracking witness where it does not, and auditing goal visibility along each perturbed camera. The evidence is reported as diagnostics and never used to remove a condition or an episode from the score.

\paragraph{Use in this paper.} The unit in \modelname{} is a cell, one task under one condition, rather than a full condition. The LIBERO-RIG course, which a Lineage run takes after LIBERO-PRO in place of RoboLab, is a development course of 18 cells at the moderate level: four or five tasks from each of the four suites, each paired with one perturbation axis, chosen so that all ten axes appear (\cref{tab:rig_course}); the control-frequency axis is excluded for the reason above. Each cell receives the family's per-task development budget (\cref{app:budgets}) and is graded on the DEV, VAL, and TEST bands of \cref{app:seed_details}, exactly as an unperturbed LIBERO task would be. The agent is given the stock LIBERO task text with only the task reference changed; nothing in the workspace names the fault. The course is run as three seed-matched runs per model (two for Gemini-3.6-flash, whose donor has two libraries), and the matched Vanilla control runs one 60-minute session per cell, three repetitions, with the same text; its host grades come every 15 minutes (the lineage runs' every 5), with agent-initiated submissions in both, so readings taken before 30 minutes favor the lineage runs and \cref{sec:finding_inheritance} quotes the 30-minute mark. The T2 results of \cref{sec:finding_perturbation} take policies practiced on unperturbed LIBERO and re-grade them, frozen, on all 40 tasks under every axis at the moderate level; the severity sweep adds the two severe levels, and \cref{tab:rig_by_axis} gives the per-axis values.

\begin{table}[h]
\centering
\caption{The 18-cell LIBERO-RIG development course, all at the moderate level. Task indices follow LIBERO's suite order.}
\label{tab:rig_course}
\small
\begin{tabular}{llll}
\toprule
Spatial & Object & Goal & LIBERO-10 \\
\midrule
00 action-frame rotation & 00 camera position & 00 gripper-pad geometry & 00 camera orientation \\
02 actuation gain & 02 camera field of view & 02 action-frame rotation & 02 camera position \\
& 04 wrist-camera mount & 04 actuation gain & 04 camera field of view \\
06 proprioception noise & 06 robot-base pose & & 06 wrist-camera mount \\
08 camera orientation & 08 proprioception noise & 08 table height & 08 robot-base pose \\
\bottomrule
\end{tabular}
\end{table}

\paragraph{Does a repair generalize to other faults?}
\label{app:rig_level2}
For every developed cell we take the best valid policy a model's adaptation runs wrote on it and run it, frozen, on every other cell of the same base task: the other axes at the moderate and the two severe levels for Opus and DeepSeek-v4-flash, at the moderate level for the other models. The reference is the same model's LIBERO-PRO-practiced policy for that task, run frozen on the identical cell (the T2 transfer of \cref{sec:finding_perturbation}), so the comparison asks whether practice under one hardware fault helps under a different one. Both sides of the comparison are validation grades, complete grades only, paired on identical cells; the score is the macro over the ten axes of the mean over cells (\cref{tab:rig_level2}). The gain is at most about ten points either way. It is positive for Opus, DeepSeek-v4-flash, and Gemini-3.6-flash, and larger at the severe levels than at the moderate level for Opus. It is negative for GPT-5.6-sol and Kimi-k3.

\begin{table}[h]
\centering
\caption{A policy repaired for one LIBERO-RIG fault, run on the same task under the other faults, against the pre-course LIBERO-PRO policy on the identical cells: validation success (\%), macro over the ten axes, and the number of paired cells.}
\label{tab:rig_level2}
\small
\begin{tabular}{llrccc}
\toprule
Model & Level & Pairs & Repaired policy & Pre-course policy & Gain \\
\midrule
Claude Opus 5 & moderate & 149 & 73.6 & 71.4 & $+2.2$ \\
 & severe (4) & 169 & 71.3 & 63.7 & $+7.6$ \\
 & severe (5) & 169 & 61.5 & 54.4 & $+7.1$ \\
GPT-5.6-sol & moderate & 142 & 66.9 & 71.0 & $-4.1$ \\
Kimi-k3 & moderate & 118 & 56.3 & 66.3 & $-10.0$ \\
DeepSeek-v4-flash & moderate & 142 & 51.1 & 41.0 & $+10.1$ \\
 & severe (4) & 160 & 48.7 & 38.5 & $+10.2$ \\
 & severe (5) & 160 & 40.6 & 33.5 & $+7.1$ \\
Gemini-3.6-flash & moderate & 128 & 68.3 & 65.0 & $+3.3$ \\
\bottomrule
\end{tabular}
\end{table}

\section{Model Portraits}
\label{app:portraits}

The five models differ in strategy as much as in score, and their failure modes follow from their strategies. The portraits below give the counts behind them. All counts are from the Lineage and Vanilla campaigns unless stated.

\textbf{Opus: measure first, build infrastructure, verify by proprioception.}
Opus makes the most tool calls (196 per two-hour RoboLab run against 101 on a one-hour LIBERO-PRO task, the one model whose effort scales with difficulty), runs the simulator locally, and views six to seven images per run. The looking is diagnostic, not reactive: image views lead to code edits 38\% of the time (baseline 24\%), concentrate in the second half of its exploration once the pipeline exists, and do not follow failure-flagged results. It compiles what it measured into geometry: only 20\% of its RoboLab lineage policies call SAM3 (73\% of no-library policies do), and its libraries hold height maps, rim fits, and footprint searches where other models hold SAM3 detector prompts. Its policies verify the grasp by gripper aperture (97\%), guard by remaining step budget (98\%), and retry (94\%), the idioms most strongly associated with success within Opus. It writes the most to the library, five to ten entries per step, phrased as sentences that carry a reason (``measure the hang, do not assume it''), and it is the only model that writes about the harness rather than the robot. It submits the least: half its runs are graded only by auto-evaluation. Its skills are almost uniformly useful (38 of 42 with three or more uses have positive lift, where a skill's lift is how much more often policies that import it succeed than policies that do not), and its best-of-3 winners lose nothing from validation to test.

\textbf{GPT-5.6-sol: fast, blind, and it overfits VAL.}
GPT-5.6-sol never reads an image by itself; it saves frames to disk and reasons about them numerically. It makes the fewest tool calls (60, and only 40 on LIBERO-PRO), edits the least (7 files), runs the simulator locally the most (16 to 21 times per run), and submits the most (3.6 times; 90\% of its best rounds are its own submissions). It treats SAM3 as a source of candidates rather than ground truth: its bundles carry one or two text prompts at the median, the fewest of any model, but the most geometric filtering (3.1 filters per bundle against 0 to 0.4 for the others, and a color gate in 83\% of RoboLab lineage bundles). Half of what it writes to memory is about what SAM3 misses, not how to phrase the prompt. This separation, SAM3 for what is present and code for which object the instruction means, made it the second-best model on RoboLab without a library. Its style is fast or never: 31\% of its RoboLab runs succeed by minute 40 and almost none after 100. Three costs follow. It shows the largest VAL-to-TEST gap of any model (4 to 7 points) and 20\% of its runs end on a policy worse than their best; its library lowers its RoboLab score ($-4.5$; \cref{tab:decomposition}), because its skills are task-named suite helpers whose most-used member has negative lift; and with Gemini-3.6-flash it is one of the two models for which best-of-3 beats Vanilla on RoboLab (\cref{sec:finding_resampling}).

\textbf{Kimi-k3: explores well, ships rarely, learns along the lineage.}
Kimi-k3 matches or beats Opus on five of seven self-exploration settings (\cref{tab:selfexp_models}) and adapts to hardware faults nearly as well as Opus when it has a library (94.4 against 97.0\% TEST), but not without one (72.0 against 96.7). Its dominant failure is delivery: 64\% of its graded RoboLab rounds never contained a policy file, because it probes the simulator in background processes and runs out of time; in the 40-minute best-of-3 draws, 90\% of attempts ship nothing. Its shipped Vanilla policies score a mean 42 VAL, against 63 for Opus, with closed-loop code (aperture check 88\%, retry 92\%, wrist-camera refinement 77\%). It looks at six to eight images per run, late, and edits after 68\% of them (baseline 25\%), the most corrective use of vision in the study. It is the model whose library gain grows along the lineage ($-26$ points in the first seven steps to $+35$ by step 45), whose skills are almost all useful (23 of 24 with three or more uses have positive lift), and the second to carry LIBERO-origin control skills onto the Isaac rig.

\textbf{DeepSeek-v4-flash: text-only, and it reads the library rather than importing it.}
Given open-set segmentation and a depth lifter, the one model that cannot see composes them into working manipulation: 98\% of its RoboLab policies call SAM3 and lift the mask through the depth camera, with eight distinct prompts per lineage bundle, it ships a policy in 85\% of runs (more often than Kimi-k3 or Gemini-3.6-flash), and on RoboLab it is ahead of Gemini-3.6-flash, the multimodal model of the same tier, without and with a library (though not in best-of-3). It writes skills it does not call again (51 to 71\% of its skills are never reused, the highest share) and imports library code in few of its policies (14\% on robosuite, 19\% on LIBERO, 34\% on LIBERO-PRO, 20\% on RoboLab, against 53 to 100\% for the other four models); its library gain on RoboLab is $+4.1$ TEST points on the 58 tasks its runs completed, and its 18-point gain on LIBERO-PRO, the largest of any model, does not arrive as copied text: the policies that copy the most from the inherited memories score lower (75) than those that copy nothing (87). Its memories are concrete constants (``descend the TCP to top\_z $-$ 0.016 for a 0.028-tall box so the jaws wrap the sides''). Its weakness is what it cannot see: when SAM3 misses it has no fallback, and it trails Gemini-3.6-flash on LIBERO and LIBERO-PRO.

\textbf{Gemini-3.6-flash: shortest policies, one perception route, skills it never reuses.}
Gemini-3.6-flash writes the shortest policies in the study (median 22 lines on the RoboLab lineages; 153 without a library), edits four times per run, opens an image in 2\% of runs, and never runs the simulator locally. SAM3 is its entire perception (39 VAL with it against 2 without), and its adaptation is the prompt: 11 distinct prompts per lineage bundle with synonym lists in 95\% of them (``grey bin'', ``gray bin'', ``plastic bin'', ``storage bin'', ``container''). Its one strong control habit is polling the success predicate to the end (86\% of lineage bundles): 68\% of its logged failures end with the motion complete and the jaws closed on air. It authors skills it mostly calls once, and on RoboLab it has called rig methods that do not exist. Its library gain on RoboLab is $+2.0$; its larger gains elsewhere ($+47$ on the LIBERO-RIG course, $+13$ on LIBERO-PRO) sit on the cells it fails alone. Its no-policy rounds (38\%) are delivery failures upstream of robotics.

\textbf{Failure by task category (RoboLab Vanilla, validation episodes).} Relative placement (stacking, spatial relations) scores 1.7 times each model's mean; tools and utensils score 28\% for Opus, 24 for GPT-5.6-sol, and 3 on average for the other three; reorientation scores 0.29 of each model's mean and thin or handled objects 0.48. Over 21,552 graded episodes the within-cell rank correlation between a layout's pass rate and the target's reach, lateral offset, yaw, or clutter averages 0.04 in magnitude. Where a policy logged its closing aperture, 78\% of failed episodes closed on air (under 5\,mm) against 18\% of successful ones. For Kimi-k3 and Gemini-3.6-flash, 54 and 55\% of zero-score runs never delivered a policy file.

\textbf{Verification schools and policy shape.} Two verification schools never mix within a model: Opus, Kimi-k3, and DeepSeek-v4-flash test the grasp by gripper aperture after closing (97, 88, and 74\%); GPT-5.6-sol and Gemini-3.6-flash test by the environment's holding flag (92 and 45\%). Opus refines from the wrist camera at pre-grasp in 75\% of RoboLab lineage bundles and Kimi-k3 in 77\%, the other three in 22\% on average. The library halves the policy and moves logic into imports: an Opus lineage policy is 293 lines importing 524 lines of skills, a GPT-5.6-sol one 124 plus 256, a Gemini-3.6-flash one 22 plus 137. Open-category selection and multi-object cells are a second source of failure, and stronger models suffer less from it: these cells score 0.83 of the model's mean for Opus and 0.66 for the others. Success falls monotonically from blocks (81\% for Opus) through bananas and bottles to mugs and bowls (55) to tools.

\textbf{Accumulation mattered more than perturbation.} Every model handled LIBERO-PRO's reworded instructions comparatively well; the harder problem was accumulating useful experience. On the hardest LIBERO-Goal task, Opus scored 0 of 5 with the stock instruction and 2 of 5 after it was rephrased seventeen steps into the LIBERO-PRO curriculum. The drawer, not the wording, was the difficulty, and what differed was the library: 171 entries at the stock run against 262 at the PRO run, including two self-authored skills that did not exist earlier. The comparison is not strictly like for like, since the two scores come from different bands and both runs timed out mid-refactor.

Every model received the same tasks, tools, budgets, and library machinery, but one model gained little from them. Inkling-small completed the robosuite and LIBERO tasks at $25.2 \pm 3.3$ and $6.4 \pm 2.2$\% TEST and was withdrawn partway through LIBERO-PRO; one of its lineages went 29 consecutive tasks without a valid policy. It wrote memories in about half its runs but almost never a code skill: after 47 tasks its three libraries held 0, 6, and 2 skills. In these runs, Inkling-small produced little working code for later
tasks to reuse. We exclude it from the remaining tables.

Most of the difficulty comes from the task rather than the model. On RoboLab, the cell explains 56\% of the variance in cell scores, the model explains 22\%, and their interaction explains the remaining 22\%. The average rank correlation of task difficulty between model pairs is 0.69. Models often struggle on the same tasks, but their failures do not form strictly nested sets. Seven cells are solved only by Opus, and five by no model.

\section{Library Contents and Reuse}
\label{app:library_contents}

Counts behind \cref{sec:finding_transfer}. On the first half of the RoboLab curriculum, pooled over each model's three lineages, Opus reused 202 distinct skills in 660 imports; the other models reused 26 to 126 skills. Across simulators (MuJoCo Panda to Isaac Sim Franka), 86 Opus imports of 12 skills reached 52\% of its RoboLab policies and 73 Kimi-k3 imports of 14 skills reached 42\%; GPT-5.6-sol made three such imports and the other two models none. The skills that cross are servo wrappers and grasp idioms, with per-skill lift of $+17$ to $+36$ points. Reuse is heavy-tailed: the top tenth of skills takes 40 to 68\% of uses, and independent lineages converge on the same concepts under different names. Memories propagate alongside code: 83 to 93\% of Opus's lineage policies carry identifiers from inherited memories that are absent from the matched no-library policies. Final programs import 16 to 21\% of a lineage's skill modules, and 96 to 133 skills per lineage have been shadowed by newer versions. A VLA library's memory file runs to a median 180{,}000 characters on RoboLab. Library gains concentrate on cells a model fails alone: on LIBERO-PRO, cells with a Vanilla score below 50\% gain $+33$ and $+39$ points for the two weakest models against $+2$ and $+4$ on cells they already solve, $+25$ against $+3$ for GPT-5.6-sol, and $+14$ against $-1$ for Kimi-k3. Kimi-k3's RoboLab library gain grows along the lineage from $-26$ points in its first seven steps to $+24$ to $+35$ from step~28 on. Frozen and re-graded on LIBERO-PRO, Opus's LIBERO-practiced lineage policies score 47.7\% TEST, against 40.7\% for its best-of-3 policies.

Most of what a library holds is never used. A VLA library is a handful of process skills and a very long memory file whose largest topic is the training recipe and checkpoint selection (\cref{tab:memory_topics}). Part of it records re-grading the same checkpoint, which gains about a point, within grading noise.

\section{Auto-Research: Supply, Time, and Task Structure}
\label{app:vla_tables}

These tables support \cref{sec:finding_code_vs_weights}. The groups are observational, not randomized interventions. A RoboLab session records 66 to 72 teacher episodes per agent-hour against about 490 on LIBERO, and ends with a median 109 to 122 successful demonstrations per task against 430 to 486. Among the 52 well-supplied RoboLab tasks, the reasoning gap within a placement count (1.2 and 9.4 percentage points) is within the task-to-task spread (permutation $p = 0.57$), while multi-placement demonstrations are roughly three times as long. Robosuite's precise multi-stage tasks also stay hard: nut assembly scores 3\% and two-arm handover 7\%, against 100\% for cube lifting.

\begin{table}[h]
\centering
\caption{More demonstrations, higher success. The 105 RoboLab tasks grouped by the number of successful demonstrations recorded by their VLA lineage session, with the VLA's mean TEST success (\%) per bin; the right half keeps only tasks whose teacher passes at least 90\% of TEST states.}
\label{tab:vla_dose}
\small
\begin{tabular}{lcccc}
\toprule
& \multicolumn{2}{c}{All tasks} & \multicolumn{2}{c}{Tasks with teacher $\geq 90\%$} \\
\cmidrule(lr){2-3} \cmidrule(lr){4-5}
Successful demonstrations per task & Tasks & VLA success & Tasks & VLA success \\
\midrule
none        &   9 &  0.0 & --- & --- \\
1 to 49     &  12 & 20.8 &  4 & 40.0 \\
50 to 99    &  20 & 41.0 & 10 & 60.0 \\
100 to 299  &  59 & 58.0 & 46 & 63.0 \\
300 or more &   5 & 94.0 &  5 & 94.0 \\
\midrule
all         & 105 & 47.2 & 65 & 63.5 \\
\bottomrule
\end{tabular}
\end{table}

\begin{table}[h]
\centering
\caption{What additional development time changes in AR. For each session with submissions before and after its first 5 hours on LIBERO or 10 hours on LIBERO-PRO and RoboLab, the best later submission is compared with the best earlier one. Entries give session counts and median in-session VAL gains in percentage points, grouped by the changes recorded in the 882 submissions. On LIBERO-PRO, retraining on the same data yields a median 7.4-point gain, close to the 8.0 points associated with collecting more data; returns are smallest on LIBERO, where scores are near the ceiling.}
\label{tab:vla_extra_time}
\small
\begin{tabular}{lcccccc}
\toprule
& \multicolumn{2}{c}{LIBERO} & \multicolumn{2}{c}{LIBERO-PRO} & \multicolumn{2}{c}{RoboLab} \\
\cmidrule(lr){2-3} \cmidrule(lr){4-5} \cmidrule(lr){6-7}
What changed & Sessions & Median & Sessions & Median & Sessions & Median \\
\midrule
Collect more data                  & 2 & 0.5 & 2 & 8.0 & 11 & 8.9 \\
Improve inference settings          & 0 & --- & 0 & --- &  4 & 4.4 \\
Average multiple model weights       & 0 & --- & 2 & 4.2 &  0 & --- \\
Re-train on the same data           & 4 & 2.5 & 4 & 7.4 &  6 & 2.3 \\
Re-grade same weights              & 2 & 0.0 & 0 & --- &  3 & 1.1 \\
\midrule
All                                  & 8 & 1.5 & 8 & 7.1 & 24 & 5.0 \\
\bottomrule
\end{tabular}
\end{table}

\begin{table}[h]
\centering
\caption{TEST success (\%) of the VLA on RoboLab by task structure. (a) The 52 tasks with a teacher at 90\% or more and at least 100 successful demonstrations per session, grouped by placements and instruction type; demo length is the median teacher demonstration in simulator steps; a \emph{reasoning} instruction requires inferring the target through an attribute, category, count, order, spatial relation, logical connective, or memory. (b) Five groups that share objects and scenes but differ in the number of placements required; scores average the Vanilla and Lineage AR sessions.}
\label{tab:vla_placements}
\small
{\small (a) By placements and instruction type}\\[3pt]
\begin{tabular}{llcccc}
\toprule
Placements & Instruction & Tasks & Demo length (steps) & Teacher & VLA \\
\midrule
one         & plain     & 16 & 204 & 96.2 & 76.9 \\
one         & reasoning & 22 & 201 & 99.1 & 75.7 \\
two or more & plain     &  6 & 585 & 98.3 & 40.0 \\
two or more & reasoning &  8 & 551 & 98.8 & 30.6 \\
\bottomrule
\end{tabular}
\\[10pt]
{\small (b) Matched tasks}\\[3pt]
\begin{tabular}{lllllll}
\toprule
& \multicolumn{3}{c}{One placement} & \multicolumn{3}{c}{Two or more placements} \\
\cmidrule(lr){2-4} \cmidrule(lr){5-7}
Matched tasks & Task & Teacher & VLA & Task & Teacher & VLA \\
\midrule
cube, banana, bowl   & cube or banana       & 100.0 & 75.0  & cube and banana         & 80.0  & 35.0 \\
                     &                      &     &     & banana then cube        & 100.0 & 30.0 \\
                     &                      &     &     & cube then banana        & 100.0 & 40.0 \\
\midrule
white bottles, pail  & any white bottle     & 100.0 & 85.0  & all three small bottles & 40.0  & 0.0 \\
\midrule
canned food, bin     & one can              & 100.0 & 100.0 & two cans                & 90.0  & 20.0 \\
                     &                      &     &     & three cans              & 90.0  & 0.0 \\
\midrule
pumpkins, bin        & small pumpkin        & 100.0 & 70.0  & both pumpkins           & 100.0 & 5.0 \\
                     & big pumpkin          & 100.0 & 40.0  &                         &     & \\
\midrule
apple, yogurt, pail  & throw away the apple & 100.0 & 95.0  & put away the snacks     & 100.0 & 65.0 \\
\bottomrule
\end{tabular}
\end{table}

Two further limits follow from the setup rather than the model. Every AR session fine-tunes $\pi_{0.5}$ on teacher demonstrations by supervised learning, so the agent's research choices all sit in the data: which tasks to collect more of, whether to add recovery rollouts and jittered starts, and how to fix the teacher. And without a judge, an agent's own teacher becomes its bound: a self-exploration session verifies episodes with camera checks it writes itself, and its teacher is its own code policy, which passes 3 to 10\% of the multi-step LIBERO-10 states, so the VLA sees few correct multi-step demonstrations and reaches 6 to 17\% there against 40 to 85\% on the single-stage suites (LIBERO-Spatial and LIBERO-Object).

\FloatBarrier
\section{Extended Related Work}
\label{app:sb2_related}
\subsection{Self-evolving Agents}
\label{sec:related-selfimprove}

\paragraph{Self-evolution in language model agents.}
The idea that an agent can improve its own performance through iterative self-evaluation predates the current generation of large language models but has been revitalized by their code-generation capabilities.
\citet{shinn2023reflexion} show that an agent can improve across repeated attempts with no weight update at all, by writing reflections on its own failures into an episodic memory that later attempts read.
\citet{zelikman2024stop} introduce the Self-Taught Optimizer (STOP), which recursively rewrites its own improvement procedure in code, demonstrating that language models can bootstrap increasingly effective optimization strategies without human-designed update rules.
More recently, \emph{scaffold engineering} has emerged as a complementary axis of self-improvement: rather than modifying model weights, agents evolve the scaffolding code that orchestrates their reasoning and tool use.
\citet{weng2026harness} provides a comprehensive survey of this paradigm, distinguishing weight-level from scaffold-level improvement and cataloging the design patterns that enable it.
\citet{lin2026agentic} operationalize the idea by building coding agents that automatically evolve their own harness code through observability-driven feedback loops.
A critical methodological concern is whether observed performance gains truly stem from improved strategies or merely from resampling.
\citet{lin2026disentangling} address this directly, showing that a substantial fraction of apparent self-evolution in LLM agents can be attributed to lucky draws rather than genuine capability acquisition, and propose controlled protocols to disentangle the two.

\subsection{Benchmarks for Autonomous Research Agents}
\label{sec:related-bench}

A parallel line of work develops benchmarks for measuring how well language model agents can conduct end-to-end research and engineering.
\citet{chan2024mlebench} introduce MLE-bench, which evaluates agents on Kaggle-style machine learning engineering tasks with automatic grading.
\citet{wijk2025rebench} propose RE-Bench, which measures frontier AI R\&D capabilities by asking agents to improve the performance of given ML systems under a time budget, emphasizing the distinction between one-shot coding and iterative self-improvement.
\citet{starace2025paperbench} scale the ambition further with PaperBench, which asks agents to replicate full AI research papers from scratch.
\citet{rank2026posttrainbench} focus specifically on post-training: PostTrainBench evaluates whether agents can automate the reward modeling, alignment, and fine-tuning pipeline for language models.
In the embodied domain, \citet{edgebench} introduce EdgeBench, which studies scaling laws for agents that learn from real-world interaction.

\section{Co-Evolving Loop: Extended Proposal}
\label{app:sb2_coevo}

\paragraph{From interaction to system updates.} Experience can enter the system through four complementary pathways (\cref{fig:sear_loop}). First, the agent consolidates reusable code, procedures, and lessons into its skill and memory libraries. Second, generated policies serve as executable teachers: repeated rollouts produce data for training VLAs or other learned action and world models, with outcome-aware selection, since teachers are imperfect; successful trajectories support imitation, and failures can inform recovery learning. Third, the interaction history (observations, tool calls, program revisions, and outcomes) could provide training signal for the reasoning agent itself, turning simulation or physical interaction into an environment for reinforcement learning over the agent's decisions. Fourth, the harness through which the agent observes and acts can itself be revised. Our rollout-driven $\pi_{0.5}$ fine-tuning instantiates one connection in this loop; establishing the full cycle requires several rounds in which updated tools return to the agent, its strategy adapts, and the system is evaluated on held-out tasks, with comparisons that freeze individual components to show whether joint adaptation helps beyond updating each component alone.

\paragraph{From task-level rollouts to reusable action skills.}
A key step in this loop is segmenting task instructions and collected trajectories into aligned, reusable action units. Rather than distilling an entire long-horizon task into a single behavior, we propose associating shorter trajectory segments with language descriptions of their local objectives. For example, placing a cup on a saucer can be decomposed into grasping the handle, lifting the cup, transporting it above the saucer, lowering it into contact, releasing the gripper, and withdrawing the arm. The reasoning agent or generated program connects these skills, checks their outcomes, and revises the sequence when execution deviates from the plan. Segmentation should therefore reflect meaningful changes in objectives and completion conditions rather than arbitrary temporal boundaries.

This decomposition establishes a useful division of labor. The language backbone handles long-horizon reasoning, task composition, and recovery decisions, while the action model translates local language-conditioned objectives into smooth, high-frequency, continuous control. The action model can become a generalist across reusable manipulation skills, retaining the feedback responsiveness needed for execution without having to determine the full task strategy. Training on instruction-aligned segments could make experience reusable across tasks that share manipulation components but require different plans. Within the proposed co-evolution loop, the agent can refine both the segmentation and the skill descriptions as it learns which abstractions its action model can execute reliably.

\paragraph{Harness evolution from experience.}
The interaction harness itself can also evolve as the agent discovers limitations in how it observes, acts, and learns. Repeated failures may reveal that static images omit important execution events, motivating video feedback or richer temporal traces. Objects falling outside the reachable workspace may motivate explicit safety zones and motion constraints that preserve the conditions for continued practice. Experience can similarly inform tool interfaces, logging, reset procedures, and failure diagnostics. These changes improve the infrastructure through which subsequent policies are developed and executed, complementing updates to skills, memories, and model tools. In \modelname{} only skills and memories persist between tasks (\cref{sec:sear}); the proposed loop would also carry model weights forward, and a system that can revise its own harness carries one more kind of state, which we would treat on the same grounds. Interaction informs changes to the harness, which in turn changes the evidence and actions available in future rounds. To evaluate this process reliably, harness revisions should be versioned and assessed while keeping final evaluation criteria and protected test information outside the agent's modification scope.

\end{document}